\documentclass[11pt]{article}

\usepackage[preprint]{acl}
\usepackage{times}
\usepackage{latexsym}
\usepackage[T1]{fontenc}
\usepackage[utf8]{inputenc}
\usepackage{microtype}
\usepackage{inconsolata}
\usepackage{graphicx}
\usepackage{amsmath}
\usepackage{booktabs}
\usepackage{multicol}
\usepackage{xcolor}
\usepackage{tabularray}
\usepackage{arydshln}
\usepackage{hyperref}
\usepackage[most]{tcolorbox}
\usepackage{setspace}
\usepackage{rotating}
\usepackage{lipsum}
\usepackage{makecell}
\usepackage{multicol}
\usepackage{xspace}
\usepackage{amsmath}
\usepackage{highlight}
\usepackage{colortbl}
\usepackage{stfloats}
\usepackage{amssymb}
\usepackage{enumitem}
\usepackage{multirow}

\newcommand{\filbench}[1]{\textsc{FilBench}}
\newcommand{\nummodels}[1]{27}
\newcommand{\numcategories}[1]{4}
\newcommand{\numsubtasks}[1]{12}
\newcommand{\numinstances}[1]{197.6k}

\newcommand*\samethanks[1][\value{footnote}]{\footnotemark[#1]}
\definecolor{cblue}{HTML}{133844}
\definecolor{cwarmblue}{HTML}{00BDB6}
\definecolor{clightblue}{HTML}{D1F9F1}
\definecolor{ccrest}{HTML}{FFE2C8}
\definecolor{cgreen}{HTML}{DFF2EA}
\definecolor{ccherry}{HTML}{F2CAD8}
\definecolor{cviolet}{HTML}{F2ECF8}
\definecolor{cdarkviolet}{HTML}{681FB1}
\definecolor{cslate}{HTML}{546072}
\newtcolorbox{promptbox}[1][]{
  colframe=cslate,    % Frame color
  colback=white,     % Background color
  coltitle=white,            % Color of the title text
  title=#1,                  % Optional title
  rounded corners,           % Corner style
  boxrule=0.5mm,             % Frame thickness
  boxsep=5pt,                % Space between content and box
  toptitle=1mm,              % Space above the title
  bottomtitle=1mm,           % Space below the title
  left=10pt,                 % Left padding
  right=10pt,                % Right padding
  top=5pt,                   % Top padding
  bottom=5pt,                % Bottom padding
}
\newtcolorbox{promptboxcn}[1][]{
  colframe=cwarmblue,    % Frame color
  colback=white,     % Background color
  coltitle=black,            % Color of the title text
  title=#1,                  % Optional title
  rounded corners,           % Corner style
  boxrule=0.5mm,             % Frame thickness
  boxsep=5pt,                % Space between content and box
  toptitle=1mm,              % Space above the title
  bottomtitle=1mm,           % Space below the title
  left=10pt,                 % Left padding
  right=10pt,                % Right padding
  top=5pt,                   % Top padding
  bottom=5pt,                % Bottom padding
}
\newtcolorbox{promptboxck}[1][]{
  colframe=ccherry,    % Frame color
  colback=white,     % Background color
  coltitle=black,            % Color of the title text
  title=#1,                  % Optional title
  rounded corners,           % Corner style
  boxrule=0.5mm,             % Frame thickness
  boxsep=5pt,                % Space between content and box
  toptitle=1mm,              % Space above the title
  bottomtitle=1mm,           % Space below the title
  left=10pt,                 % Left padding
  right=10pt,                % Right padding
  top=5pt,                   % Top padding
  bottom=5pt,                % Bottom padding
}
\newtcolorbox{promptboxrc}[1][]{
  colframe=ccrest,    % Frame color
  colback=white,     % Background color
  coltitle=black,            % Color of the title text
  title=#1,                  % Optional title
  rounded corners,           % Corner style
  boxrule=0.5mm,             % Frame thickness
  boxsep=5pt,                % Space between content and box
  toptitle=1mm,              % Space above the title
  bottomtitle=1mm,           % Space below the title
  left=10pt,                 % Left padding
  right=10pt,                % Right padding
  top=5pt,                   % Top padding
  bottom=5pt,                % Bottom padding
}
\newtcolorbox{promptboxgn}[1][]{
  colframe=cviolet,    % Frame color
  colback=white,     % Background color
  coltitle=black,            % Color of the title text
  title=#1,                  % Optional title
  rounded corners,           % Corner style
  boxrule=0.5mm,             % Frame thickness
  boxsep=5pt,                % Space between content and box
  toptitle=1mm,              % Space above the title
  bottomtitle=1mm,           % Space below the title
  left=10pt,                 % Left padding
  right=10pt,                % Right padding
  top=5pt,                   % Top padding
  bottom=5pt,                % Bottom padding
}
\newcommand{\cold}[1]{\gradientcelld{#1}{-0.1}{0}{0.8}{ccherry}{white}{clightblue}{100}}

\newcommand{\sea}{\raisebox{-1.5pt}{\includegraphics[height=1.05em]{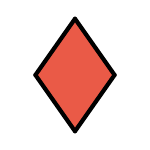}}\xspace}
\newcommand{\mul}{\raisebox{-1.5pt}{\includegraphics[height=1.05em]{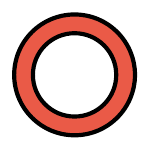}}\xspace}
\newcommand{\huggingface}{\raisebox{-1.5pt}{\includegraphics[height=1.05em]{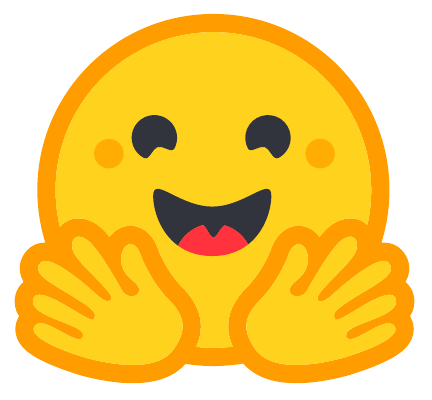}}}
\newcommand{\github}{\raisebox{-1.5pt}{\includegraphics[height=1.05em]{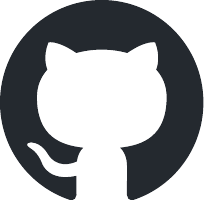}}}

\title{
    \filbench{}: Can LLMs Understand and Generate Filipino?
}

\author{
 Lester James V. Miranda\textsuperscript{1}\thanks{Equal contributions.}\quad
 Elyanah Aco\textsuperscript{2}\samethanks{}\quad
 Conner Manuel\textsuperscript{3}\samethanks{}\\
 \bf Jan Christian Blaise Cruz\textsuperscript{4,5}\thanks{Senior authors.}\quad
 \bf Joseph Marvin Imperial\textsuperscript{4,6,7}\samethanks{}\\\\
 \textsuperscript{1}Allen Institute for AI\quad
 \textsuperscript{2}Nara Institute of Science and Technology\quad
 \textsuperscript{3}Together AI\quad\\
 \textsuperscript{4}SEACrowd\quad
 \textsuperscript{5}MBZUAI\quad
 \textsuperscript{6}University of Bath\quad
 \textsuperscript{7}National University, Philippines\quad
 \\\\
{\small\github{}~~\textbf{Code}\hspace{0.5em}\href{https://github.com/filbench/filbench-eval}{\texttt{filbench/filbench-eval}}}\quad
{\small\huggingface{}~~\textbf{Leaderboard}\hspace{0.5em}\href{https://huggingface.co/spaces/UD-Filipino/filbench-leaderboard}{\texttt{UD-Filipino/filbench-leaderboard}}}
}

\begin{document}
\maketitle
\begin{abstract}
    Despite the impressive performance of LLMs on English-based tasks, little is known about their capabilities in specific languages such as Filipino.
    In this work, we address this gap by introducing \textbf{\filbench{}}, a Filipino-centric benchmark designed to evaluate LLMs across a diverse set of tasks and capabilities in Filipino, Tagalog, and Cebuano.
    We carefully curate the tasks in \filbench{} to reflect the priorities and trends of NLP research in the Philippines such as Cultural Knowledge, Classical NLP, Reading Comprehension, and Generation.
    By evaluating \nummodels{} state-of-the-art LLMs on \filbench{}, we find that several LLMs suffer from reading comprehension and translation capabilities.
    Our results indicate that \filbench{} is challenging, with the best model, GPT-4o, achieving only a score of 72.23\%.
    Moreover, we also find that models trained specifically for Southeast Asian languages tend to underperform on \filbench{}, with the highest-performing model, SEA-LION v3 70B, achieving only a score of 61.07\%.
    Our work demonstrates the value of curating language-specific LLM benchmarks to aid in driving progress on Filipino NLP and increasing the inclusion of Philippine languages in LLM development.
\end{abstract}

\addtocontents{toc}{\protect\setcounter{tocdepth}{0}}

\begin{figure}[t]
    \centering
    \includegraphics[width=0.85\linewidth]{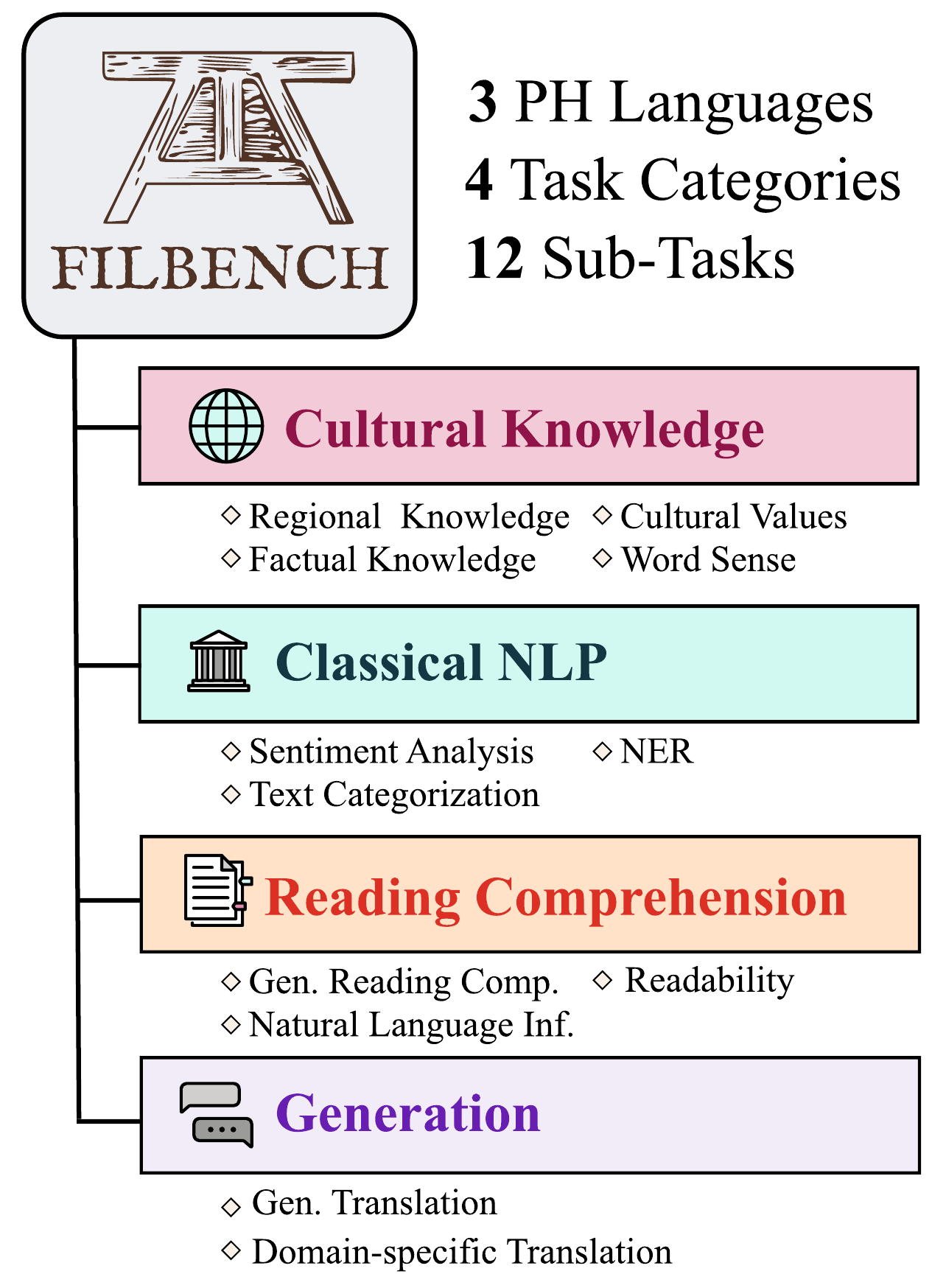}
    \caption{\textbf{Overview of \filbench{}.} In order to comprehensively assess the full capabilities of LLMs on Philippine languages, we curate an evaluation suite consisting of \numcategories{} categories and \numsubtasks{} subtasks across \textbf{Filipino, Tagalog, and Cebuano} based on the research priorities of the Philippine NLP community (\S\ref{sec:filbench-research-priorities}).}
    \label{fig:filbench_main}
\end{figure}

\section{Introduction}
While large language models (LLMs) have shown impressive performance on a variety of English-based tasks and capabilities, their effectiveness remains largely unexplored for low-resource languages such as Filipino.
This knowledge gap exists for two reasons.
First, most low-resource languages, especially Filipino-centric benchmarks, developed before the ChatGPT-era \citep[\textemdash 2022][]{gururaja-etal-2023-build} are ill-posed for current LLM evaluation despite their value in understanding language system capabilities.
% First, early Filipino-centric benchmarks developed before the ChatGPT-era \citep[\textemdash 2022][]{gururaja-etal-2023-build} lack the necessary formulation for current LLM evaluation despite their value in understanding language system capabilities.
Second, existing multilingual LLM benchmarks either exclude Filipino entirely \citep[\textit{inter alia}]{liu2025seaexam,huang2025benchmax} or fail to provide sufficient task and/or language diversity \citep{susanto2025sea}.
Filipino is an important language to consider for LLM evaluation not only because of its unique linguistic properties such as its voice marking system \citep{bardaji2024}, but also due to its large speaker population with more than 28 million speakers in the Philippines and over 2 million speakers abroad \citep{PSA2020Household}.

In this work, we perform a comprehensive study of the strengths and limitations of LLM capabilities on Filipino-centric tasks.
We introduce an evaluation suite called \textbf{\filbench{}}, consisting of \numcategories{} categories and \numsubtasks{} diverse sub-tasks that are formulated for LLM evaluation.
The choice of tasks to include in \filbench{} is based on our study of research trends and priorities in Filipino NLP (\S\ref{sec:filbench-research-priorities}, \S\ref{sec:appendix-fil-nlp-research}).
Evaluating models on \filbench{} reveals significant gaps in LLM performance, for instance, in text generation capabilities.
The contributions of this study are three-fold:

\begin{itemize}[leftmargin=3mm]
    \item We close the \textbf{resource gap} by curating Filipino test sets across four broad task categories: Cultural Knowledge Assessment (CK), Classical NLP (CN), Reading Comprehension (RC), and Generation (GN).
          We transform these datasets into a unified task format aligned with standard LLM evaluation practices across literature.
          Our evaluation suite, \textbf{\filbench{}}, consists of test instances across \numcategories{} categories and \numsubtasks{} sub-tasks (\S\ref{sec:filbench}).
    \item We bridge the \textbf{evaluation gap} by evaluating \nummodels{} state-of-the-art LLMs on \filbench{} (\S\ref{sec:results}).
          We find that the best model, GPT-4o, only achieve around 72.23\% aggregated performance while the best Southeast-Asian model, SEA-LION v3 70B, only obtains a score of 61.07\%.
    \item We provide \textbf{analyses and insights} to the strengths and weaknesses of LLMs when presented with Filipino-centric tasks and test cases (\S\ref{sec:analysis}).
          Notably, we find that text generation suffers the most, with the lowest scores across models due to failure modes such as hallucination and poor instruction-following.
\end{itemize}

\filbench{} demonstrates the value of constructing language-specific benchmarks to reveal gaps in language model capabilities and benefit the wider speaker community.
More importantly, we hope that this work aids in improving the state of Filipino NLP and increase the inclusion of Philippine languages in LLM development.

\section{Background}

\paragraph{Languages in the Philippines.}
The Philippines is home to approximately 117 million language speakers across more than 185 distinct languages \citep{ethnologue,mcfarland2008linguistic,metila2016challenge}.
One of its official languages is Filipino (\texttt{FIL}), which is a standardized form of Tagalog (\texttt{TGL}) and used mainly in Metro Manila.\footnote{The designation of Filipino and Tagalog as separate languages is often a point of contention, although they are linguistically similar \citep{Villafania2007}. We follow the official view of the \textit{Komisyon ng Wikang Filipino} (Commision on the Filipino Language) and treat them as separate.}
Aside from Filipino and Tagalog, Cebuano (\texttt{CEB}) is the second most widely spoken language in the Philippines with over 28 million speakers.
It is part of the Visayan language family and is spoken mainly in regions of Cebu, Siquijor, and Bohol among many others \citep{pilar-etal-2023-cebuaner}. As part of the same subgroup of Philippine languages, Tagalog and Cebuano share similar linguistic characteristics such as shared vocabulary and comparable word formulation processes and affixation rules, among others \cite{bacalla2019morpoanalisis, imperial-kochmar-2023-automatic}.
In our work, we focus on the these three languages because they cover the majority of Filipino speakers, representing approximately 61\% of the country's population \citep{PSA2020Household}.

\paragraph{Task Formulation in LLM Evaluation.}
In order to standardize how each test example is presented to an LLM, it must first be formatted into a consistent prompt structure or \textit{formulation}.
Multiple-choice formulation (MCF) is a common standard in evaluating LLMs across a vast array of tasks \citep{gu2024olmes,open-llm-leaderboard-v2}.
In MCF, a question is posed with answers presented as labeled choices, where scoring is done by comparing the LLM's choice to the gold label.
For evaluating LLMs on generative tasks such as translation, one approach is to write an instruction prompting an LLM to translate a given text from a source language to a target language.
Then, the generated output by the LLM is compared against the reference translation using various machine translation metrics \citep{papineni-etal-2002-bleu,lin-2004-rouge}.

\begin{table*}
    \centering
    \resizebox*{\linewidth}{!}{
        \begin{tabular}{llllr}
            \toprule
            \textbf{Category}                  & \textbf{Sub-Task}         & \textbf{Dataset}                                              & \textbf{Languages}         & \textbf{\# Instances} \\
            \midrule
            Classical NLP (CN)                 & Text Classification       & Dengue Filipino \cite{livelo-cheng-2018-dengue}               & \texttt{FIL}               & 4,015                 \\
                                               &                           & BalitaNLP \cite{bunag2023news}                                & \texttt{TGL}               & 70,352                \\
                                               &                           & SIB-200 \cite{adelani-etal-2024-sib}                          & \texttt{CEB}, \texttt{FIL} & 99                    \\
                                               & Named-Entity Recognition  & CebuaNER \cite{pilar-etal-2023-cebuaner}                      & \texttt{CEB}               & 1,310                 \\
                                               &                           & TLUnified-NER \cite{miranda-2023-developing}                  & \texttt{TGL}               & 1,579                 \\
                                               &                           & Universal NER \cite{mayhew-etal-2024-universal}               & \texttt{CEB}, \texttt{TGL} & 105                   \\
                                               & Sentiment Analysis        & FiReCS \cite{cosme-deleon-2024-products}                      & \texttt{FIL}               & 7,340                 \\
            \midrule
            Cultural Knowledge Assessment (CK) & Regional Knowledge        & INCLUDE \cite{romanou2024include}                             & \texttt{TGL}               & 510                   \\
                                               & Factual Knowledge         & Global MMLU \cite{singh2024global}                            & \texttt{TGL}               & 14,042                 \\
                                               & Cultural Values           & KALAHI \cite{montalan2024kalahi}                              & \texttt{TGL}               & 150                   \\
                                               & Word-sense Disambiguation & StingrayBench \cite{cahyawijaya2024thank}                     & \texttt{TGL}               & 100                   \\
            \midrule
            Reading Comprehension (RC)         & Readability               & Cebuano Readability Corpus \cite{imperial-etal-2022-baseline} & \texttt{CEB}               & 350                   \\
                                               & Reading Comprehension     & Belebele \cite{bandarkar-etal-2024-belebele}                  & \texttt{CEB, FIL}               & 1,800                 \\
                                               & NLI                       & NewsPH NLI \cite{cruz-etal-2021-news}                         & \texttt{FIL}               & 90,000                \\
            \midrule
            Generation (GN)                    & Document translation      & NTREX-128 \cite{federmann-etal-2022-ntrex}                    & \texttt{FIL}               & 1,997                 \\
                                               & Realistic translation     & Tatoeba \cite{tiedemann-2020-tatoeba}                         & \texttt{CEB}, \texttt{TGL} & 2,876                 \\
                                               & Domain-specific transl.   & TICO-19 \cite{anastasopoulos-etal-2020-tico}                  & \texttt{TGL}               & 971                   \\
            \bottomrule
        \end{tabular}
    }
    \caption{
        \textbf{Fine-grained overview of \filbench{}.}
        Our curation effort involves expert-annotated or validated datasets across a diverse range of sub-tasks and categories basd on a quantitative analysis of the priorities of the Filipino NLP community (\S\ref{sec:appendix-fil-nlp-research}), allowing us
        to comprehensively evaluate LLM capabilities on Filipino-centric tasks.
    }
    \label{table:filbench-datasets}
\end{table*}

\section{The \filbench{} Evaluation Suite}
\label{sec:filbench}

Our design philosophy for \filbench{} centers on two core principles:
(1) developing an impactful benchmark that aligns with the research priorities within the Philippine context (\S\ref{sec:filbench-research-priorities}), ensuring that a model excelling in \filbench{} is likely to perform effectively across a wide range of Filipino applications and
(2) maintaining data quality and richness by incorporating diverse sub-tasks (\S\ref{sec:filbench-categories}) that were annotated by experts or native speakers.
\autoref{table:filbench-datasets} shows all the datasets and tasks included in \filbench{}.
Example task formulation for each sub-task is shown in Appendix \ref{sec:appendix-prompts}.

\subsection{Research Priorities in Filipino NLP}
\label{sec:filbench-research-priorities}

In order to determine which tasks to include in \filbench{}, we perform a survey of the research trends in NLP research on Philippine languages from 2006--2023.
Our methodology involves scraping Scopus-indexed papers and \textsuperscript{$\star$}ACL/EMNLP publications and classifying their NLP sub-field based on common ACL tracks.
We find that classical NLP tasks such as information extraction and sentiment analysis are widely studied, as well as a variety of translation tasks.
Then, we devise a taxonomy consisting of four major categories that encompass more recent trends in Philippine NLP research.
More details about our methodology and findings can be found in Appendix \ref{sec:appendix-fil-nlp-research}.

\subsection{\filbench{} Categories}
\label{sec:filbench-categories}

\paragraph{Cultural Knowledge Assessment (CK).}
This category tests a language model's ability to recall factual and culturally-specific information.
Studies have consistently found that LLMs predominantly trained on English text are strongly biased towards Western values and perspectives, especially when prompted in English \cite{cao-etal-2023-assessing}.
Cultural misalignment between LLMs and users can lead to unintended harms such as norm violations \cite{qiu-etal-2025-evaluating} and socio-economic exclusion \cite{dammu-etal-2024-uncultured}.
For CK, we curate a variety of examples that test an LLM's regional and factual knowledge \citep{romanou2024include,singh2024global}, understanding of Filipino-centric values \citep{montalan2024kalahi}, and word-sense disambiguation \citep{cahyawijaya2024thank}.

\paragraph{Classical NLP (CN).}
This category encompasses a variety of information extraction and linguistic tasks such as named entity recognition (NER), sentiment analysis, and text categorization that were traditionally performed using specialized trained models.
These tasks have been prominent in Philippine NLP research over the past decade \citep{roxas-etal-2021-science}, and LLMs have recently begun to be employed in this domain \citep[\textit{inter alia}]{ashok2023promptner,zhang2023sentiment,wang2023gpt}.
For CN, we include expert-annotated NER datasets such as CebuaNER \citep{pilar-etal-2023-cebuaner}, TLUnified-NER \citep{miranda-2023-developing}, and Universal NER \citep{mayhew-etal-2024-universal}.
We also take the Filipino and Cebuano subsets of SIB-200 \citep{adelani-etal-2024-sib}, and the text-only subset of Balita NLP \citep{bunag2023news}.

\paragraph{Reading Comprehension (RC).}
This category evaluates a language model's ability to understand and interpret Filipino text, focusing on tasks such as readability, comprehension, and natural language inference (NLI).
These tasks are crucial for assessing how well a model can process and generate human-like understanding of written content.
For RC, we include datasets like the Cebuano Readability Corpus \citep{imperial-etal-2022-baseline}, Belebele \citep{bandarkar-etal-2024-belebele}, and NewsPH NLI \citep{cruz-etal-2021-news}, which provide a comprehensive evaluation of reading comprehension capabilities in the Filipino context.

\paragraph{Generation (GN).}
Although generative LLM tasks usually include summarization and conversational generation, evaluation test sets in Filipino are sparse.
However, machine translation is one of the most dominant areas of NLP research in the Philippines \citep[\textit{inter alia}]{oco-roxas-2018-survey,baliber-etal-2020-bridging,aji-etal-2023-current}.
Recently, LLMs have gained traction for its use as automatic translators, as opposed to training specialized translation models \citep{zhu2023multilingual, he-etal-2024-exploring,alves2024tower}.
Hence, we dedicate a large portion of \filbench{} for testing an LLM's ability to faithfully translate texts, either from English to Filipino (\texttt{ENG} $\to$ \texttt{FIL}) or from Cebuano to English (\texttt{CEB} $\to$ \texttt{ENG}).
We include a diverse set of test examples, ranging from documents \citep{federmann-etal-2022-ntrex}, realistic texts collected from volunteers \citep{tiedemann-2020-tatoeba}, and domain-specific text \citep{anastasopoulos-etal-2020-tico}.

\subsection{\filbench{} Scoring}
The CN, CK, and RC categories follow the MCF task formulation, so we score an LLM's performance for these categories by computing the accuracy, i.e., the number of correct answers divided by the total number of examples.
For GN, we compute the ROUGE-L score between the LLM-generated text and the gold reference text.
%We chose ROUGE-L instead of other metrics such as BERTScore because it does not require Filipino-specific embedding models that are still unavailable.
All per-category metrics range from $0$ to $1$.
In order to create a representative, single evaluation score, we perform a weighted average based on the number of examples across results as shown in \autoref{eq:filbench}:

{\small
\begin{equation}
    \label{eq:filbench}
    \text{\filbench{} Score} = 100 \times \frac{\sum_{i \in \{\text{CN}, \text{CK}, \text{GN}, \text{RC}\}} n_i \cdot S_i}{\sum_{i \in \{\text{CN}, \text{CK}, \text{GN}, \text{RC}\}} n_i}
\end{equation}}

where $n_i$ is the number of examples in category $i$ and $S_i$ is the score for category $i$.

\begin{table*}
    \centering
    \resizebox*{\linewidth}{!}{
        \begin{tabular}{lrrrrr}
            \toprule
            % \begin{noindent}
        \textbf{Model}                              & \textbf{\thead[r]{\textsc{FilBench}\\Score}} & \textbf{\thead[r]{Cultural\\Knowledge}} & \textbf{\thead[r]{Classical\\NLP}} & \textbf{\thead[r]{Reading\\Comp.}} & \textbf{Generation} \\
        % \end{noindent}
            \midrule
            \mul\href{https://platform.openai.com/docs/models}{gpt-4o-2024-08-06}                                                                  & \textbf{72.73{\small$\pm$1.66}} & 73.29{\small$\pm$3.01}          & 89.03{\small$\pm$2.05}          & \textbf{80.12{\small$\pm$0.90}} & \textbf{46.48{\small$\pm$0.60}} \\
            \mul\href{https://huggingface.co/meta-llama/Llama-4-Maverick-17B-128E-Instruct-FP8}{meta-llama/Llama-4-Maverick-17B-128E-Instruct-FP8} & 67.67{\small$\pm$1.04}          & \textbf{76.75{\small$\pm$3.04}} & 87.28{\small$\pm$0.26}          & 72.99{\small$\pm$0.18}          & 33.67{\small$\pm$0.71}          \\
            \mul\href{https://huggingface.co/meta-llama/Llama-4-Scout-17B-16E-Instruct }{meta-llama/Llama-4-Scout-17B-16E-Instruct}                & 63.20{\small$\pm$1.05}          & 74.31{\small$\pm$3.14}          & 87.88{\small$\pm$0.25}          & 70.86{\small$\pm$0.18}          & 19.75{\small$\pm$0.63}          \\
            \mul\href{https://huggingface.co/Qwen/Qwen2.5-72B-Instruct}{Qwen/Qwen2.5-72B-Instruct}                                                 & 63.08{\small$\pm$0.99}          & 73.11{\small$\pm$3.22}          & 88.60{\small$\pm$0.24}          & 75.62{\small$\pm$0.17}          & 14.98{\small$\pm$0.33}          \\
            \sea\href{https://huggingface.co/aisingapore/Llama-SEA-LION-v3-70B-IT}{aisingapore/Llama-SEA-LION-v3-70B-IT}                           & 61.07{\small$\pm$0.95}          & 76.78{\small$\pm$3.02}          & 89.99{\small$\pm$0.23}          & 53.56{\small$\pm$0.19}          & 23.95{\small$\pm$0.34}          \\
            \mul\href{https://huggingface.co/Tower-Babel/Babel-83B-Chat}{Tower-Babel/Babel-83B-Chat}                                               & 60.85{\small$\pm$0.96}          & 75.21{\small$\pm$3.11}          & 88.81{\small$\pm$0.25}          & 64.85{\small$\pm$0.19}          & 14.53{\small$\pm$0.29}          \\
            \mul\href{https://huggingface.co/meta-llama/Llama-3.1-70B-Instruct}{meta-llama/Llama-3.1-70B-Instruct}                                 & 59.66{\small$\pm$1.17}          & 72.16{\small$\pm$3.21}          & \textbf{90.27{\small$\pm$0.83}} & 52.17{\small$\pm$0.28}          & 24.03{\small$\pm$0.37}          \\
            \sea\href{https://huggingface.co/sail/Sailor2-20B-Chat}{sail/Sailor2-20B-Chat}                                                         & 58.61{\small$\pm$1.06}          & 66.43{\small$\pm$3.41}          & 89.03{\small$\pm$0.25}          & 63.03{\small$\pm$0.19}          & 15.95{\small$\pm$0.38}          \\
            \mul\href{https://huggingface.co/Qwen/Qwen2.5-32B-Instruct}{Qwen/Qwen2.5-32B-Instruct}                                                 & 57.88{\small$\pm$1.45}          & 66.83{\small$\pm$3.45}          & 89.32{\small$\pm$1.99}          & 70.59{\small$\pm$0.18}          & 4.79{\small$\pm$0.17}           \\
            \sea\href{https://huggingface.co/aisingapore/Gemma-SEA-LION-v3-9B-IT}{aisingapore/Gemma-SEA-LION-v3-9B-IT}                             & 56.14{\small$\pm$1.53}          & 64.44{\small$\pm$3.43}          & 88.55{\small$\pm$0.25}          & 54.46{\small$\pm$0.20}          & 17.10{\small$\pm$2.25}          \\
            \bottomrule
        \end{tabular}
    }
    \caption{
        \textbf{Performance of state-of-the-art LLMs on Filipino-centric tasks.}
        We evaluate several models with different multilingual capabilities (multilingual \mul, SEA-specific \sea), sizes (1.5B to 400B), and accessibility (open-source vs. commercial).
        Full results can be found in \autoref{table:main_results_all_agg}.
    }
    \label{table:main_results_top10}
\end{table*}

\section{Results: Performance of State-of-the-Art LLMs on \filbench{}}
\label{sec:results}

In order to understand what kind of LLMs perform well in Filipino,  we select a variety of open-source and commercial LLMs to ensure broad coverage across parameter sizes and language capabilities.
We also include a number of SEA-specific models that were trained to cater to Southeast Asian languages, including Filipino.
% For parameter size, we classify models as low ($\leq$13B), medium (13B-32B), or high ($\geq$70B).
% We also classify LLMs based on their language capabilities, categorizing them as either Multilingual or SEA-specific, with the latter being a subset of the former.
% We tag a model as Multilingual if it was presented as such and if the technical report includes multilingual evaluations.
% In contrast, we classify a model as SEA-specific if the majority of the training data used to fine-tune the model originated from Southeast Asia, or if it was specifically developed for languages in the region.
A total of \nummodels{} models are chosen for evaluation.
\autoref{table:filbench_models} in the Appendix shows the full details of the evaluated models.

\autoref{table:main_results_top10} shows the scores obtained by the top ten models on \filbench{}.
The full results for all \nummodels{} models can be seen in \autoref{table:main_results_all_agg} of the Appendix.
The best performing model is GPT-4o (72.23\%), closely followed by Llama 4 Maverick (67.67\%).
Moreover, the highest scoring open-source dense model is Qwen2.5 72B (63.08\%), while SEA-LION v3 70B is the best SEA-specific model (61.07\%).

\begin{figure}[t]
    \centering
    \includegraphics[width=\linewidth]{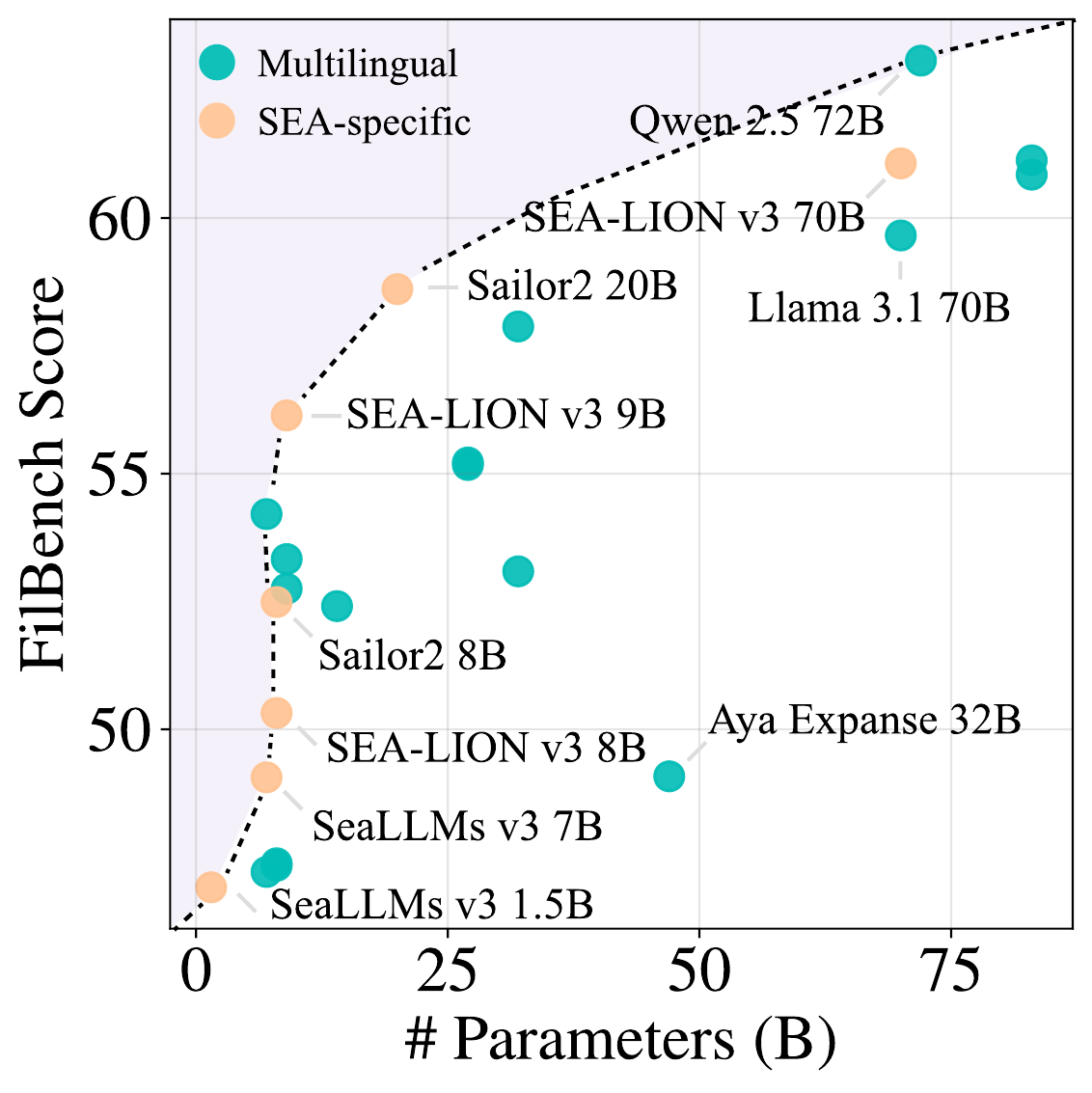}
    \caption{
        % \filbench{} score to Parameter Size (B) for several dense open-source language models.
        \textbf{Parameter-efficiency of LLMs with respect to \filbench{}.}
        SEA-specific models are at the \textcolor{cdarkviolet}{Pareto frontier} of parameter-efficiency.
        However, the best SEA-specific model still underperforms on \filbench{} with a score of 61.07\%.
    }
    \label{fig:impact_of_lm_size}
\end{figure}

\paragraph{Finding \#1: Larger models dominate \filbench{}.}
\autoref{fig:impact_of_lm_size} shows the \filbench{} score to Parameter Size (B) for several dense open-source language models with known sizes.
Our findings suggest that parameter size strongly correlates with \filbench{} performance, with a Spearman $\rho$ of 0.810.
However, this correlation is not perfect as we observe some smaller models to be competitive with larger counterparts as observed in Qwen 2.5 32B having similar performance to Llama 3.1 70B.

\begin{figure}[t]
    \centering
    \includegraphics[width=0.9\linewidth]{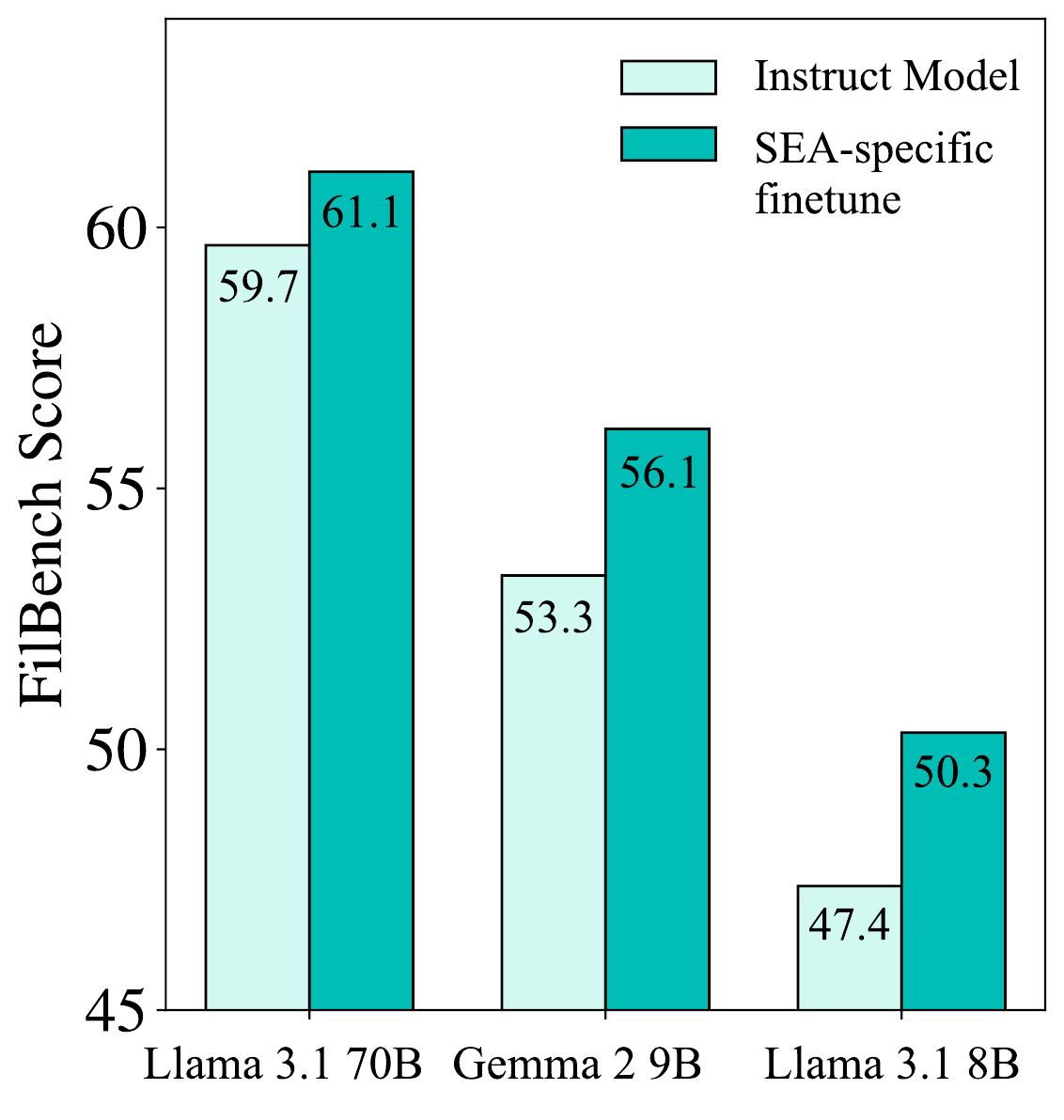}
    \caption{
        \textbf{Effect of language-specific finetuning.}
        Performance comparison between a base instruction model and its finetuned version (SEA-LION v3).
        Language-specific finetuning from a multilingual base model can improve performance on \filbench{}.
    }
    \label{fig:continuous_ft}
\end{figure}

\begin{figure*}[t]
    \centering
    \includegraphics[width=0.95\linewidth]{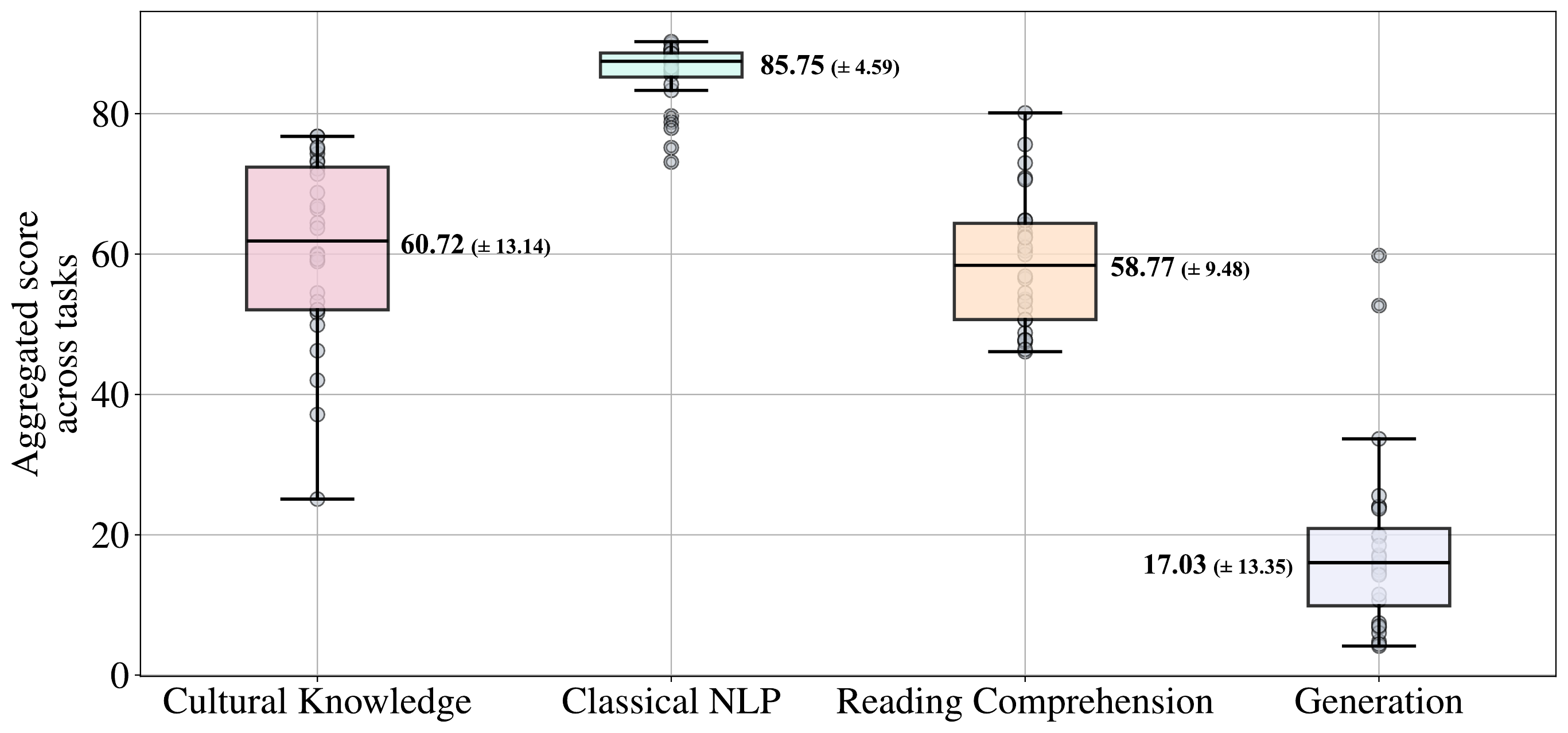}
    \caption{
        \textbf{Performance trends in \filbench{}.}
        Model performance (aggregated) across the four categories of \filbench{}, along with the average performance for each category.
        LLMs tend to perform well in Classical NLP tasks, but suffer poor performance in Generation tasks.
    }
    \label{fig:performance_trends}
\end{figure*}

\paragraph{Finding \#2: Language-specific finetuning improves \filbench{} performance.}
SEA-specific models tend to be more parameter-efficient as they perform better than non-specialized LLMs on \filbench{}.
This trend is more apparent for smaller models within the 7B to 9B range, as shown in \autoref{fig:impact_of_lm_size}.
In addition, SEA-specific models such as Sailor2 20B, SEA-LION v3 9B, and SeaLLMs v3 1.5B sit near the Pareto frontier in terms of performance and size.
Despite these results, the best performing SEA-specific model still underperforms on \filbench{}, as in the case of SEA-LION v3 70B with a score of 61.07\%.
In addition, we also find that continuous finetuning of an existing multilingual LLM on SEA-specific data improve \filbench{} performance, as observed in the SEA-LION model family, which are finetunes of Llama 3.1 and Gemma 2, in  \autoref{fig:continuous_ft}.
These findings show a promising direction for building Filipino-focused LLMs, as it provides a resource-efficient path without training entirely new models from scratch.

\paragraph{Finding \#3: Models tend to follow a consistent trend in \filbench{} performance across categories.}
\autoref{fig:performance_trends} suggests that most models have a consistent trend in \filbench{} performance, i.e., they tend to score well in CK, CN, and RC categories, yet are worse on GN.
This is more apparent in generative (GN) tasks, where most models tend to struggle with an average performance of 17.03\%.
On the other hand, models tend to perform well in CK (60.72\%) and CN (85.75\%) categories, indicating high-level of understanding of Filipino-centric cultural entities and values.
Model performance on CK tends to be more dispersed with one of the largest standard deviation ($\pm$13.14).
These findings suggest that model capabilities are not uniform across categories for Filipino, indicating significant room for improvement on model training.

\section{Analysis: When do LLMs Perform Well or Worse on Filipino Language Tasks?}
\label{sec:analysis}

% In this section, we look into key aspects of model performance such as the consistency of model answers in CN, CK and RC categories (\S\ref{sec:analysis-consistency}), the common model failure modes on the GN category (\S\ref{sec:analysis-failure}), and performance trends across Tagalog and Cebuano (\S\ref{sec:analysis-languages}), where the latter is less-resourced than the former.

\subsection{Do models consistently agree with one another on Filipino language tasks?}
\label{sec:analysis-consistency}

\paragraph{Set-up.}
In order to understand whether models are consistently reliable in answering test cases in \filbench{}, we compute the inter-rater reliability using Fleiss' $\kappa$ across a given set of models.
The first group consists of SEA-specific models (see models marked with \sea in \autoref{table:filbench_models}) while the second group includes the top-five non-SEA models on \filbench{} (\autoref{table:main_results_top10}).
To increase granularity, we compute the Fleiss' $\kappa$ for each sub-task.

\paragraph{Results.}
The results in \autoref{table:model-agreement} show that the SEA-specific group consistently demonstrate higher agreement on several sub-tasks than the non-SEA models.
This suggests that SEA-specific finetuning can improve model reliability and consistency in outputs.
However, both groups show alarming disagreement on cultural tasks, indicating fundamentally different interpretations of culturally-nuanced content.
We show some examples of model disagreement for the SEA-specific group in Appendix \ref{sec:appendix-disagreement}.
This implies that while regional specialization improves reliability, deeper cultural adaptation and more sophisticated training approaches may be needed to achieve reliable performance on Filipino.

\begin{table}[t]
    \centering
    \resizebox*{\linewidth}{!}{
        \begin{tabular}{lrr}
            \toprule
            \textbf{Sub-Task}   & \multicolumn{2}{c}{\bfseries Model Agreement (Fleiss' $\kappa$)}                 \\
                                & SEA-Specific                                                     & Top-Five      \\
            \midrule
            \multicolumn{3}{l}{\textit{Classical NLP (CN)}}                                                        \\
            Text Classification & \cold{0.513}                                                     & \cold{0.174}  \\
            Named-Entity Recog. & \cold{0.639}                                                     & \cold{0.273}  \\
            Sentiment Analysis  & \cold{0.598}                                                     & \cold{0.212}  \\
            \midrule
            \multicolumn{3}{l}{\textit{Cultural Knowledge (CK)}}                                                   \\
            Regional Knowledge  & \cold{0.393}                                                     & \cold{0.209}  \\
            Factual Knowledge   & \cold{0.224}                                                     & \cold{0.115}  \\
            Cultural Values     & \cold{0.403}                                                     & \cold{0.187}  \\
            Word-sense Disamb.  & \cold{0.072}                                                     & \cold{-0.041} \\
            \midrule
            \multicolumn{3}{l}{\textit{Reading Comprehension (RC)}}                                                \\
            Readability         & \cold{0.207}                                                     & \cold{-0.119} \\
            Reading Comp.       & \cold{0.377}                                                     & \cold{0.248}  \\
            NLI                 & \cold{0.438}                                                     & \cold{0.201}  \\
            \bottomrule
        \end{tabular}
    }
    \caption{
        \textbf{Inter-model agreement on MCF-based tasks.}
        Inter-model agreement, as measured by Fleiss' $\kappa$, for each sub-task in \filbench{}.
        Despite good performance on \filbench{}, models tend to disagree with one another, highlighting gaps in reliability.
        % The list of SEA-specific models evaluated can be found in \autoref{table:filbench_models}, whereas the Top-Five models are those with the highest \filbench{} scores in \autoref{table:main_results_top10}.
    }
    \label{table:model-agreement}
\end{table}

% \begin{figure*}[t]
%     \centering
%     \includegraphics[width=0.5\linewidth]{example-image} % Adjust width as needed
%     \caption{Failure modes in Generation tasks in \filbench{}.}
%     \label{fig:failure_modes}
% \end{figure*}

\begin{figure}[t]
    \centering
    \includegraphics[width=0.95\linewidth]{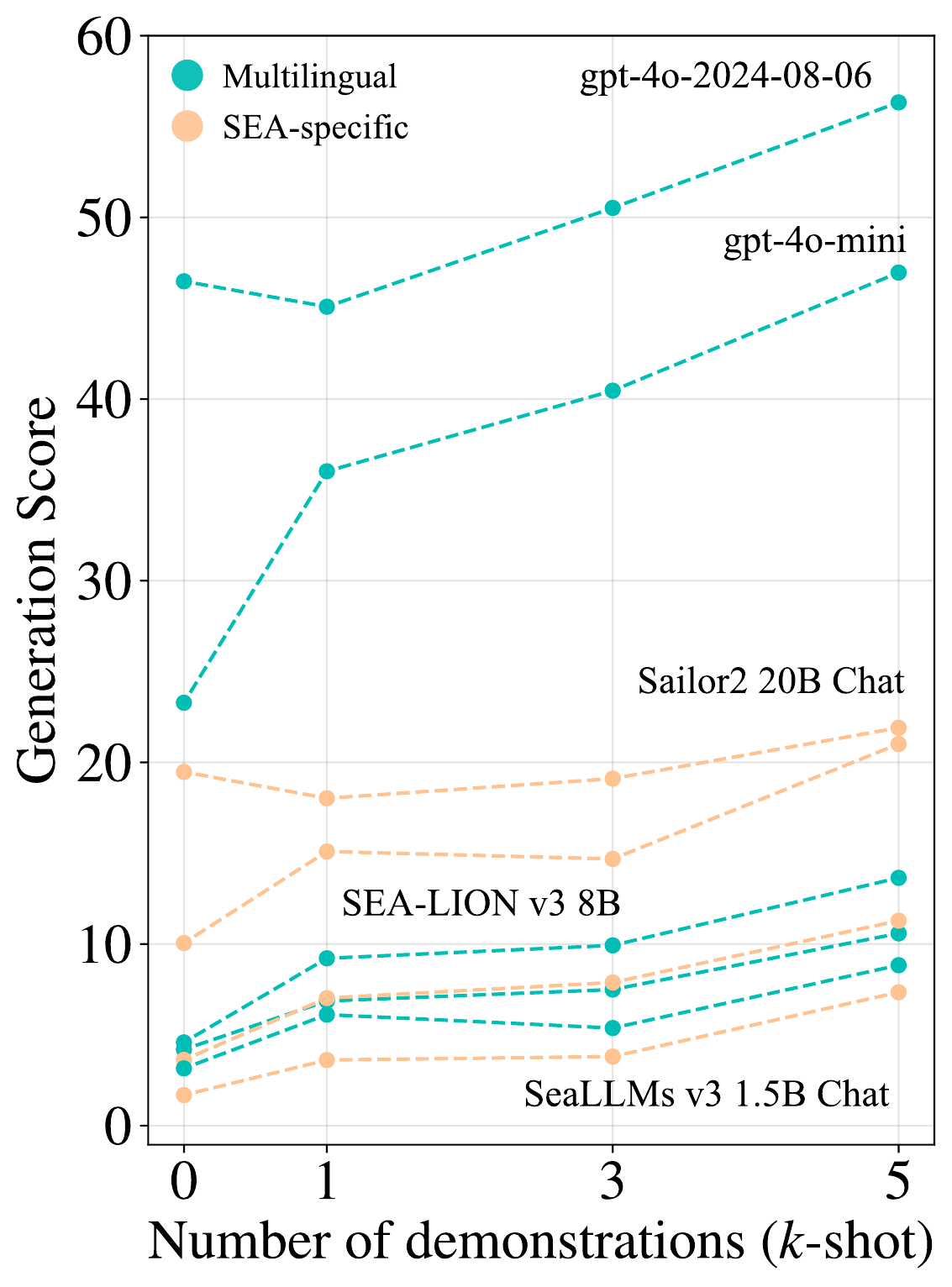} % Adjust width as needed
    \caption{
        \textbf{Effect of few-shot examples on Generation score.}
        Incorporating a single ($k=1$) demonstration during generation generally mitigates poor
        model performance.
    }
    \label{fig:fewshot}
\end{figure}

\subsection{Why do models fail in Generation tasks?}
\label{sec:analysis-failure}

\paragraph{Set-up.}
Given the low performance of most models in GN tasks, we qualitiatively analyze example outputs in order to provide a taxonomy of common failure cases in Filipino generative tasks.
In addition, we also assess whether we can mitigate low Generation performance by incorporating demonstrations or few-shot examples in the prompt.
To do so, we include $k=\{1,3,5\}$ examples on the prompt during evaluation.

\begin{table*}[ht]
    \centering
    \renewcommand{\arraystretch}{0.95}
    % \resizebox*{\linewidth}{!}{
    \small
    \begin{tabular}{p{0.2\linewidth}  p{0.80\linewidth}}
        \toprule
        \textbf{Failure mode} & \textbf{Example}                                                                                                      \\
        \midrule
        Poor instruction-     & ENG: Tom winked at Mary. FIL:                                                                                         \\
        following             &                                                                                                                       \\
                              & \textbf{Gold:} \textit{Si Tom ay kumindat kay Mary.}                                                                  \\
                              & \textbf{Pred:} \textit{Tom mutiara kepada Mary. (mutiara = to wink).}                                                 \\
                              & \textbf{Explanation:} Output is not in Filipino.                                                                      \\
        \midrule
        Overgeneration        & ENG: Did you hear the lecture in that class yesterday? asked the student. FIL:                                        \\
                              &                                                                                                                       \\
                              & \textbf{Gold:} \textit{Narinig mo ba ang lecture sa klaseng yan kahapon?, tanong ng estudyante.}                      \\
                              & \textbf{Pred:} \textit{ Sige, narinig ko na. ANG LECTURA YESTERDAY BA? Saan po ba ito naganap?} (continued...)        \\
                              & \textbf{Explanation:} Output is incoherent and generated highly verbose text.                                         \\
        \midrule
        Hallucination         & ENG: The more you read the book, the less you will understand it. FIL:                                                \\
                              &                                                                                                                       \\
                              & \textbf{Gold:} \textit{Nang mas na binabasa mo ang libro, menos mong maiintindihan.}                                  \\
                              & \textbf{Pred:} \textit{Ang higit na nagbabasa ng libro, mas nagiging delikado at hindi kayang tukuyin ang kahulugan.} \\
                              & \textbf{Explanation:} Output hallucinated the concept of danger (\textit{delikado}).                                  \\
        \bottomrule
    \end{tabular}
    % }
    \caption{
        \textbf{Common failure modes in translation tasks in the Generation category.}
        We find three common failure modes in most models in the Generation category of \filbench{}.
        The predictions in this table show the outputs of SEA-LION v3 70B, the currently best-performing SEA-specific model.
    }
    \label{table:generation-fail-modes-main}
\end{table*}

\paragraph{Results.}
We find common failure modes on Generation tasks in \filbench{} and show some examples from the outputs of the best-performing SEA-specific model, SEA-LION v3 70B, in \autoref{table:generation-fail-modes-main}:

\begin{itemize}[leftmargin=3mm]
    \item \textbf{Poor instruction-following.}
          When presented with Generation tasks from \filbench{}, models tend to misinterpret instructions or generate translations in an incorrect target language.
    \item \textbf{Overgeneration.}
          In the case of translation, models tend to produce overly verbose text than necessary, usually until the maximum generation length is reached.
          This usually results in incoherent text even if properly translated to the target language.
    \item \textbf{Hallucination.}
          Models often fail in Generation tasks due to spurious artifacts in the generated text.
          These tend to diminish the faithfulness of the model's output, especially in the case of translation tasks.
\end{itemize}

By manually inspecting a sample of 100 failure cases from GPT-4o, we find that overgeneration and poor instruction-following are the most dominant,  with 47\% and 34\% respectively, while hallucination occurs 19\% of the time.
We hypothesize that overgeneration can be caused by training data imbalance, as suggested in \cite{bawden-yvon-2023-investigating} and \cite{alves-etal-2023-steering} work in the case for BLOOM and LLaMA 7B.
We defer the ablation of training data quality and its effect on translation performance to future work.

In addition, we also find that \textbf{few-shot prompting can mitigate drop in Generation performance} (\autoref{fig:fewshot}).
We find that poor instruction-following, which is common especially in zero-shot \texttt{ENG} $\to$ \texttt{FIL} decreases once examples are provided. Full few-shot experiment results are shown in Appendix \ref{sec:few-shot}.
Despite these results, model performance on generation tasks remain generally poor, with frequent instances of overgeneration and semantically inaccurate translations. We further explain reasons for this using the Tatoeba dataset, which models consistently underperform on, in Appendix \ref{sec:generation-analysis}.

\subsection{Human evaluation of \filbench{}}
When curating test instances for \filbench{}, we ensured that the majority of sources in underwent human annotation and evaluation. 
However, we want to verify that strong agreement between native speakers and the gold answers persisted after the instances were converted into our task-specific formulations (Appendix \ref{sec:appendix-prompts}). 

\paragraph{Set-up.}
In order to evaluate the agreement between native speakers and \filbench{}'s gold answers, we sample 150 instances from \filbench{} with similar sub-task distribution.
Then, three authors (all native speakers of Filipino) served as annotators to label each instance.
For MCF tasks, the annotators choose the letter-option of the correct answer.
For GN tasks, we provide the annotators with a free-form text field to input their answers.
Then, we compute the inter-annotator agreement via Fleiss' $\kappa$ across two settings: (i) among annotators (\textit{intra-group}) and the (ii) majority response of human annotators to \filbench{}'s gold answer (\textit{inter-group}).
For GN, we compute the average ROUGE-L score for each annotator pair (\textit{intra-group}) and the average of the ROUGE-L score between the gold reference translation and each of the annotator translation (\textit{inter-group}).

\begin{table}[t]
\centering
\resizebox*{\linewidth}{!}{
\begin{tabular}{lrr}
\toprule
\textbf{Task Formulation}                   & \textbf{Intra-group} & \textbf{Inter-group} \\
\midrule
MCF, Fleiss' $\kappa$ &  0.8163           &    0.8756         \\
Generation, Avg. ROUGE-L   &  0.7604           &  0.7806          \\
\bottomrule
\end{tabular}
}
\caption{
    \textbf{Inter-rater agreement of native-speakers to a subset of \filbench{}.}
    We show that \filbench{} instances have a strong agreement with native speakers on both MFC-based (Cultural Knowledge, Classical NLP, Reading Comprehension) and Generation tasks.
}
\label{table:human-eval}
\end{table}

\paragraph{Results.}
\autoref{table:human-eval} shows the agreement scores among the three annotators (\textit{intra-group}) and their overall agreement with the gold reference answer (\textit{inter-group}).
The Fleiss' $\kappa$ indicate high agreement \citep{Landis1977TheMO}, suggesting that the instances in FilBench are reliable and aligns with native-speakers.
In addition, the ROUGE-L scores between annotators are also high, suggesting that the Generation instances can be reproducibly translated.
Furthermore, the inter-group ROUGE-L score supports this claim, as evidenced by similar performance given that most of the Generation instances were originally translated by other native speakers.
In general, the results suggest that the agreement between native speakers and the gold answers are preserved even after converting it into our task-specific formulations.

\section{Discussion}

\paragraph{On what to prioritize next when collecting data for Filipino-centric post-training.}
Our findings, through \filbench{},  reveal critical gaps in existing LLM's capabilities to process Filipino text, particularly in generation tasks where the best models achieved $\leq60\%$ performance.
In addition, we also find that continuous finetuning helps improve LLM performance on \filbench{}.
This suggests that post-training data collection efforts should prioritize high-quality translation pairs and generative content across diverse domains.
Furthermore, gathering training data from a wide range of Philippine languages, beyond just Tagalog, can enhance the performance of LLMs as demonstrated in \citet{buzaaba2025lugha}.
We posit that this can be achieved by taking advantage of cross-lingual transfer \citep{artetxe-etal-2020-cross} across typologically-similar languages.

\paragraph{On the importance of building language-community specific evaluation suites.}
Our findings strongly reinforce the necessity of developing language-community specific evaluation suites rather than relying on general multilingual benchmarks.
\filbench{} demonstrates that even state-of-the-art models like GPT-4o achieve only 75.56\% overall performance, indicating that Filipino presents unique challenges not captured in broader evaluations.
By creating focused evaluation suites like \filbench{}, the research community can more accurately identify model limitations and track progress in ways that respect the linguistic particularities of Philippine languages.
Furthermore, the performance variations across Filipino, Tagalog, and Cebuano emphasize the importance of fine-grained attention to linguistic diversity even within regions.
This points to the need for training approaches that recognize intra-regional linguistic boundaries rather than treating Southeast Asian languages as a homogeneous group.

%\paragraph{On ``what's next'' for FilBench and for Filipino NLP}
%\ljm{TODO}

\section{Related Work}
\paragraph{State of LLM Research for Philippine Languages.}
Progress in the NLP research landscape for Philippine languages such as Tagalog and Cebuano is seeing a promising growth, which can be attributed to democratization and access to LLM artifacts, particularly data and open models \citep{lovenia-etal-2024-seacrowd}.
The first works to release open-source artifacts include tasks such as sentiment analysis, hate speech detection, and natural language inference (NLI) \citep{cruz2019evaluating,cruz2020establishing,cruz-cheng-2022-improving}.
% The works of \citet{cruz2019evaluating}, \citet{cruz2020establishing}, and \citet{cruz-cheng-2022-improving} were the first to release open-source artifacts, including pre-trained Tagalog LLMs trained from Wikipedia, news articles, and social media data for baseline tasks such as sentiment analysis, hate speech detection, and natural language inference (NLI).
Further release of multilingual LLMs supporting Tagalog, allowed researchers to explore further linguistic phenomena from classical NLP tasks \citep[\textit{inter alia}]{pilar-etal-2023-cebuaner,mayhew-etal-2024-universal} to language model applications \citep{catapang-visperas-2023-emotion,montalan2024kalahi}.
% such as part-of-speech tagging, \cite{layacan-etal-2024-zero}, named-entity recognition \cite{pilar-etal-2023-cebuaner,mayhew-etal-2024-universal}, and code-switching \cite{zhang-etal-2023-multilingual,yong-etal-2023-prompting} as well as applications such as moral reasoning \cite{catapang-visperas-2023-emotion}, sexism and biases \cite{gamboa-lee-2025-filipino}, and cultural understanding \cite{montalan2024kalahi}.
% More recently, data-centric initiatives and collaborations such as SEACrowd \cite{lovenia-etal-2024-seacrowd} allowed the community to have better access to Philippine language data for LLM research through the release of its accessible one-stop-shop data hub\footnote{\url{https://seacrowd.github.io/seacrowd-catalogue/}}.

\paragraph{Language-specific LLM Evaluation Benchmarks.}
Global research communities are following the trend of releasing language-specific benchmarks in order to assess and track LLM progress in their respective languages.
Notable examples include AfroBench for African languages \citep{ojo2023good}, BenCzechMark for the Czech \citep{fajcik2024benczechmark}, the Open Arabic LLM Leaderboard for Arabic \citep{OALL-2} and Le Leaderboard for French \citep{openllm-French-leaderboard}.
These benchmarks usually contain curated tasks that may include translated versions of existing datasets or subsets of larger evaluation suites.
\filbench{} takes inspiration from these efforts by curating a comprehensive evaluation suite for Philippine languages.

Region-specific benchmarks also exist such as SeaBench and SeaExam \citep{liu2025seaexam} for Southeast Asia, although they do not contain any Filipino-specific subset.
The most recent effort related to \filbench{} is Batayan \cite{montalan2025batayan}, which is part of SEA-HELM \citep{susanto2025sea}.
\filbench{} takes a complementary approach by systematically curating existing benchmarks, enabling not only greater efficiency in resource utilization but also facilitating a wider diversity of task types and expanded coverage of Philippine languages beyond Filipino (in this case, Tagalog and Cebuano).
% This methodology allows us to incorporate established tasks with proven metrics while simultaneously representing the linguistic diversity of the Philippines.
% Moreover, our approach provides a replicable framework that researchers can efficiently adapt to other low-resource languages, as the process of identifying and consolidating existing resources requires significantly less time and fewer specialized personnel than creating new annotated datasets from scratch.
% \filbench{} focuses on evaluating tasks prioritized by Filipino NLP researchers, specifically in Tagalog, Filipino, and Cebuano.

\section{Conclusion}
In this work, we present a comprehensive evaluation of LLMs on Filipino-centric tasks to investigate their strengths and limitations, which still remains underexplored. We curate a benchmark called \filbench{} across four categories and \numsubtasks{} sub-tasks, based on our analyses of research priorities in Philippine NLP. Through \filbench{}, we discovered weaknesses in the current open and commercial state-of-the-art LLMs, such as low reliability and poor generation capabilities. \filbench{} emphasizes the value of creating language-specific LLM benchmarks, as it allows us to find promising avenues for models to improve their Filipino-centric performance. Specifically, this includes language-specific post-training and collecting relevant training datasets for text generation. We hope that \filbench{} aids in driving the progress in Filipino NLP.

\section*{Limitations}

\paragraph{Influence of training data on downstream \filbench{} performance.}
When selecting models for evaluation on \filbench{}, we categorized based on whether these models were originally presented as multilingual or SEA-specific, rather than considering the proportion of Filipino-centric training data used for fine-tuning.
The training data provenance is difficult to track, especially for closed-source models.
This explains why models that top other multilingual leaderboards such as Aya Expanse 32B \citep{dang2024aya} perform poorly on \filbench{}, because it was not explicitly trained on Filipino.
Our experiments hinge on the assumption that cross-lingual transfer \citep{artetxe-etal-2020-cross} happens during different states of language modeling, as evidenced in \citet{chirkova-nikoulina-2024-zero-shot}.
We leave the systematic exploration of the influence of the proportion of language-specific training data to a language-specific benchmark (i.e., Filipino-centric training data to \filbench{} performance) for future work.

\paragraph{Focus on Tagalog and Cebuano.}
While some of the datasets in \filbench{} support other Philippine languages (i.e. Ilokano for Belebele), data for these (labeled or otherwise) remain sparse.
We focus our suite on the relatively better-resourced Filipino and Cebuano, with the hope of supporting more languages once more datasets become available.
Future work might explore data augmentation techniques and other community-driven data collection initiatives to extend \filbench{}'s coverage to languages like Hiligaynon, Bikolano, and others.

\section*{Ethics Statement}
The development and evaluation of language technologies for Filipino, Cebuano, and other Philippine languages addresses important issues of linguistic inclusion and technological access. \filbench{} aims to support the development of more capable Filipino language technologies that can serve the significant population of Filipino speakers worldwide. The benchmark deliberately includes culturally-specific knowledge assessment to address known biases in LLMs toward Western values and content. This highlights the importance of evaluating models in their cultural context rather than assuming universal applicability. Datasets included in \filbench{} are from publicly accessible sources, and the authors obtained explicit approval from dataset creators when license information was unclear. Overall, we do not see any serious ethical issues with this work.

\section*{Acknowledgments}
The authors would like to thank Cohere Labs for providing credits through the Cohere Research Grant to run the Aya model series, and Together AI for additional computational credits for running several open models. 
We also acknowledge the Hugging Face team, particularly the OpenEvals team (Cl\'ementine Fourrier and Nathan Habib) and Daniel van Strien, for their support in publishing the \filbench{} blog post. 
Finally, we thank the reviewers from the May ARR cycle for their helpful feedback and insightful comments that improved this paper.

\bibliography{custom}

\begin{thebibliography}{85}
\providecommand{\natexlab}[1]{#1}

\bibitem[{Adelani et~al.(2024)Adelani, Liu, Shen, Vassilyev, Alabi, Mao, Gao, and Lee}]{adelani-etal-2024-sib}
David~Ifeoluwa Adelani, Hannah Liu, Xiaoyu Shen, Nikita Vassilyev, Jesujoba~O. Alabi, Yanke Mao, Haonan Gao, and En-Shiun~Annie Lee. 2024.
\newblock {SIB}-200: A simple, inclusive, and big evaluation dataset for topic classification in 200+ languages and dialects.
\newblock In \emph{Proceedings of the 18th Conference of the European Chapter of the Association for Computational Linguistics (Volume 1: Long Papers)}, pages 226--245.

\bibitem[{Aji et~al.(2023)Aji, Forde, Loo, Sutawika, Wang, Winata, Yong, Zhang, Do{\u{g}}ru{\"o}z, Tan, and Cruz}]{aji-etal-2023-current}
Alham~Fikri Aji, Jessica~Zosa Forde, Alyssa~Marie Loo, Lintang Sutawika, Skyler Wang, Genta~Indra Winata, Zheng-Xin Yong, Ruochen Zhang, A.~Seza Do{\u{g}}ru{\"o}z, Yin~Lin Tan, and Jan Christian~Blaise Cruz. 2023.
\newblock \href {https://doi.org/10.18653/v1/2023.ijcnlp-tutorials.2} {Current status of {NLP} in south {E}ast {A}sia with insights from multilingualism and language diversity}.
\newblock In \emph{Proceedings of the 13th International Joint Conference on Natural Language Processing and the 3rd Conference of the Asia-Pacific Chapter of the Association for Computational Linguistics: Tutorial Abstract}, pages 8--13, Nusa Dua, Bali. Association for Computational Linguistics.

\bibitem[{Alves et~al.(2023)Alves, Guerreiro, Alves, Pombal, Rei, de~Souza, Colombo, and Martins}]{alves-etal-2023-steering}
Duarte Alves, Nuno Guerreiro, Jo{\~a}o Alves, Jos{\'e} Pombal, Ricardo Rei, Jos{\'e} de~Souza, Pierre Colombo, and Andre Martins. 2023.
\newblock \href {https://doi.org/10.18653/v1/2023.findings-emnlp.744} {Steering large language models for machine translation with finetuning and in-context learning}.
\newblock In \emph{Findings of the Association for Computational Linguistics: EMNLP 2023}, pages 11127--11148, Singapore. Association for Computational Linguistics.

\bibitem[{Alves et~al.(2024)Alves, Pombal, Guerreiro, Martins, Alves, Farajian, Peters, Rei, Fernandes, Agrawal et~al.}]{alves2024tower}
Duarte~M Alves, Jos{\'e} Pombal, Nuno~M Guerreiro, Pedro~H Martins, Jo{\~a}o Alves, Amin Farajian, Ben Peters, Ricardo Rei, Patrick Fernandes, Sweta Agrawal, and 1 others. 2024.
\newblock Tower: An open multilingual large language model for translation-related tasks.
\newblock \emph{arXiv preprint arXiv:2402.17733}.

\bibitem[{Anastasopoulos et~al.(2020)Anastasopoulos, Cattelan, Dou, Federico, Federmann, Genzel, Guzm{\'a}n, Hu, Hughes, Koehn, Lazar, Lewis, Neubig, Niu, {\"O}ktem, Paquin, Tang, and Tur}]{anastasopoulos-etal-2020-tico}
Antonios Anastasopoulos, Alessandro Cattelan, Zi-Yi Dou, Marcello Federico, Christian Federmann, Dmitriy Genzel, Franscisco Guzm{\'a}n, Junjie Hu, Macduff Hughes, Philipp Koehn, Rosie Lazar, Will Lewis, Graham Neubig, Mengmeng Niu, Alp {\"O}ktem, Eric Paquin, Grace Tang, and Sylwia Tur. 2020.
\newblock {TICO}-19: the translation initiative for {CO}vid-19.
\newblock In \emph{Proceedings of the 1st Workshop on {NLP} for {COVID}-19 (Part 2) at {EMNLP} 2020}.

\bibitem[{Artetxe et~al.(2020)Artetxe, Ruder, and Yogatama}]{artetxe-etal-2020-cross}
Mikel Artetxe, Sebastian Ruder, and Dani Yogatama. 2020.
\newblock \href {https://doi.org/10.18653/v1/2020.acl-main.421} {On the cross-lingual transferability of monolingual representations}.
\newblock In \emph{Proceedings of the 58th Annual Meeting of the Association for Computational Linguistics}, pages 4623--4637, Online. Association for Computational Linguistics.

\bibitem[{Ashok and Lipton(2023)}]{ashok2023promptner}
Dhananjay Ashok and Zachary~C Lipton. 2023.
\newblock Promptner: Prompting for named entity recognition.
\newblock \emph{arXiv preprint arXiv:2305.15444}.

\bibitem[{Bacalla(2019)}]{bacalla2019morpoanalisis}
Lita Bacalla. 2019.
\newblock \href {https://doi.org/10.5861/ijrse.2019.4902} {Morpo-analisis ng wikang tagalog at wikang sugbuanun'g binisaya: Pahambing na pag-aaral}.
\newblock \emph{International Journal of Resarch Studies in Education}, 8:55--65.

\bibitem[{Baliber et~al.(2020)Baliber, Cheng, Adlaon, and Mamonong}]{baliber-etal-2020-bridging}
Renz~Iver Baliber, Charibeth Cheng, Kristine~Mae Adlaon, and Virgion Mamonong. 2020.
\newblock \href {https://doi.org/10.18653/v1/2020.loresmt-1.2} {Bridging {P}hilippine languages with multilingual neural machine translation}.
\newblock In \emph{Proceedings of the 3rd Workshop on Technologies for MT of Low Resource Languages}, pages 14--22, Suzhou, China. Association for Computational Linguistics.

\bibitem[{Bandarkar et~al.(2024)Bandarkar, Liang, Muller, Artetxe, Shukla, Husa, Goyal, Krishnan, Zettlemoyer, and Khabsa}]{bandarkar-etal-2024-belebele}
Lucas Bandarkar, Davis Liang, Benjamin Muller, Mikel Artetxe, Satya~Narayan Shukla, Donald Husa, Naman Goyal, Abhinandan Krishnan, Luke Zettlemoyer, and Madian Khabsa. 2024.
\newblock {The Belebele Benchmark: a Parallel Reading Comprehension Dataset in 122 Language Variants}.
\newblock In \emph{Proceedings of the 62nd Annual Meeting of the Association for Computational Linguistics (Volume 1: Long Papers)}, pages 749--775.

\bibitem[{Bardaj\'{i} et~al.(2024)Bardaj\'{i}, Or, Aquino, and Himmelmann}]{bardaji2024}
Maria Bardaj\'{i}, Elsie Or, Angelina Aquino, and Nikolaus Himmelmann. 2024.
\newblock The challenges of symmetrical voice languages for universal dependencies.
\newblock In \emph{Proceedings of the 15th International Conference of the Association for Linguistic Typology}.

\bibitem[{Bawden and Yvon(2023)}]{bawden-yvon-2023-investigating}
Rachel Bawden and Fran{\c{c}}ois Yvon. 2023.
\newblock \href {https://aclanthology.org/2023.eamt-1.16/} {Investigating the translation performance of a large multilingual language model: the case of {BLOOM}}.
\newblock In \emph{Proceedings of the 24th Annual Conference of the European Association for Machine Translation}, pages 157--170.

\bibitem[{Buzaaba et~al.(2025)Buzaaba, Wettig, Adelani, and Fellbaum}]{buzaaba2025lugha}
Happy Buzaaba, Alexander Wettig, David~Ifeoluwa Adelani, and Christiane Fellbaum. 2025.
\newblock Lugha-llama: Adapting large language models for african languages.
\newblock \emph{arXiv preprint arXiv:2504.06536}.

\bibitem[{Buñag and Esquivel(2023)}]{bunag2023news}
Kenrick~Lance Buñag and Rosanna Esquivel. 2023.
\newblock Transformer-based conditional language models to generate filipino news article.
\newblock In \emph{Proceedings of the International Conference on International Engineering and Operations Management}.

\bibitem[{Cahyawijaya et~al.(2024)Cahyawijaya, Zhang, Lovenia, Cruz, Gilbert, Nomoto, and Aji}]{cahyawijaya2024thank}
Samuel Cahyawijaya, Ruochen Zhang, Holy Lovenia, Jan Christian~Blaise Cruz, Elisa Gilbert, Hiroki Nomoto, and Alham~Fikri Aji. 2024.
\newblock Thank you, stingray: Multilingual large language models can not (yet) disambiguate cross-lingual word sense.
\newblock \emph{arXiv preprint arXiv:2410.21573}.

\bibitem[{Cao et~al.(2023)Cao, Zhou, Lee, Cabello, Chen, and Hershcovich}]{cao-etal-2023-assessing}
Yong Cao, Li~Zhou, Seolhwa Lee, Laura Cabello, Min Chen, and Daniel Hershcovich. 2023.
\newblock \href {https://doi.org/10.18653/v1/2023.c3nlp-1.7} {Assessing cross-cultural alignment between {C}hat{GPT} and human societies: An empirical study}.
\newblock In \emph{Proceedings of the First Workshop on Cross-Cultural Considerations in NLP (C3NLP)}, pages 53--67.

\bibitem[{Catapang and Visperas(2023)}]{catapang-visperas-2023-emotion}
Jasper~Kyle Catapang and Moses Visperas. 2023.
\newblock \href {https://aclanthology.org/2023.nlp4dh-1.1/} {Emotion-based morality in {T}agalog and {E}nglish scenarios ({EM}o{TES}-3{K}): A parallel corpus for explaining (im)morality of actions}.
\newblock In \emph{Proceedings of the Joint 3rd International Conference on Natural Language Processing for Digital Humanities and 8th International Workshop on Computational Linguistics for Uralic Languages}, pages 1--6, Tokyo, Japan. Association for Computational Linguistics.

\bibitem[{Chirkova and Nikoulina(2024)}]{chirkova-nikoulina-2024-zero-shot}
Nadezhda Chirkova and Vassilina Nikoulina. 2024.
\newblock \href {https://aclanthology.org/2024.inlg-main.53/} {Zero-shot cross-lingual transfer in instruction tuning of large language models}.
\newblock In \emph{Proceedings of the 17th International Natural Language Generation Conference}, pages 695--708, Tokyo, Japan. Association for Computational Linguistics.

\bibitem[{Cosme and De~Leon(2023)}]{cosme-deleon-2024-products}
Camilla~Johnine Cosme and Marlene De~Leon. 2023.
\newblock Sentiment analysis of code-switched filipino-english product and service reviews using transformers-based large language models.
\newblock In \emph{Proceedings of World Conference on Information Systems for Business Management}, pages 123--135.

\bibitem[{Cruz and Cheng(2019)}]{cruz2019evaluating}
Jan Christian~Blaise Cruz and Charibeth Cheng. 2019.
\newblock \href {https://arxiv.org/abs/1907.00409} {Evaluating language model finetuning techniques for low-resource languages}.
\newblock \emph{arXiv preprint arXiv:1907.00409}.

\bibitem[{Cruz and Cheng(2020)}]{cruz2020establishing}
Jan Christian~Blaise Cruz and Charibeth Cheng. 2020.
\newblock \href {https://arxiv.org/abs/2005.02068} {Establishing baselines for text classification in low-resource languages}.
\newblock \emph{arXiv preprint arXiv:2005.02068}.

\bibitem[{Cruz and Cheng(2022)}]{cruz-cheng-2022-improving}
Jan Christian~Blaise Cruz and Charibeth Cheng. 2022.
\newblock \href {https://aclanthology.org/2022.lrec-1.703/} {Improving large-scale language models and resources for {F}ilipino}.
\newblock In \emph{Proceedings of the Thirteenth Language Resources and Evaluation Conference}, pages 6548--6555, Marseille, France. European Language Resources Association.

\bibitem[{Cruz et~al.(2021)Cruz, Resabal, Lin, Velasco, and Cheng}]{cruz-etal-2021-news}
Jan Christian~Blaise Cruz, Jose~Kristian Resabal, James Lin, Dan~John Velasco, and Charibeth Cheng. 2021.
\newblock Exploiting news article structure for automatic corpus generation of entailment datasets.
\newblock In \emph{{PRICAI} 2021: {T}rends in {A}rtificial {I}ntelligence}, pages 86--99.

\bibitem[{Cucio and Hennig(2025)}]{cucio2025artificial}
Micholo Cucio and Tristan Hennig. 2025.
\newblock {Artificial Intelligence and the Philippine Labor Market: Mapping Occupational Exposure and Complementarity}.
\newblock Technical report, International Monetary Fund.

\bibitem[{Dammu et~al.(2024)Dammu, Jung, Singh, Choudhury, and Mitra}]{dammu-etal-2024-uncultured}
Preetam Prabhu~Srikar Dammu, Hayoung Jung, Anjali Singh, Monojit Choudhury, and Tanu Mitra. 2024.
\newblock \href {https://doi.org/10.18653/v1/2024.emnlp-main.1134} {{\textquotedblleft}they are uncultured{\textquotedblright}: Unveiling covert harms and social threats in {LLM} generated conversations}.
\newblock pages 20339--20369.

\bibitem[{Dang et~al.(2024)Dang, Singh, D'souza, Ahmadian, Salamanca, Smith, Peppin, Hong, Govindassamy, Zhao et~al.}]{dang2024aya}
John Dang, Shivalika Singh, Daniel D'souza, Arash Ahmadian, Alejandro Salamanca, Madeline Smith, Aidan Peppin, Sungjin Hong, Manoj Govindassamy, Terrence Zhao, and 1 others. 2024.
\newblock Aya expanse: Combining research breakthroughs for a new multilingual frontier.
\newblock \emph{arXiv preprint arXiv:2412.04261}.

\bibitem[{Dou et~al.(2025)Dou, Liu, Zhou, Chen, Wang, Jin, Liu, Zhu, Du, Yang et~al.}]{dou2025sailor2}
Longxu Dou, Qian Liu, Fan Zhou, Changyu Chen, Zili Wang, Ziqi Jin, Zichen Liu, Tongyao Zhu, Cunxiao Du, Penghui Yang, and 1 others. 2025.
\newblock {Sailor2: Sailing in South-East Asia with Inclusive Multilingual LLMs}.
\newblock \emph{arXiv preprint arXiv:2502.12982}.

\bibitem[{Eberhard et~al.(2024)Eberhard, Simons, and Fennig}]{ethnologue}
David~M. Eberhard, Gary~F. Simons, and Charles~D. Fennig, editors. 2024.
\newblock \href {http://www.ethnologue.com} {\emph{Ethnologue: {L}anguages of the {W}orld}}, 27 edition.
\newblock SIL International, Dallas, Texas.

\bibitem[{El~Filali et~al.(2025)El~Filali, ALOUI, Husaain, Alzubaidi, Boussaha, Cojocaru, Fourrier, Habib, and Hacid}]{OALL-2}
Ali El~Filali, Manel ALOUI, Tarique Husaain, Ahmed Alzubaidi, Basma El~Amel Boussaha, Ruxandra Cojocaru, Clémentine Fourrier, Nathan Habib, and Hakim Hacid. 2025.
\newblock Open arabic llm leaderboard 2.
\newblock https://huggingface.co/spaces/OALL/Open-Arabic-LLM-Leaderboard.

\bibitem[{Eronen et~al.(2023)Eronen, Ptaszynski, and Masui}]{ERONEN2023103250}
Juuso Eronen, Michal Ptaszynski, and Fumito Masui. 2023.
\newblock Zero-shot cross-lingual transfer language selection using linguistic similarity.
\newblock \emph{Information Processing and Management}, 60(3):103250.

\bibitem[{Fajcik et~al.(2024)Fajcik, Docekal, Dolezal, Ondrej, Bene{\v{s}}, Kapsa, Smrz, Polok, Hradis, Neverilova et~al.}]{fajcik2024benczechmark}
Martin Fajcik, Martin Docekal, Jan Dolezal, Karel Ondrej, Karel Bene{\v{s}}, Jan Kapsa, Pavel Smrz, Alexander Polok, Michal Hradis, Zuzana Neverilova, and 1 others. 2024.
\newblock Benczechmark: A czech-centric multitask and multimetric benchmark for large language models with duel scoring mechanism.
\newblock \emph{arXiv preprint arXiv:2412.17933}.

\bibitem[{Federmann et~al.(2022)Federmann, Kocmi, and Xin}]{federmann-etal-2022-ntrex}
Christian Federmann, Tom Kocmi, and Ying Xin. 2022.
\newblock {NTREX}-128 {--} news test references for (mt) evaluation of 128 languages.
\newblock In \emph{Proceedings of the First Workshop on Scaling Up Multilingual Evaluation}, pages 21--24.

\bibitem[{Fourrier et~al.(2024)Fourrier, Habib, Lozovskaya, Szafer, and Wolf}]{open-llm-leaderboard-v2}
Clémentine Fourrier, Nathan Habib, Alina Lozovskaya, Konrad Szafer, and Thomas Wolf. 2024.
\newblock Open llm leaderboard v2.
\newblock \url{https://huggingface.co/spaces/open-llm-leaderboard/open_llm_leaderboard}.

\bibitem[{Grattafiori et~al.(2024)Grattafiori, Dubey, Jauhri, Pandey, Kadian, Al-Dahle, Letman, Mathur, Schelten, Vaughan et~al.}]{grattafiori2024llama}
Aaron Grattafiori, Abhimanyu Dubey, Abhinav Jauhri, Abhinav Pandey, Abhishek Kadian, Ahmad Al-Dahle, Aiesha Letman, Akhil Mathur, Alan Schelten, Alex Vaughan, and 1 others. 2024.
\newblock {The LLaMa 3 herd of models}.
\newblock \emph{arXiv preprint arXiv:2407.21783}.

\bibitem[{Group(2024)}]{bcg2024consumers}
Boston~Consulting Group. 2024.
\newblock \href {https://www.bcg.com/publications/2024/consumers-know-more-about-ai-than-businesses-think} {Consumers know more about ai than businesses think}.

\bibitem[{Gu et~al.(2024)Gu, Tafjord, Kuehl, Haddad, Dodge, and Hajishirzi}]{gu2024olmes}
Yuling Gu, Oyvind Tafjord, Bailey Kuehl, Dany Haddad, Jesse Dodge, and Hannaneh Hajishirzi. 2024.
\newblock {OLMES}: A standard for language model evaluations.
\newblock \emph{arXiv preprint arXiv:2406.08446}.

\bibitem[{Gururaja et~al.(2023)Gururaja, Bertsch, Na, Widder, and Strubell}]{gururaja-etal-2023-build}
Sireesh Gururaja, Amanda Bertsch, Clara Na, David Widder, and Emma Strubell. 2023.
\newblock \href {https://doi.org/10.18653/v1/2023.emnlp-main.822} {To build our future, we must know our past: Contextualizing paradigm shifts in natural language processing}.
\newblock In \emph{Proceedings of the 2023 Conference on Empirical Methods in Natural Language Processing}, pages 13310--13325, Singapore. Association for Computational Linguistics.

\bibitem[{Habib et~al.(2023)Habib, Fourrier, Kydlíček, Wolf, and Tunstall}]{lighteval}
Nathan Habib, Clémentine Fourrier, Hynek Kydlíček, Thomas Wolf, and Lewis Tunstall. 2023.
\newblock \href {https://github.com/huggingface/lighteval} {Lighteval: A lightweight framework for llm evaluation}.

\bibitem[{He et~al.(2024)He, Liang, Jiao, Zhang, Yang, Wang, Tu, Shi, and Wang}]{he-etal-2024-exploring}
Zhiwei He, Tian Liang, Wenxiang Jiao, Zhuosheng Zhang, Yujiu Yang, Rui Wang, Zhaopeng Tu, Shuming Shi, and Xing Wang. 2024.
\newblock \href {https://doi.org/10.1162/tacl_a_00642} {Exploring human-like translation strategy with large language models}.
\newblock \emph{Transactions of the Association for Computational Linguistics}, 12:229--246.

\bibitem[{Huang et~al.(2025)Huang, Zhu, Hu, He, Li, Huang, and Yuan}]{huang2025benchmax}
Xu~Huang, Wenhao Zhu, Hanxu Hu, Conghui He, Lei Li, Shujian Huang, and Fei Yuan. 2025.
\newblock Benchmax: A comprehensive multilingual evaluation suite for large language models.
\newblock \emph{arXiv preprint arXiv:2502.07346}.

\bibitem[{Hurst et~al.(2024)Hurst, Lerer, Goucher, Perelman, Ramesh, Clark, Ostrow, Welihinda, Hayes, Radford et~al.}]{hurst2024gpt}
Aaron Hurst, Adam Lerer, Adam~P Goucher, Adam Perelman, Aditya Ramesh, Aidan Clark, AJ~Ostrow, Akila Welihinda, Alan Hayes, Alec Radford, and 1 others. 2024.
\newblock {GPT-4o System Card}.
\newblock \emph{arXiv preprint arXiv:2410.21276}.

\bibitem[{Imperial and Kochmar(2023)}]{imperial-kochmar-2023-automatic}
Joseph~Marvin Imperial and Ekaterina Kochmar. 2023.
\newblock \href {https://doi.org/10.18653/v1/2023.findings-acl.331} {Automatic readability assessment for closely related languages}.
\newblock In \emph{Findings of the Association for Computational Linguistics: ACL 2023}, pages 5371--5386, Toronto, Canada. Association for Computational Linguistics.

\bibitem[{Imperial et~al.(2022)Imperial, Reyes, Ibañez, Sapinit, and Hussien}]{imperial-etal-2022-baseline}
Joseph~Marvin Imperial, Lloyd Lois~Antonie Reyes, Michael~Antoinio Ibañez, Ranz Sapinit, and Mohammed Hussien. 2022.
\newblock A baseline readability model for cebuano.
\newblock In \emph{Proceedings of the 17th Workshop on Innovative Use of NLP for Building Educational Applications}.

\bibitem[{Jiang et~al.(2024)Jiang, Sablayrolles, Roux, Mensch, Savary, Bamford, Chaplot, Casas, Hanna, Bressand et~al.}]{jiang2024mixtral}
Albert~Q Jiang, Alexandre Sablayrolles, Antoine Roux, Arthur Mensch, Blanche Savary, Chris Bamford, Devendra~Singh Chaplot, Diego de~las Casas, Emma~Bou Hanna, Florian Bressand, and 1 others. 2024.
\newblock Mixtral of experts.
\newblock \emph{arXiv preprint arXiv:2401.04088}.

\bibitem[{Kwon et~al.(2023)Kwon, Li, Zhuang, Sheng, Zheng, Yu, Gonzalez, Zhang, and Stoica}]{kwon2023efficient}
Woosuk Kwon, Zhuohan Li, Siyuan Zhuang, Ying Sheng, Lianmin Zheng, Cody~Hao Yu, Joseph Gonzalez, Hao Zhang, and Ion Stoica. 2023.
\newblock Efficient memory management for large language model serving with pagedattention.
\newblock In \emph{Proceedings of the 29th Symposium on Operating Systems Principles}, pages 611--626.

\bibitem[{Landis and Koch(1977)}]{Landis1977TheMO}
J~Richard Landis and Gary~G. Koch. 1977.
\newblock \href {https://api.semanticscholar.org/CorpusID:11077516} {The measurement of observer agreement for categorical data.}
\newblock \emph{Biometrics}, 33 1:159--74.

\bibitem[{Lin(2004)}]{lin-2004-rouge}
Chin-Yew Lin. 2004.
\newblock \href {https://aclanthology.org/W04-1013/} {{ROUGE}: A package for automatic evaluation of summaries}.
\newblock In \emph{Text Summarization Branches Out}, pages 74--81, Barcelona, Spain. Association for Computational Linguistics.

\bibitem[{Liu et~al.(2025)Liu, Zhang, Ying, Aljunied, Luu, and Bing}]{liu2025seaexam}
Chaoqun Liu, Wenxuan Zhang, Jiahao Ying, Mahani Aljunied, Anh~Tuan Luu, and Lidong Bing. 2025.
\newblock Seaexam and seabench: Benchmarking llms with local multilingual questions in southeast asia.
\newblock \emph{arXiv preprint arXiv:2502.06298}.

\bibitem[{Liu and Wang(2024)}]{liu2024earth}
Yan Liu and He~Wang. 2024.
\newblock \emph{{Who on Earth Is Using Generative AI?}}
\newblock World Bank.

\bibitem[{Livelo and Cheng(2018)}]{livelo-cheng-2018-dengue}
Evan~Dennison Livelo and Charibeth Cheng. 2018.
\newblock Intelligent dengue infoveillance using gated recurrent neural learning and cross-label frequencies.
\newblock In \emph{2018 IEEE International Conference on Agents}.

\bibitem[{Lovenia et~al.(2024)Lovenia, Mahendra, Akbar, Miranda, Santoso, Aco, Fadhilah, Mansurov, Imperial, Kampman, Moniz, Habibi, Hudi, Montalan, Hadiwijaya, Lopo, Nixon, Karlsson, Jaya, Diandaru, Gao, Irawan, Wang, Cruz, Whitehouse, Parmonangan, Khelli, Zhang, Susanto, Ryanda, Hermawan, Velasco, Kautsar, Hendria, Moslem, Flynn, Adilazuarda, Li, Lee, Damanhuri, Sun, Qorib, Djanibekov, Leong, Do, Muennighoff, Pansuwan, Putra, Xu, Chia, Purwarianti, Ruder, Tjhi, Limkonchotiwat, Aji, Keh, Winata, Zhang, Koto, Yong, and Cahyawijaya}]{lovenia-etal-2024-seacrowd}
Holy Lovenia, Rahmad Mahendra, Salsabil~Maulana Akbar, Lester James~Validad Miranda, Jennifer Santoso, Elyanah Aco, Akhdan Fadhilah, Jonibek Mansurov, Joseph~Marvin Imperial, Onno~P. Kampman, Joel Ruben~Antony Moniz, Muhammad Ravi~Shulthan Habibi, Frederikus Hudi, Railey Montalan, Ryan~Ignatius Hadiwijaya, Joanito~Agili Lopo, William Nixon, B{\"o}rje~F. Karlsson, James Jaya, and 42 others. 2024.
\newblock \href {https://doi.org/10.18653/v1/2024.emnlp-main.296} {{SEAC}rowd: A multilingual multimodal data hub and benchmark suite for {S}outheast {A}sian languages}.
\newblock In \emph{Proceedings of the 2024 Conference on Empirical Methods in Natural Language Processing}, pages 5155--5203, Miami, Florida, USA. Association for Computational Linguistics.

\bibitem[{Mayhew et~al.(2024)Mayhew, Blevins, Liu, Suppa, Gonen, Imperial, Karlsson, Lin, Ljube{\v{s}}i{\'c}, Miranda, Plank, Riabi, and Pinter}]{mayhew-etal-2024-universal}
Stephen Mayhew, Terra Blevins, Shuheng Liu, Marek Suppa, Hila Gonen, Joseph~Marvin Imperial, B{\"o}rje Karlsson, Peiqin Lin, Nikola Ljube{\v{s}}i{\'c}, Lester~James Miranda, Barbara Plank, Arij Riabi, and Yuval Pinter. 2024.
\newblock Universal {NER}: A gold-standard multilingual named entity recognition benchmark.
\newblock In \emph{Proceedings of the 2024 Conference of the North American Chapter of the Association for Computational Linguistics: Human Language Technologies (Volume 1: Long Papers)}, pages 4322--4337.

\bibitem[{McFarland(2008)}]{mcfarland2008linguistic}
Curtis~D McFarland. 2008.
\newblock Linguistic diversity and english in the philippines.
\newblock \emph{Philippine English: Linguistic and literary perspectives}, 1:131.

\bibitem[{{Meta AI}(2025)}]{metaai2025llama4blog}
{Meta AI}. 2025.
\newblock {The Llama 4 herd: The beginning of a new era of natively multimodal AI innovation}.
\newblock \url{https://ai.meta.com/blog/llama-4-multimodal-intelligence/}.
\newblock Blog post, accessed May 16, 2025.

\bibitem[{Metila et~al.(2016)Metila, Pradilla, and Williams}]{metila2016challenge}
Romylyn~A Metila, Lea Angela~S Pradilla, and Alan~B Williams. 2016.
\newblock The challenge of implementing mother tongue education in linguistically diverse contexts: The case of the philippines.
\newblock \emph{The Asia-Pacific Education Researcher}, 25:781--789.

\bibitem[{Miranda(2023)}]{miranda-2023-developing}
Lester~James Miranda. 2023.
\newblock \href {https://doi.org/10.18653/v1/2023.sealp-1.2} {Developing a named entity recognition dataset for {T}agalog}.
\newblock In \emph{Proceedings of the First Workshop in South East Asian Language Processing}, pages 13--20, Nusa Dua, Bali, Indonesia. Association for Computational Linguistics.

\bibitem[{{Mistral AI}(2024)}]{mistralai2024ministrauxblog}
{Mistral AI}. 2024.
\newblock {Mixtral of experts}.
\newblock \url{https://mistral.ai/news/ministraux}.
\newblock Blog post, accessed May 16, 2025.

\bibitem[{Mohamad~Alhajar(2024)}]{openllm-French-leaderboard}
Alexandre~Lavallée Mohamad~Alhajar. 2024.
\newblock Open llm french leaderboard v0.2.
\newblock \url{https://huggingface.co/spaces/le-leadboard/OpenLLMFrenchLeaderboard}.

\bibitem[{Montalan et~al.(2025)Montalan, Layacan, Africa, Flores, Lopez~II, Magsajo, Cayabyab, and Tjhi}]{montalan2025batayan}
Jann~Railey Montalan, Jimson~Paulo Layacan, David~Demitri Africa, Richell~Isaiah Flores, Michael~T Lopez~II, Theresa~Denise Magsajo, Anjanette Cayabyab, and William~Chandra Tjhi. 2025.
\newblock Batayan: A filipino nlp benchmark for evaluating large language models.
\newblock \emph{arXiv preprint arXiv:2502.14911}.

\bibitem[{Montalan et~al.(2024)Montalan, Ngui, Leong, Susanto, Rengarajan, Aji, and Tjhi}]{montalan2024kalahi}
Jann~Railey Montalan, Jian~Gang Ngui, Wei~Qi Leong, Yosephine Susanto, Hamsawardhini Rengarajan, Alham~Fikri Aji, and William~Chandra Tjhi. 2024.
\newblock Kalahi: A handcrafted, grassroots cultural {LLM} evalutation suite for filipino.
\newblock \emph{arXiv preprint arXiv:2409.15380}.

\bibitem[{Ng et~al.(2025)Ng, Nguyen, Huang, Tai, Leong, Leong, Yong, Ngui, Susanto, Cheng, Rengarajan, Limkonchotiwat, Hulagadri, Teng, Tong, Siow, Teo, Lau, Tan, Ong, Ong, Montalan, Chan, Antonyrex, Lee, Choa, Tat-Wee, Liu, Tjhi, Cambria, and Teo}]{sealionv3}
Raymond Ng, Thanh~Ngan Nguyen, Yuli Huang, Ngee~Chia Tai, Wai~Yi Leong, Wei~Qi Leong, Xianbin Yong, Jian~Gang Ngui, Yosephine Susanto, Nicholas Cheng, Hamsawardhini Rengarajan, Peerat Limkonchotiwat, Adithya~Venkatadri Hulagadri, Kok~Wai Teng, Yeo~Yeow Tong, Bryan Siow, Wei~Yi Teo, Wayne Lau, Choon~Meng Tan, and 12 others. 2025.
\newblock \href {https://arxiv.org/abs/2504.05747} {Sea-lion: Southeast asian languages in one network}.
\newblock \emph{Preprint}, arXiv:2504.05747.

\bibitem[{Oco and Roxas(2018)}]{oco-roxas-2018-survey}
Nathaniel Oco and Rachel Roxas. 2018.
\newblock \href {https://aclanthology.org/W18-2204/} {A survey of machine translation work in the {P}hilippines: From 1998 to 2018}.
\newblock In \emph{Proceedings of the {AMTA} 2018 Workshop on Technologies for {MT} of Low Resource Languages ({L}o{R}es{MT} 2018)}, pages 30--36, Boston, MA. Association for Machine Translation in the Americas.

\bibitem[{Ojo et~al.(2023)Ojo, Ogueji, Stenetorp, and Adelani}]{ojo2023good}
Jessica Ojo, Kelechi Ogueji, Pontus Stenetorp, and David~Ifeoluwa Adelani. 2023.
\newblock How good are large language models on african languages?
\newblock \emph{arXiv preprint arXiv:2311.07978}.

\bibitem[{Papineni et~al.(2002)Papineni, Roukos, Ward, and Zhu}]{papineni-etal-2002-bleu}
Kishore Papineni, Salim Roukos, Todd Ward, and Wei-Jing Zhu. 2002.
\newblock \href {https://doi.org/10.3115/1073083.1073135} {{B}leu: a method for automatic evaluation of machine translation}.
\newblock In \emph{Proceedings of the 40th Annual Meeting of the Association for Computational Linguistics}, pages 311--318, Philadelphia, Pennsylvania, USA. Association for Computational Linguistics.

\bibitem[{{Philippine Statistics Authority}(2020)}]{PSA2020Household}
{Philippine Statistics Authority}. 2020.
\newblock \href {https://psa.gov.ph/content/household-population-number-households-and-average-household-size-philippines-2020-census} {Household population, number of households and average household size of the philippines (2020 census of population and housing)}.
\newblock Accessed: 2025-04-03.

\bibitem[{Philippy et~al.(2023)Philippy, Guo, and Haddadan}]{philippy-etal-2023-towards}
Fred Philippy, Siwen Guo, and Shohreh Haddadan. 2023.
\newblock \href {https://doi.org/10.18653/v1/2023.acl-long.323} {Towards a common understanding of contributing factors for cross-lingual transfer in multilingual language models: A review}.
\newblock In \emph{Proceedings of the 61st Annual Meeting of the Association for Computational Linguistics (Volume 1: Long Papers)}, pages 5877--5891.

\bibitem[{Pilar et~al.(2023)Pilar, Dedoroy, Papas, Buenaventura, Montefalcon, Padilla, Imperial, Abisado, and Maceda}]{pilar-etal-2023-cebuaner}
Ma. Beatrice~Emanuela Pilar, Dane Dedoroy, Ellyza~Mari Papas, Mary~Loise Buenaventura, Myron~Darrel Montefalcon, Jay~Rhald Padilla, Joseph~Marvin Imperial, Mideth Abisado, and Lany Maceda. 2023.
\newblock {C}ebua{NER}: A new baseline {C}ebuano named entity recognition model.
\newblock In \emph{Proceedings of the 37th Pacific Asia Conference on Language, Information and Computation}, pages 792--800.

\bibitem[{Qiu et~al.(2025)Qiu, Fabbri, Agarwal, Huang, Tan, Peng, and Wu}]{qiu-etal-2025-evaluating}
Haoyi Qiu, Alexander~R. Fabbri, Divyanish Agarwal, Kung-Hsiang Huang, Sarah Tan, Nanyun Peng, and Chien-Sheng Wu. 2025.
\newblock Evaluating cultural and social awareness of llm web agents.
\newblock In \emph{Findings of the Association for Computational Linguistics: NAACL 2025}.

\bibitem[{Romanou et~al.(2024)Romanou, Foroutan, Sotnikova, Chen, Nelaturu, Singh, Maheshwary, Altomare, Haggag, Amayuelas et~al.}]{romanou2024include}
Angelika Romanou, Negar Foroutan, Anna Sotnikova, Zeming Chen, Sree~Harsha Nelaturu, Shivalika Singh, Rishabh Maheshwary, Micol Altomare, Mohamed~A Haggag, Alfonso Amayuelas, and 1 others. 2024.
\newblock {INCLUDE}: Evaluating multilingual language understanding with regional knowledge.
\newblock \emph{arXiv preprint arXiv:2411.19799}.

\bibitem[{Roxas et~al.(2021)Roxas, Imperial, and De~La~Cruz}]{roxas-etal-2021-science}
Rachel Edita~O. Roxas, Joseph~Marvin Imperial, and Angelica~H. De~La~Cruz. 2021.
\newblock \href {https://aclanthology.org/2021.paclic-1.76/} {Science mapping of publications in natural language processing in the {P}hilippines: 2006 to 2020}.
\newblock In \emph{Proceedings of the 35th Pacific Asia Conference on Language, Information and Computation}, pages 721--730, Shanghai, China. Association for Computational Lingustics.

\bibitem[{Singh et~al.(2024)Singh, Romanou, Fourrier, Adelani, Ngui, Vila-Suero, Limkonchotiwat, Marchisio, Leong, Susanto et~al.}]{singh2024global}
Shivalika Singh, Angelika Romanou, Cl{\'e}mentine Fourrier, David~I Adelani, Jian~Gang Ngui, Daniel Vila-Suero, Peerat Limkonchotiwat, Kelly Marchisio, Wei~Qi Leong, Yosephine Susanto, and 1 others. 2024.
\newblock Global {MMLU}: Understanding and addressing cultural and linguistic biases in multilingual evaluation.
\newblock \emph{arXiv preprint arXiv:2412.03304}.

\bibitem[{Susanto et~al.(2025)Susanto, Hulagadri, Montalan, Ngui, Yong, Leong, Rengarajan, Limkonchotiwat, Mai, and Tjhi}]{susanto2025sea}
Yosephine Susanto, Adithya~Venkatadri Hulagadri, Jann~Railey Montalan, Jian~Gang Ngui, Xian~Bin Yong, Weiqi Leong, Hamsawardhini Rengarajan, Peerat Limkonchotiwat, Yifan Mai, and William~Chandra Tjhi. 2025.
\newblock {SEA-HELM}: Southeast asian holistic evaluation of language models.
\newblock \emph{arXiv preprint arXiv:2502.14301}.

\bibitem[{Team et~al.(2025)Team, Kamath, Ferret, Pathak, Vieillard, Merhej, Perrin, Matejovicova, Ram{\'e}, Rivi{\`e}re et~al.}]{team2025gemma}
Gemma Team, Aishwarya Kamath, Johan Ferret, Shreya Pathak, Nino Vieillard, Ramona Merhej, Sarah Perrin, Tatiana Matejovicova, Alexandre Ram{\'e}, Morgane Rivi{\`e}re, and 1 others. 2025.
\newblock Gemma 3 technical report.
\newblock \emph{arXiv preprint arXiv:2503.19786}.

\bibitem[{Team et~al.(2024)Team, Riviere, Pathak, Sessa, Hardin, Bhupatiraju, Hussenot, Mesnard, Shahriari, Ram{\'e} et~al.}]{team2024gemma}
Gemma Team, Morgane Riviere, Shreya Pathak, Pier~Giuseppe Sessa, Cassidy Hardin, Surya Bhupatiraju, L{\'e}onard Hussenot, Thomas Mesnard, Bobak Shahriari, Alexandre Ram{\'e}, and 1 others. 2024.
\newblock Gemma 2: {I}mproving open language models at a practical size.
\newblock \emph{arXiv preprint arXiv:2408.00118}.

\bibitem[{Tiedemann(2020)}]{tiedemann-2020-tatoeba}
Jörg Tiedemann. 2020.
\newblock The tatoeba translation challenge {--} realistic data sets for low resoure and multilingual {MT}.
\newblock In \emph{Proceedings of the Fifth Conference on Machine Translation}, pages 1174--1182.

\bibitem[{Villafania(2007)}]{Villafania2007}
Sonny Villafania. 2007.
\newblock \href {https://web.archive.org/web/20140522052247/http://svillafania.philippinepen.ph/2007/08/articles-filipino-and-tagalog-not-so.html} {Filipino and {Tagalog}, not so different}.
\newblock Archived from the original on 2014-05-22.

\bibitem[{Wan et~al.(2022)Wan, Yang, Wong, Chao, Yao, Zhang, and Chen}]{wan-etal-2022-challenges}
Yu~Wan, Baosong Yang, Derek~Fai Wong, Lidia~Sam Chao, Liang Yao, Haibo Zhang, and Boxing Chen. 2022.
\newblock \href {https://doi.org/10.1162/coli_a_00435} {Challenges of neural machine translation for short texts}.
\newblock \emph{Computational Linguistics}, 48(2):321--342.

\bibitem[{Wang et~al.(2023)Wang, Sun, Li, Ouyang, Wu, Zhang, Li, and Wang}]{wang2023gpt}
Shuhe Wang, Xiaofei Sun, Xiaoya Li, Rongbin Ouyang, Fei Wu, Tianwei Zhang, Jiwei Li, and Guoyin Wang. 2023.
\newblock Gpt-ner: Named entity recognition via large language models.
\newblock \emph{arXiv preprint arXiv:2304.10428}.

\bibitem[{Yang et~al.(2024)Yang, Yang, Zhang, Hui, Zheng, Yu, Li, Liu, Huang, Wei et~al.}]{yang2024qwen2}
An~Yang, Baosong Yang, Beichen Zhang, Binyuan Hui, Bo~Zheng, Bowen Yu, Chengyuan Li, Dayiheng Liu, Fei Huang, Haoran Wei, and 1 others. 2024.
\newblock Qwen2.5 {T}echnical {R}eport.
\newblock \emph{arXiv preprint arXiv:2412.15115}.

\bibitem[{Yue et~al.(2024)Yue, Song, Asai, Kim, de~Dieu~Nyandwi, Khanuja, Kantharuban, Sutawika, Ramamoorthy, and Neubig}]{yue2024pangea}
Xiang Yue, Yueqi Song, Akari Asai, Seungone Kim, Jean de~Dieu~Nyandwi, Simran Khanuja, Anjali Kantharuban, Lintang Sutawika, Sathyanarayanan Ramamoorthy, and Graham Neubig. 2024.
\newblock Pangea: {A} {F}ully {O}pen {M}ultilingual {M}ultimodal {LLM} for 39 {L}anguages.
\newblock In \emph{The Thirteenth International Conference on Learning Representations}.

\bibitem[{Zhang et~al.(2023{\natexlab{a}})Zhang, Haddow, and Birch}]{zhang2023prompting}
Biao Zhang, Barry Haddow, and Alexandra Birch. 2023{\natexlab{a}}.
\newblock {Prompting Large Language Model for Machine Translation: A Case Study}.
\newblock \emph{arXiv preprint arXiv:2301.07069}.

\bibitem[{Zhang et~al.(2024)Zhang, Chan, Zhao, Aljunied, Wang, Liu, Deng, Hu, Xu, Chia et~al.}]{zhang2024seallms}
Wenxuan Zhang, Hou~Pong Chan, Yiran Zhao, Mahani Aljunied, Jianyu Wang, Chaoqun Liu, Yue Deng, Zhiqiang Hu, Weiwen Xu, Yew~Ken Chia, and 1 others. 2024.
\newblock {SeaLLMs 3: Open Foundation and Chat Multilingual Large Language Models for Southeast Asian Languages}.
\newblock \emph{arXiv preprint arXiv:2407.19672}.

\bibitem[{Zhang et~al.(2023{\natexlab{b}})Zhang, Deng, Liu, Pan, and Bing}]{zhang2023sentiment}
Wenxuan Zhang, Yue Deng, Bing Liu, Sinno~Jialin Pan, and Lidong Bing. 2023{\natexlab{b}}.
\newblock Sentiment analysis in the era of large language models: A reality check.
\newblock \emph{arXiv preprint arXiv:2305.15005}.

\bibitem[{Zhao et~al.(2025)Zhao, Liu, Deng, Ying, Aljunied, Li, Bing, Chan, Rong, Zhao et~al.}]{zhao2025babel}
Yiran Zhao, Chaoqun Liu, Yue Deng, Jiahao Ying, Mahani Aljunied, Zhaodonghui Li, Lidong Bing, Hou~Pong Chan, Yu~Rong, Deli Zhao, and 1 others. 2025.
\newblock {Babel: Open Multilingual Large Language Models Serving Over 90\% of Global Speakers}.
\newblock \emph{arXiv preprint arXiv:2503.00865}.

\bibitem[{Zhu et~al.(2023)Zhu, Liu, Dong, Xu, Huang, Kong, Chen, and Li}]{zhu2023multilingual}
Wenhao Zhu, Hongyi Liu, Qingxiu Dong, Jingjing Xu, Shujian Huang, Lingpeng Kong, Jiajun Chen, and Lei Li. 2023.
\newblock Multilingual machine translation with large language models: Empirical results and analysis.
\newblock \emph{arXiv preprint arXiv:2304.04675}.

\end{thebibliography}
% \bibliography{custom,anthology}

% =======================================================
\clearpage
\appendix

\onecolumn
\addtocontents{toc}{\protect\setcounter{tocdepth}{2}}
\section*{Appendix}
\renewcommand{\contentsname}{}
\vspace{-2em}
\tableofcontents
\twocolumn

\onecolumn
\section{Details of Models Evaluated on \filbench{}}

\autoref{table:filbench_models} shows the details of all models evaluated on \filbench{}.

\begin{table}[h]
    \centering
    \renewcommand{\arraystretch}{0.6}
    \resizebox*{\linewidth}{!}{
        \begin{tabular}{lrrrr}
            \toprule
            \textbf{Model} & \textbf{\# Params (B)}       & \textbf{\# Lang.} & \textbf{License} & \textbf{Reference} \\
            \midrule
            \mul\href{https://platform.openai.com/docs/models}{gpt-4o-2024-08-06}
                           & $-$
                           & $-$
                           & OpenAI ToS
                           & \citet{hurst2024gpt}
            \\
            \mul\href{https://platform.openai.com/docs/models}{gpt-4o-mini}
                           & $-$
                           & $-$
                           & OpenAI ToS
                           & \citet{hurst2024gpt}
            \\
            \mul\href{https://huggingface.co/CohereForAI/aya-expanse-32b}{CohereForAI/aya-expanse-32b}
                           & 32
                           & 23
                           & CC BY NC 4.0
                           & \citet{dang2024aya}
            \\
            \mul\href{https://huggingface.co/meta-llama/Llama-4-Maverick-17B-128E-Instruct-FP8}{meta-llama/Llama-4-Maverick-17B-128E-Instruct-FP8}
                           & 400 (17)
                           & 200
                           & Llama 4 License
                           & \citet{metaai2025llama4blog}
            \\
            \mul\href{https://huggingface.co/meta-llama/Llama-4-Scout-17B-16E-Instruct}{meta-llama/Llama-4-Scout-17B-16E-Instruct}
                           & 109 (17)
                           & 200
                           & Llama 4 License
                           & \citet{metaai2025llama4blog}
            \\
            \mul\href{https://huggingface.co/meta-llama/Llama-3.1-70B-Instruct}{meta-llama/Llama-3.1-70B-Instruct}
                           & 70
                           & 30
                           & Llama 3.1 License
                           & \citet{grattafiori2024llama}
            \\
            \mul\href{https://huggingface.co/meta-llama/Llama-3.1-8B-Instruct}{meta-llama/Llama-3.1-8B-Instruct}
                           & 8
                           & 30
                           & Llama 3.1 License
                           & \citet{grattafiori2024llama}
            \\
            \mul\href{https://huggingface.co/Qwen/Qwen2.5-72B-Instruct}{Qwen/Qwen2.5-72B-Instruct}
                           & 72
                           & 29
                           & Qwen License
                           & \citet{yang2024qwen2}
            \\
            \mul\href{https://huggingface.co/Qwen/Qwen2.5-32B-Instruct}{Qwen/Qwen2.5-32B-Instruct}
                           & 32
                           & 29
                           & Apache 2.0
                           & \citet{yang2024qwen2}
            \\
            \mul\href{https://huggingface.co/Qwen/Qwen2.5-14B-Instruct}{Qwen/Qwen2.5-14B-Instruct}
                           & 14
                           & 29
                           & Apache 2.0
                           & \citet{yang2024qwen2}
            \\
            \mul\href{https://huggingface.co/Qwen/Qwen2.5-7B-Instruct}{Qwen/Qwen2.5-7B-Instruct}
                           & 7
                           & 29
                           & Apache 2.0
                           & \citet{yang2024qwen2}
            \\
            \mul\href{https://huggingface.co/Tower-Babel/Babel-83B-Chat}{Tower-Babel/Babel-83B-Chat}
                           & 83
                           & 25
                           & SeaLLM License
                           & \citet{zhao2025babel}
            \\
            \mul\href{https://huggingface.co/Tower-Babel/Babel-9B-Chat}{Tower-Babel/Babel-9B-Chat}
                           & 9
                           & 25
                           & SeaLLM License
                           & \citet{zhao2025babel}
            \\
            \mul\href{https://huggingface.co/google/gemma-3-27b-it}{google/gemma-3-27b-it}
                           & 27
                           & 73
                           & Gemma License
                           & \citet{team2025gemma}
            \\
            \mul\href{https://huggingface.co/google/gemma-2-27b-it}{google/gemma-2-27b-it}
                           & 27
                           & 73
                           & Gemma License
                           & \citet{team2024gemma}
            \\
            \mul\href{https://huggingface.co/google/gemma-2-9b-it}{google/gemma-2-9b-it}
                           & 9
                           & 73
                           & Gemma License
                           & \citet{team2024gemma}
            \\
            \mul\href{https://huggingface.co/mistralai/Ministral-8B-Instruct-2410}{mistralai/Ministral-8B-Instruct-2410}
                           & 8
                           & 10
                           & Mistral AI License
                           & \citet{mistralai2024ministrauxblog}
            \\
            \mul\href{https://huggingface.co/mistralai/Mixtral-8x22B-Instruct-v0.1}{mistralai/Mixtral-8x22B-Instruct-v0.1}
                           & 141 (39)
                           & 5
                           & Apache 2.0
                           & \citet{jiang2024mixtral}
            \\

            \mul\href{https://huggingface.co/mistralai/Mixtral-8x7B-Instruct-v0.1}{mistralai/Mixtral-8x7B-Instruct-v0.1}
                           & 47 (13)
                           & 5
                           & Apache 2.0
                           & \citet{jiang2024mixtral}
            \\
            \mul\href{https://huggingface.co/neulab/Pangea-7B}{neulab/Pangea-7B}
                           & 7
                           & 39
                           & Apache 2.0
                           & \citet{yue2024pangea}
            \\
            \sea\href{https://huggingface.co/aisingapore/Llama-SEA-LION-v3-70B-IT}{aisingapore/Llama-SEA-LION-v3-70B-IT}
                           & 70
                           & 13
                           & Llama 3.1 License
                           & \citet{sealionv3}
            \\
            \sea\href{https://huggingface.co/aisingapore/Gemma-SEA-LION-v3-9B-IT}{aisingapore/Gemma-SEA-LION-v3-9B-IT}
                           & 9
                           & 13
                           & Gemma License
                           & \citet{sealionv3}
            \\
            \sea\href{https://huggingface.co/aisingapore/Llama-SEA-LION-v3-8B-IT}{aisingapore/Llama-SEA-LION-v3-8B-IT}
                           & 8
                           & 13
                           & Llama 3.1 License
                           & \citet{sealionv3}
            \\
            \sea\href{https://huggingface.co/sail/Sailor2-20B-Chat}{sail/Sailor2-20B-Chat}
                           & 20
                           &  12
                           & Apache 2.0
                           & \citet{dou2025sailor2}
            \\
            \sea\href{https://huggingface.co/sail/Sailor2-8B-Chat}{sail/Sailor2-8B-Chat}
                           & 8
                           & 12
                           & Apache 2.0
                           & \citet{dou2025sailor2}
            \\
            \sea\href{https://huggingface.co/SeaLLMs/SeaLLMs-v3-7B-Chat}{SeaLLMs/SeaLLMs-v3-7B-Chat}
                           & 7
                           & 12
                           & SeaLLM License
                           & \citet{zhang2024seallms}
            \\
            \sea\href{https://huggingface.co/SeaLLMs/SeaLLMs-v3-1.5B-Chat}{SeaLLMs/SeaLLMs-v3-1.5B-Chat}
                           & 1.5
                           & 12
                           & SeaLLM License
                           & \citet{zhang2024seallms}
            \\
            \bottomrule
        \end{tabular}
    }
    \caption{
        All models evaluated on \filbench{}. 
        We evaluate several models with different multilingual capabilities (multilingual \mul, SEA-specific \sea), sizes (1.5B to 400B), and accessibility (open-source vs. commercial). 
        For Mixture-of-Experts models, parameters are denoted as "Total Parameters (Active Parameters)". 
        Models that are finetuned on top of a pre-trained model have the number of languages supported based on their fine-tuning data. 
    }
    \label{table:filbench_models}
\end{table}

% Sources
% AI Singapore Gemma: https://huggingface.co/aisingapore/gemma2-9b-cpt-sea-lionv3-instruct
% Gemma: https://developers.googleblog.com/en/building-more-inclusive-llms-using-gemma-open-models/

\section{\filbench{} Dataset Licenses}
\autoref{table:filbench-datasets-licenses} provides information for all datasets in \filbench{}, such as their license and data collection process.
\begin{table}[h]
    \centering
    \renewcommand{\arraystretch}{0.8}
    \resizebox*{\linewidth}{!}{
        \begin{tabular}{llllr}
            \toprule
            \textbf{Category} & \textbf{Dataset}                                              & \textbf{Source}           & \textbf{Annotation}        & \textbf{License} \\
            \midrule
            CN                & Dengue Filipino \cite{livelo-cheng-2018-dengue}               & Social media (Twitter)    & Expert-annotated           & Unknown          \\
                              & BalitaNLP \cite{bunag2023news}                                & News articles             & Included from source       & Unknown          \\
                              & SIB-200 \cite{adelani-etal-2024-sib}                          & Human-translation         & Expert-annotated           & CC BY SA 4.0     \\
                              & CebuaNER \cite{pilar-etal-2023-cebuaner}                      & News articles             & Expert-annotated           & CC BY NC SA 4.0  \\
                              & TLUnified-NER \cite{miranda-2023-developing}                  & News articles             & Expert-annotated           & GPL v3.0         \\
                              & Universal NER \cite{mayhew-etal-2024-universal}               & Universal Dependencies    & Expert-annotated           & CC BY SA 4.0     \\
                              & FiReCS \cite{cosme-deleon-2024-products}                      & Reviews (Maps and Shopee) & Expert-annotated           & CC BY 4.0        \\
            \midrule
            CK                & INCLUDE \cite{romanou2024include}                             & Local exams               & Expert-annotated           & Apache 2.0       \\
                              & Global MMLU \cite{singh2024global}                            & MMLU dataset              & Translated with validation & Apache 2.0       \\
                              & KALAHI \cite{montalan2024kalahi}                              & Human-provided            & Expert-annotated           & CC BY 4.0        \\
                              & StingrayBench \cite{cahyawijaya2024thank}                     & Human-provided            & Expert-annotated           & CC BY SA 4.0     \\
            \midrule
            RC                & Cebuano Readability Corpus \cite{imperial-etal-2022-baseline} & Book repositories         & Expert-annotated           & MIT              \\
                              & Belebele \cite{bandarkar-etal-2024-belebele}                  & Wikipedia                 & Expert-annotated           & CC BY SA 4.0     \\
                              & NewsPH NLI \cite{cruz-etal-2021-news}                         & News articles             & Semi-supervised            & Unknown          \\
            \midrule
            GN                & NTREX-128 \cite{federmann-etal-2022-ntrex}                    & Translated from WMT19     & Expert-annotated           & CC BY SA 4.0     \\
                              & Tatoeba \cite{tiedemann-2020-tatoeba}                         & Crowd-sourced             & Crowd-sourced              & CC BY 2.0        \\
                              & TICO-19 \cite{anastasopoulos-etal-2020-tico}                  & News, Wikipedia, PubMed   & Semi-supervised            & CC0 1.0          \\
            \bottomrule
        \end{tabular}
    }
    \caption{
        Supplemental information for all datasets included in \filbench{}.
        For datasets with ``Unknown'' licenses, we obtained explicit approval from the authors to include them in our evaluation suite.
    }
    \label{table:filbench-datasets-licenses}
\end{table}

\clearpage
\section{Full results on \filbench{}}

\autoref{table:main_results_all_agg} shows the full aggregated results for the \nummodels{} models evaluated on \filbench{}.
%We plan to release a public version of the leaderboard after the review period.

\begin{table*}[h]
    \centering
    \resizebox*{\linewidth}{!}{
        \begin{tabular}{lrrrrr}
            \toprule
            % \begin{noindent}
        \textbf{Model}                              & \textbf{\thead[r]{\textsc{\filbench{}}\\Score}} & \textbf{\thead[r]{Cultural\\Knowledge}} & \textbf{\thead[r]{Classical\\NLP}} & \textbf{\thead[r]{Reading\\Comp.}} & \textbf{Generation} \\
        % \end{noindent}
            \midrule
            \mul\href{https://platform.openai.com/docs/models}{gpt-4o-2024-08-06}                                                                  & 72.73{\small$\pm$1.66} & 73.29{\small$\pm$3.01} & 89.03{\small$\pm$2.05} & 80.12{\small$\pm$0.90} & 46.48{\small$\pm$0.60} \\
            \mul\href{https://huggingface.co/meta-llama/Llama-4-Maverick-17B-128E-Instruct-FP8}{meta-llama/Llama-4-Maverick-17B-128E-Instruct-FP8} & 67.67{\small$\pm$1.04} & 76.75{\small$\pm$3.04} & 87.28{\small$\pm$0.26} & 72.99{\small$\pm$0.18} & 33.67{\small$\pm$0.71} \\
            \mul\href{https://huggingface.co/meta-llama/Llama-4-Scout-17B-16E-Instruct }{meta-llama/Llama-4-Scout-17B-16E-Instruct}                & 63.20{\small$\pm$1.05} & 74.31{\small$\pm$3.14} & 87.88{\small$\pm$0.25} & 70.86{\small$\pm$0.18} & 19.75{\small$\pm$0.63} \\
            \mul\href{https://huggingface.co/Qwen/Qwen2.5-72B-Instruct}{Qwen/Qwen2.5-72B-Instruct}                                                 & 63.08{\small$\pm$0.99} & 73.11{\small$\pm$3.22} & 88.60{\small$\pm$0.24} & 75.62{\small$\pm$0.17} & 14.98{\small$\pm$0.33} \\
            \sea\href{https://huggingface.co/aisingapore/Llama-SEA-LION-v3-70B-IT}{aisingapore/Llama-SEA-LION-v3-70B-IT}                           & 61.07{\small$\pm$0.95} & 76.78{\small$\pm$3.02} & 89.99{\small$\pm$0.23} & 53.56{\small$\pm$0.19} & 23.95{\small$\pm$0.34} \\
            \mul\href{https://huggingface.co/Tower-Babel/Babel-83B-Chat}{Tower-Babel/Babel-83B-Chat}                                               & 60.85{\small$\pm$0.96} & 75.21{\small$\pm$3.11} & 88.81{\small$\pm$0.25} & 64.85{\small$\pm$0.19} & 14.53{\small$\pm$0.29} \\
            \mul\href{https://huggingface.co/meta-llama/Llama-3.1-70B-Instruct}{meta-llama/Llama-3.1-70B-Instruct}                                 & 59.66{\small$\pm$1.17} & 72.16{\small$\pm$3.21} & 90.27{\small$\pm$0.83} & 52.17{\small$\pm$0.28} & 24.03{\small$\pm$0.37} \\
            \sea\href{https://huggingface.co/sail/Sailor2-20B-Chat}{sail/Sailor2-20B-Chat}                                                         & 58.61{\small$\pm$1.06} & 66.43{\small$\pm$3.41} & 89.03{\small$\pm$0.25} & 63.03{\small$\pm$0.19} & 15.95{\small$\pm$0.38} \\
            \mul\href{https://huggingface.co/Qwen/Qwen2.5-32B-Instruct}{Qwen/Qwen2.5-32B-Instruct}                                                 & 57.88{\small$\pm$1.45} & 66.83{\small$\pm$3.45} & 89.32{\small$\pm$1.99} & 70.59{\small$\pm$0.18} & 4.79{\small$\pm$0.17}  \\
            \sea\href{https://huggingface.co/aisingapore/Gemma-SEA-LION-v3-9B-IT}{aisingapore/Gemma-SEA-LION-v3-9B-IT}                             & 56.14{\small$\pm$1.53} & 64.44{\small$\pm$3.43} & 88.55{\small$\pm$0.25} & 54.46{\small$\pm$0.20} & 17.10{\small$\pm$2.25} \\
            \mul\href{https://huggingface.co/google/gemma-2-27b-it}{google/gemma-2-27b-it}                                                         & 55.22{\small$\pm$1.04} & 68.76{\small$\pm$3.32} & 87.99{\small$\pm$0.25} & 48.77{\small$\pm$0.19} & 15.38{\small$\pm$0.38} \\
            \mul\href{https://huggingface.co/google/gemma-3-27b-it}{google/gemma-3-27b-it}                                                         & 55.17{\small$\pm$0.99} & 71.41{\small$\pm$3.24} & 88.61{\small$\pm$0.24} & 53.23{\small$\pm$0.19} & 7.42 {\small$\pm$0.30} \\
            \mul\href{https://huggingface.co/mistralai/Mixtral-8x22B-Instruct-v0.1}{mistralai/Mixtral-8x22B-Instruct-v0.1}                         & 54.28{\small$\pm$1.09} & 54.47{\small$\pm$3.62} & 87.19{\small$\pm$0.25} & 64.78{\small$\pm$0.19} & 10.70{\small$\pm$0.31} \\
            \mul\href{https://huggingface.co/google/gemma-2-9b-it}{google/gemma-2-9b-it}                                                           & 53.33{\small$\pm$1.08} & 63.69{\small$\pm$3.47} & 87.47{\small$\pm$0.25} & 50.65{\small$\pm$0.20} & 11.51{\small$\pm$0.40} \\
            \mul\href{https://huggingface.co/Tower-Babel/Babel-9B-Chat}{Tower-Babel/Babel-9B-Chat}                                                 & 52.75{\small$\pm$1.48} & 60.06{\small$\pm$3.57} & 87.67{\small$\pm$1.90} & 56.49{\small$\pm$0.20} & 6.79 {\small$\pm$0.26} \\
            \sea\href{https://huggingface.co/sail/Sailor2-8B-Chat}{sail/Sailor2-8B-Chat}                                                           & 52.49{\small$\pm$1.10} & 58.94{\small$\pm$3.57} & 86.03{\small$\pm$0.27} & 50.69{\small$\pm$0.23} & 14.29{\small$\pm$0.36} \\
            \mul\href{https://huggingface.co/Qwen/Qwen2.5-14B-Instruct}{Qwen/Qwen2.5-14B-Instruct}                                                 & 52.41{\small$\pm$1.63} & 59.27{\small$\pm$3.61} & 86.27{\small$\pm$2.56} & 59.95{\small$\pm$0.20} & 4.14 {\small$\pm$0.14} \\
            \mul\href{https://huggingface.co/Qwen/Qwen2.5-7B-Instruct}{Qwen/Qwen2.5-7B-Instruct}                                                   & 50.46{\small$\pm$1.08} & 51.61{\small$\pm$3.68} & 85.58{\small$\pm$0.27} & 60.47{\small$\pm$0.20} & 4.19 {\small$\pm$0.15} \\
            \sea\href{https://huggingface.co/aisingapore/llama3.1-8b-cpt-sea-lionv3-instruct}{aisingapore/Llama-SEA-LION-v3-8B-IT}                 & 50.32{\small$\pm$1.08} & 59.89{\small$\pm$3.56} & 83.33{\small$\pm$0.28} & 47.47{\small$\pm$0.10} & 10.60{\small$\pm$0.29} \\
            \mul\href{https://huggingface.co/mistralai/Mixtral-8x7B-Instruct-v0.1}{mistralai/Mixtral-8x7B-Instruct-v0.1}                           & 50.26{\small$\pm$1.09} & 49.88{\small$\pm$3.67} & 84.19{\small$\pm$0.29} & 60.95{\small$\pm$0.19} & 6.02 {\small$\pm$0.31} \\
            \sea\href{https://huggingface.co/SeaLLMs/SeaLLMs-v3-7B-Chat}{SeaLLMs/SeaLLMs-v3-7B-Chat}                                               & 49.06{\small$\pm$1.06} & 52.04{\small$\pm$3.66} & 79.68{\small$\pm$0.33} & 62.47{\small$\pm$0.19} & 2.08 {\small$\pm$0.10} \\
            \mul\href{https://huggingface.co/CohereForAI/aya-expanse-32b}{CohereForAI/aya-expanse-32b}                                             & 47.84{\small$\pm$1.41} & 53.22{\small$\pm$3.65} & 87.47{\small$\pm$1.60} & 46.09{\small$\pm$0.21} & 4.58 {\small$\pm$0.16} \\
            \mul\href{https://huggingface.co/meta-llama/Llama-3.1-8B-Instruct}{meta-llama/Llama-3.1-8B-Instruct}                                   & 47.38{\small$\pm$1.51} & 52.08{\small$\pm$3.68} & 86.61{\small$\pm$1.90} & 46.42{\small$\pm$0.24} & 4.42 {\small$\pm$0.20} \\
            \mul\href{https://huggingface.co/mistralai/Ministral-8B-Instruct-2410}{mistralai/Ministral-8B-Instruct-2410}                           & 47.33{\small$\pm$1.66} & 42.02{\small$\pm$3.62} & 77.95{\small$\pm$2.59} & 62.33{\small$\pm$0.20} & 7.00 {\small$\pm$0.25} \\
            \mul\href{https://huggingface.co/neulab/Pangea-7B}{neulab/Pangea-7B}                                                                   & 43.98{\small$\pm$1.08} & 46.23{\small$\pm$3.70} & 78.80{\small$\pm$0.29} & 47.74{\small$\pm$0.22} & 3.15 {\small$\pm$0.15} \\
            \sea\href{https://huggingface.co/SeaLLMs/SeaLLMs-v3-1.5B-Chat}{SeaLLMs/SeaLLMs-v3-1.5B-Chat}                                           & 43.20{\small$\pm$1.07} & 37.14{\small$\pm$3.61} & 75.17{\small$\pm$0.33} & 56.85{\small$\pm$0.20} & 2.08 {\small$\pm$0.14} \\
            \mul\href{https://platform.openai.com/docs/models}{gpt-4o-mini}                                                                        & 42.32{\small$\pm$1.81} & 25.09{\small$\pm$3.26} & 73.12{\small$\pm$3.18} & 47.78{\small$\pm$0.34} & 23.29{\small$\pm$0.59} \\
            \bottomrule
        \end{tabular}
    }
    \caption{
        Model performance on \filbench{}.
        We evaluate several models with different multilingual capabilities (multilingual \mul, SEA-specific \sea), sizes (8B to 400B), and accessibility (open-source vs. commercial).
    }
    \label{table:main_results_all_agg}
\end{table*}

\section{Generation Few-shot Results}
\label{sec:few-shot}

\autoref{table:generation-few-shot} shows the full few-shot experiment results on the Generation category of \filbench{} for 9 selected models.

\begin{table*}[!h]
    \centering
    \resizebox*{\linewidth}{!}{
        \begin{tabular}{@{}lrrrrrrrrrrrrrrrr@{}}
            \toprule
                                                                                                     & \multicolumn{4}{c}{\textbf{Tatoeba - TGL}}            & \multicolumn{4}{c}{\textbf{Tatoeba - CEB}}            & \multicolumn{4}{c}{\textbf{NTREX-128}}                & \multicolumn{4}{c}{\textbf{TICO-19}}                                                                                                                                                                                                                  \\
                                                                                                     & \multicolumn{4}{c}{(\texttt{ENG} $\to$ \texttt{FIL})} & \multicolumn{4}{c}{(\texttt{CEB} $\to$ \texttt{ENG})} & \multicolumn{4}{c}{(\texttt{ENG} $\to$ \texttt{FIL})} & \multicolumn{4}{c}{(\texttt{ENG} $\to$ \texttt{FIL})}                                                                                                                                                                                                 \\
            \cmidrule(lr){2-5}\cmidrule(lr){6-9}\cmidrule(lr){10-13}\cmidrule(lr){14-17}
            \textbf{Model / $k$-shot \#}       & \multicolumn{1}{c}{0}                                                       & \multicolumn{1}{c}{1} & \multicolumn{1}{c}{3} & \multicolumn{1}{c}{5} & 
            \multicolumn{1}{c}{0} &
            \multicolumn{1}{c}{1} & \multicolumn{1}{c}{3} & \multicolumn{1}{c}{5} &
            \multicolumn{1}{c}{0} &
            \multicolumn{1}{c}{1} & \multicolumn{1}{c}{3} & \multicolumn{1}{c}{5} & 
            \multicolumn{1}{c}{0} &
            \multicolumn{1}{c}{1} & \multicolumn{1}{c}{3} & \multicolumn{1}{c}{5} \\ \midrule
            \mul \href{https://platform.openai.com/docs/models}{gpt-4o-2024-08-06}  & 51.88                      & 60.23                                                 & 60.62                                                   & 61.65 & 33.78                                               & 59.37                                                 & 62.99                   & 63.98  & 38.96               & 57.09                 & 58.47                  & 58.56 &   53.03              & 64.42                 & 64.08                   & 65.15                 \\
            \mul \href{https://platform.openai.com/docs/models}{gpt-4o-mini} & 12.13                     & 51.69                                                 & 55.20                                                  & 60.23    & 27.07                                             & 49.57                                                 & 58.30                   & 58.71  & 27.83               & 54.67                 & 57.81                   & 58.30  & 41.16                & 52.24                 & 64.08                   & 64.43                 \\
            \midrule
            \sea \href{https://huggingface.co/Sailor/Sailor2-20B-Chat}{Sailor/Sailor2-20B-Chat} & 15.88          & 17.13                                                 & 18.31                                                   & 22.34   & 13.67                                              & 10.60                                                 & 12.07                   & 13.19  & 23.41                & 44.45                 & 43.84                   & 44.50 & 22.88                & 54.21                 & 53.08                   & 55.05                 \\
            \sea \href{https://huggingface.co/aisingapore/Llama-SEA-LION-v3-8B-IT}{aisingapore/Llama-SEA-LION-v3-8B-IT} & 1.45  & 14.93                                                 & 15.25                                                  & 15.10  & 9.01                                               & 10.75                                                 & 12.33                   & 12.26  & 14.79               & 39.74                 & 40.41                   &  40.52 & 22.84               & 44.04                & 43.91                   & 44.05                 \\
            \mul \href{https://huggingface.co/CohereForAI/aya-expanse-32b}{CohereForAI/aya-expanse-32b} & 0.80 & 14.03                                                 & 13.86                                                  & 13.60 & 8.31                                                & 10.27                                                 & 11.51                   & 11.93 & 6.72                & 33.71                 & 33.70                   & 36.12 & 8.48                 & 39.55                 & 38.88                   & 39.92                 \\

            \sea \href{https://huggingface.co/SeaLLMs/SeaLLMs-v3-7B-Chat}{SeaLLMs/SeaLLMs-v3-7B-Chat}  & 0.65 & 11.17                                                 & 11.70                                                 & 12.07  & 6.44                                                & 7.62                                                  & 9.47                  & 9.61  & 5.65                & 32.46                 & 33.88                 & 36.50 & 6.01                 &           39.41         & 39.13                 & 39.83                   \\
            \mul \href{https://huggingface.co/Qwen/Qwen-2.5-7B-Instruct}{Qwen/Qwen-2.5-7B-Instruct} & 0.72 & 8.37                                                  & 8.72                                                   & 9.38 & 6.60                                                 & 7.38                                                  & 8.93                   & 9.72  & 6.99                & 28.72                 & 29.59                   & 30.67 & 6.43                 & 32.06                 & 32.40                   & 33.16                 \\
            \mul \href{https://huggingface.co/neulab/Pangea-7B}{neulab/Pangea-7B} & 0.53                        & 5.73                                                  & 7.56                                                   & 7.69 & 7.06                                                 & 6.65                                                  & 8.29                   & 8.41 & 4.59                 & 23.36                 & 24.15                   & 25.60 & 5.40                & 31.28                 & 28.95                   & 28.73                 \\

            \sea \href{https://huggingface.co/SeaLLMs/SeaLLMs-v3-1.5B-Chat}{SeaLLMs/SeaLLMs-v3-1.5B-Chat} & 0.78 & 4.9                                                   & 6.43                                                   & 6.99 & 4.51                                                  & 6.72   & 6.36                   & 6.34  & 2.04                & 22.97                 & 26.10   & 28.16 & 2.15                 & 24.45                 & 32.27                   & 32.72                 \\

            \bottomrule
        \end{tabular}
    }
    \caption{Generation scores for few-shot prompting on selected models (multilingual \mul, SEA-specific \sea).}
    \label{table:generation-few-shot}
\end{table*}

\twocolumn

\section{Evaluation Infrastructure and Runtime}

\begin{figure}[h]
    \centering
    \includegraphics[width=0.95\linewidth]{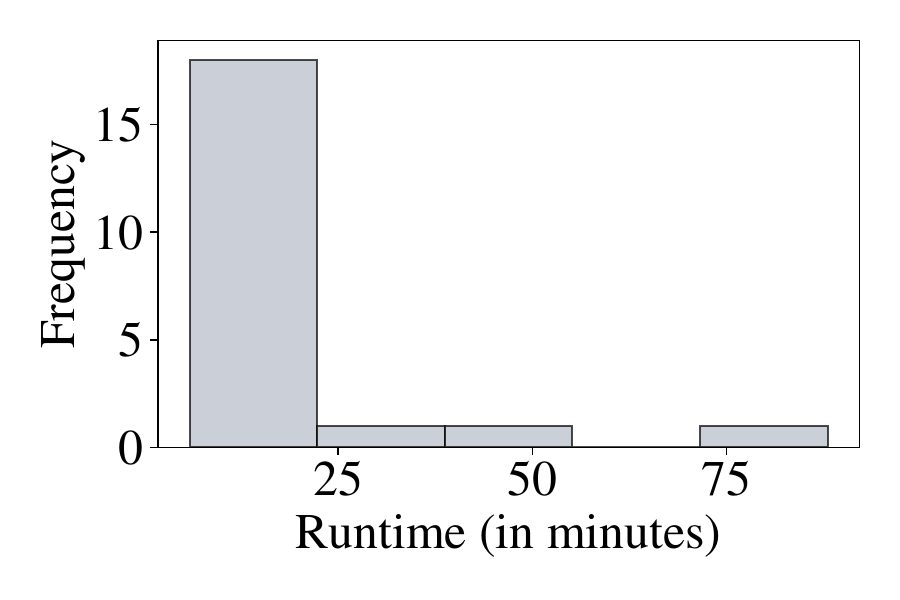} % Adjust width as needed
    \caption{Runtime of different benchmarks for a 32B model on \filbench{} (2 $\times$ H100 NVIDIA GPU).}
    \label{fig:runtime}
\end{figure}

We built \filbench{} on top of LightEval \citep{lighteval}.
When using the vLLM backend \citep{kwon2023efficient}, evaluating on the whole suite \textit{sequentially} can take 4.93 hours on 2 NVIDIA H100 GPUs for models under 83B parameters.
However, the evaluation suite can be parallelized per benchmark, with the runtime distribution shown in \autoref{fig:runtime}.
The longest-running task can take approximately 1 hour and 28 minutes and the shortest task takes only 5.86 minutes.

\begin{figure*}[t]
    \centering
    \includegraphics[width=0.95\linewidth]{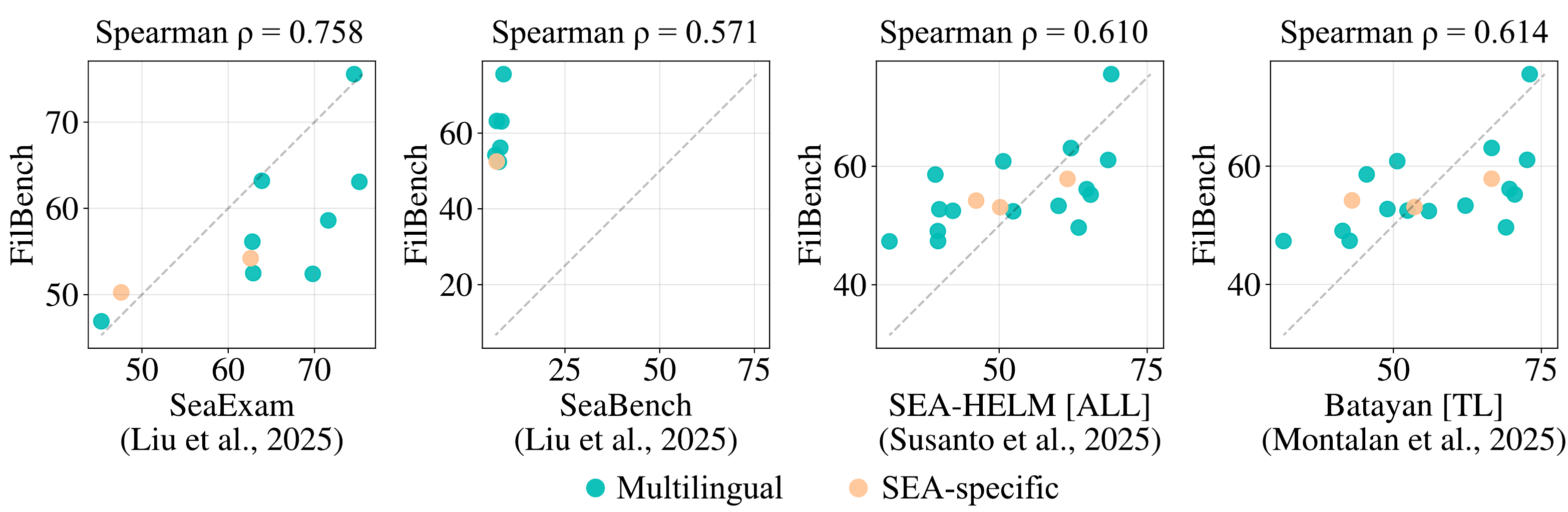} % Adjust width as needed
    \caption{
        Performance of different Multilingual and SEA-Specific models on FilBench and other SEA-specific / Tagalog benchmarks such as SeaExam and SeaBench \citep{liu2025seaexam}, SEA-HELM \citep{susanto2025sea}, and Batayan \citep{montalan2025batayan}.
    }
    \label{fig:rank_correlation}
\end{figure*}

\section{Extended Related Work}

In this section, we focus on other benchmarking efforts related to \filbench{}.
First, we compare the differences across these efforts (\S\ref{sec:comparison_sea}) and show \filbench{}'s value in providing a more focused evaluation for Filipino.
Then, we discuss whether there is a transferability in performance when evaluating from one benchmark to another (\S\ref{sec:comparison_sea_perf}).

\begin{table*}[b]
    \centering
     \resizebox*{\linewidth}{!}{
    \begin{tabular}{lrrrr}
        \toprule
        % \begin{noindent}
        \textbf{Benchmark}                              & \textbf{\# Tasks} & \textbf{\# Instances} & \textbf{PH Languages} & \textbf{Data Collection Procedure} \\
        % \end{noindent}
        \midrule
        \textbf{\filbench{} (\textsc{Ours})} & \numsubtasks{} & \numinstances{} & \texttt{FIL/TGL}, \texttt{CEB} & Curated from expert-annotated datasets \\
        SeaBench \citep{liu2025seaexam}      & 1              & 300             & -  & Collected from native-speakers                       \\
        SeaExam  \citep{liu2025seaexam}      & 3              & 5.5k            & -  &  Collected from native-speakers                       \\
        Batayan \citep{montalan2025batayan}  & 8              & 3.8k            & \texttt{FIL}  & Curated with human annotation             \\
        \bottomrule
    \end{tabular}
     }
    \caption{
        Comparison of multilingual benchmarks related to Filipino-centric tasks.
        Our data collection procedure allows us to scale the diversity of tasks in our suite.
        % We evaluate several models with different multilingual capabilities (multilingual \mul, SEA-specific \sea), sizes (1.5B to 400B), and accessibility (open-source vs. commercial).
        % Full results can be found in \autoref{table:main_results_all_agg}.
    }
    \label{table:benchmark_comparison}
\end{table*}

\subsection{Comparison to other SEA-specific / Filipino benchmarks}
\label{sec:comparison_sea}

\autoref{table:benchmark_comparison} shows benchmarking efforts orthogonal to \filbench{}.
These efforts focus on a specific region, i.e., Southeast Asia or SEA, and comprises of datasets from countries other than the Philippines.
In general, we find that SEA-specific benchmarks do not contain any Philippine language at all (as in the case of the SeaLLM Leaderboard, \citealt{liu2025seaexam}) or is limited to a single Filipino language (Tagalog, as in the case of SEA-HELM, \citealt{susanto2025sea}).
\filbench{} aims to provide a more realistic evaluation of Filipino-centric tasks by having a principled approach in choosing categories that reflect the current trends and priorities of the Philippine NLP research community.

\subsection{Does high performance in one benchmark translates similarly to \filbench{}?}
\label{sec:comparison_sea_perf}

\paragraph{Set-up.}
In order to understand whether high performance in one benchmark translates to similar performance in \filbench{}, we compute the Spearman $\rho$ rank correlation of models that were evaluated in both benchmarks.
For the SeaLLM leaderboard, we treat SeaBench and SeaExam separately.
For SEA-HELM, we compute the correlation for the full evaluation suite and its Tagalog-only subset (Batayan).

\paragraph{Results.}
\autoref{fig:rank_correlation} shows the raw scores for \filbench{} with respect to another benchmark, alongside its Spearman $\rho$ rank correlation.
The results show moderate to strong positive correlations ($\rho$ = 0.571 to 0.758) between FilBench and other SEA language benchmarks, with SeaExam demonstrating the strongest predictive relationship.
This suggests that model performance on one benchmark does meaningfully transfer to performance on Filipino language tasks, though the scattered distribution of data points indicates that different benchmarks capture distinct aspects of language ability.
Furthermore, our findings highlight that while some transferability exists across Southeast Asian language benchmarks, benchmark-specific optimization may still be necessary for optimal performance on FilBench.

\section{Analysis of Model Dis/Agreement}
\label{sec:appendix-disagreement}

In this section, we show examples of agreement and disagreement from the SEA-specific models we analyzed in \S\ref{sec:analysis-consistency}.

\subsection{Set-up: Qualitative Analysis of Model Outputs during Dis/Agreement}
In order to understand model behavior, we qualitatively analyze per-instance agreement between select sub-tasks within the \filbench{} evaluation suite.
This involves examining instances where models either consistently agree or disagree on their outputs.
By focusing on specific sub-tasks, such as Readability and Cultural Knowledge Assessment, we aim to identify patterns and potential sources of error or divergence in model predictions.
We hope that this analysis helps in understanding the nuances of model performance and the challenges posed by different task types.

\begin{figure}[t]
    \begin{promptbox}[High agreement among models but incorrect answer]
        Tanong: Ano ang ibig sabihin ng berdeng arrow sa signal na pang-trapiko?\\
        A. Hindi pinapayagan ang pagpasok sa interseksyong itinuturo ng arrow.\\
        B. Napapahintulot sa mga sasakyan na kumaliwa o kumanan.\\
        C. Nagpapahintulot sa pagtawid ng mga taong tatawid.\\
        D. Wala sa nabanggit.\\
        Sagot:\\
        ---\\
        \textbf{Model Pred (Majority):} A\\
        \textbf{Gold:} B

    \end{promptbox}
    \caption{
        In this example from a driving license exam, all SEA-specific models agree that the correct answer is A.
        However, the gold label is B.
    }
    \label{fig:regional_high_agreement_wrong_answer}
\end{figure}

\begin{figure}[t]
    \begin{promptbox}[High disagreement among models]
        Tanong: Sa isang sangandaan/interseksyon na walang senyas trapiko, dalawang sasakyan ang dumarating sa magkabilang kalye, aling sasakyan ang dapat magbigay?\\
        A. Ang unang dumating\\
        B. Ang unang nagmarahan\\
        C. Ang huling dumating\\
        D. Wala sa nabanggit\\
        Sagot:\\
        ---\\
        \textbf{Gold:} C
    \end{promptbox}
    \caption{
        In this example from a driving license exam, all SEA-specific models disagree on their answers.
    }
    \label{fig:regional_high_disagreement}
\end{figure}

\subsection{Results: Examples of Model Dis/Agreement}

\paragraph{Regional Knowledge (Fleiss' $\kappa=0.393$)}
For this task, models are required to answer questions taken from a sample of a driving exam in the Philippines.
\autoref{fig:regional_high_agreement_wrong_answer} shows an example where models agree on a specific answer, yet they are all incorrect.
The question asks what a green arrow (\textit{berdeng arrow}) indicates as a traffic signal.
All models answered Option A (Vehicles are \textbf{not allowed} to enter the intersection as pointed by the arrow), yet the correct answer is Option B (Vehicles are \textbf{allowed} to turn left or right).

We also show an example where most SEA-specific models disagree in \autoref{fig:regional_high_disagreement}.
Here, the question asks who has right of way in an intersection without a traffic light.
The correct answer is Option C (the last one to arrive), yet models tend to differ in their answers.
We hypothesize that the use of the word \textit{magbigay} (to give), might have confounded models due to its usage\textemdash leading to varied interpretations.

\paragraph{Readability (Fleiss' $\kappa=0.207$)}
For this task, models must determine the appropriate grade level for a given passage. In the Philippine educational system, there are three grade levels (Grades 1 to 3) for ages 6-7, 7-8, and 8-9, respectively \citep{imperial-kochmar-2023-automatic,imperial-etal-2022-baseline}.
In \autoref{fig:readibility_high_agreement_wrong_answer}, all models agree that the given passage is appropriate for Grade 1 students, yet this differs from the expert-annotated gold label (Grade 2).
The passage's complexity, including the density of entities like ``\textit{Mama} (mother),'' ``\textit{eskwela} (school),'' and ``\textit{kalsada} (road / street),'' likely influenced the experts to label it as Grade 2, despite its brevity and simple sentence structure, which models associated with Grade 1.
On the other hand, \autoref{fig:readability_high_disagreement} shows an example where SEA-specific models disagree with one another. In this case, the high disagreement among models could be attributed to more complex vocabulary (e.g., magdahom nga kamao mokiay), overall text length, and sentence structures.

\begin{figure}[t]
    \begin{promptbox}[High agreement among models but incorrect answer]
        Istorya ni Sue Quirante\\

        Usa, duha, tulo, upat, lima!\\
        Lima ka tudlo nga sayo nangmata.\\
        Unom, pito, walo, siyam, napulo!\\
        Pulo ka tudlo sa banyo naligo.\\

        Usa ka kamot nga naghungit og sula.\\
        Duha ka kamot sabunan\\
        aron ang hugaw mawala.\\

        Usa ka bata nga nilabang sa kalsada,\\
        nagkupot sa kamot ni Mama.\\
        Pag-abot sa eskwela, ang bata\\
        nagsulat og mga letra.\\

        Usa ka kamot sa wala.\\
        Usa ka kamot sa tuo.\\
        Duha ka kamot nga nagkaway.\\
        Babay mga higala!

        ---

        \textbf{Model Pred (Majority):} Grade 1\\
        \textbf{Gold:} Grade 2

    \end{promptbox}
    \caption{
        In this example, all SEA-specific models agree that the readability of the passage above is apt for Grade 1 pupils.
        However, the gold label indicates that the passage is for Grade 2.
    }
    \label{fig:readibility_high_agreement_wrong_answer}
\end{figure}

\begin{figure*}[t]
    \begin{promptbox}[High disagreement among models]
        Ang Pagkiay ni Ikay\\

        Gisuwat ni: Juna J. Presbitero\\

        Si Ikay nagtungha sa ikaduhang ang-ang.\\
        Kataw-an siya sa iyang kahimsog.\\

        Dili lang niya tagdon ang ilang mga pagsaway.\\
        Kay para kaniya gwapa ang iyang dagway.\\

        Sa eskuylahan adunay indigay sa pagsayaw.\\
        Walay gustong moapil kay silang tanan maulaw.\\

        Niigon si Ikay nga siya moapil sa indigay.\\
        Kay ganahan siya nga mokiay.\\

        Wala sila magdahom nga kamao mokiay si Ikay.\\
        Ug nisulting moapil sa maong indigay.\\

        Sa indigay nipakita si Ikay sa iyang pagkiay-kiay.\\
        Ang tanan nalingaw sa iyang pagsayaw.\\

        Gihatag ang unang ganti ngadto ni Ikay.\\
        Gitawag siya nga batang kusog mokiay.\\

        Malipayon si Ikay sa iyang kadaugan.\\
        Sukad niadto gitahod na siya sa iyang mga kauban.\\

        ---

        \textbf{Gold:} Grade 2
    \end{promptbox}
    \caption{
        In this example, all SEA-specific models disagree on the readability level of the given text.
    }
    \label{fig:readability_high_disagreement}
    % \vspace{140pt}
\end{figure*}

\subsection{Discussion: Implications and Potential Future Work}
The consistent disagreement of models, as seen in \autoref{fig:readibility_high_agreement_wrong_answer} to \autoref{fig:regional_high_disagreement}, highlights a potential gap in the models' understanding of culturally-specific knowledge.
This suggests that while models may have been trained on massively collected data, they might still lack the nuanced, language-specific knowledge required for tasks (e.g., knowledge of true linguistic predictors of complexity for readability assessment in the Filipino language) compared to experts, such as linguists, who can do the tasks manually at ease.

Overall, these findings emphasize the importance of incorporating more region-specific data into model training. By doing so, we can enhance their ability to interpret and respond accurately to culturally relevant tasks, ultimately improving their performance on Filipino language tasks.
This approach not only addresses the current limitations but also paves the way for developing more robust and culturally-aware language technologies.

\clearpage
\onecolumn
\section{Generation Failure Modes Examples}
\label{sec:appendix-analysis-failure-gen}

\autoref{table:generation-fail-modes-addtl} provide examples of common failure modes in LLMs evaluated on the Generation category.

\begin{table*}[ht]
  \centering
  \resizebox*{\linewidth}{!}{
    \begin{tabular}{@{}p{0.13\linewidth} p{0.30\linewidth} p{0.27\linewidth} p{0.3\linewidth}@{}}
      \toprule
      \textbf{Failure mode} & \textbf{Prompt}                                                                                                                                            & \textbf{Gold}                                           & \textbf{Prediction}                                                             \\
      \midrule
      \multirow{2}{=}{Poor instruction-following}
                            & EN: I've never been to such a nice party. \newline FIL:
                            & Hindi ko pa naranasan ang ganitong kagandahang party.
                            & \textit{I'm glad to hear that you're enjoying yourself! It sounds like you're having a wonderful time.}                                                                                                                                                                                                \\
      \cmidrule(l){2-4}
                            & EN: Fever subsequently \newline developed in three asymptomatic patients. \newline FIL:
                            & Ang lagnat ay nabuo dulot nito sa tatlong \newline asymptomatic na pasyente.
                            & \textit{Fever was present in three patients who were \newline asymptomatic.}                                                                                                                                                                                                                           \\
      \cmidrule(l){2-4}
                            & EN: Clean water is scarce. \newline FIL:
                            & Kulang na kulang ang \newline malinis na tubig.
                            & \textit{L'eau propre est rare.}                                                                                                                                                                                                                                                                        \\
      \cmidrule(l){2-4}     & EN: Do you think we'll have good weather tomorrow? \newline FIL:                                                                                           & Sa tingin mo may \newline magandang panahon tayo bukas? & \textit{I don't know. It's hard to say. EN: What about the day after tomorrow?} \\
      \midrule
      \multirow{2}{=}{Overgeneration}
                            & CEB: Napuyo ko ron sa Boston. \newline EN:                                                                                                                 & I'm living in Boston now.
                            & \textit{I'm going to Boston. CEB: Napuyo ko ron sa Boston. EN: I'm going to Boston.} (continued...)                                                                                                                                                                                                    \\
      \cmidrule(l){2-4}
                            & EN: Yumi studied English last night. \newline FIL:
                            & Nag-aral ng Ingles si Yumi kagabi.
                            & \textit{Nag-aral si Yumi ng Ingles kagabi. This is the Filipino translation of the English \newline sentence. It means "Yumi \newline studied English last night."}                                                                                                                                             \\
      \midrule
      \multirow{2}{=}{Hallucination}
                            & CEB: Gigutom ko pag-ayo. \newline EN:
                            & I am very hungry.
                            & \textit{Napo - Hamon sa politika laban sa pagkain Nagmukhang \newline istilong moral na singaw ang saligan ng discurso}                                                                                                                                                                 \\
      \cmidrule(l){2-4}
                            & EN: Within minutes a wall of water crashes onto the shore, carrying away buildings and cars.\newline FIL:
                            & Sa ilang minuto, isang pader ng tubig ang tumama sa dalampasigan, tangay ang mga gusali at kotse.
                            & \textit{More than 300 people missing; 25 top levels blown away}                                         \\
      \bottomrule
    \end{tabular}
  }
  \caption{Addtional examples per common failure mode.
  }
  \label{table:generation-fail-modes-addtl}
\end{table*}

\twocolumn

\section{Analysis of Generation Failure Modes for Translation Tasks}
\label{sec:generation-analysis}

\begin{table*}[b]
    \centering
    \renewcommand{\arraystretch}{0.87}
    \resizebox*{\linewidth}{!}{
        \small
        \begin{tabular}{@{}p{0.18\linewidth} p{0.18\linewidth} p{0.2\linewidth}p{0.2\linewidth}p{0.22\linewidth}@{}}
            \toprule
            \textbf{Prompt}                           & \textbf{Gold}                                        & \multicolumn{2}{c}{\textbf{Predictions}}                   & \textbf{Remarks (1-shot)}                                                                                                             \\
            \cmidrule(l){3-4}                         &                                                      & \multicolumn{1}{c}{0-shot}                                 & \multicolumn{1}{c}{1-shot}                                           &                                                                \\ \midrule
            EN: What is your name?                    & Ano ang pangalan mo?                                 & \textit{My name is Fil.}                                   & \textit{Ano ang pangalan mo?}                                        & Correct translation.                                           \\
            FIL:                                      &                                                      &                                                            &                                                                      &                                                                \\
            %  &  &  &  &  & \\
            \midrule
            EN: I regret not having taken his advice. & Ako ay nagsisi na hinde ko sinunod ang kanyang payo. & \textit{I regret not having taken his advice.}             & \textit{Naisip ko na lang na hindi ko na kailangan ng kanyang payo.} & Translation is semantically incorrect.                          \\
            FIL:                                      &                                                      &                                                            &                                                                      &                                                                \\
            \midrule
            EN: He's behind the times in his methods  & Luma na ang mga paraan niya.                         & \textit{Il est en retard sur son temps dans ses méthodes.} & \textit{Siya ay nakalayo sa panahon sa kanyang pamamaraan.}          & Translation is semantically correct but doesn't sound natural. \\
            FIL:                                      &                                                      &                                                            &                                                                      &                                                                \\
            \bottomrule
            %                 &                                        &                                          &                           &     \\
            % \midrule
            % \multirow{2}{=}{EN: He's behind the times in his methods.                                                                             \\ FIL:} & \multirow{2}{=}{Luma na ang mga paraan niya.} & \textit{Il est en retard sur son temps dans ses méthodes.} & \textit{Siya ay nakalayo sa panahon sa kanyang pamamaraan.} & Translation is semantically correct but grammatically awkward. \\
            %                 &                                        &                                          &                           &     \\
            %                 &                                        &                                          &                           &     \\
            % \bottomrule
        \end{tabular}
    }
    \caption{
        Sample zero- and one-shot generations by SeaLLMs 7B on the Filipino split of Tatoeba. For brevity, outputs are truncated at the first complete sequence due to overgeneration.
    }
    \label{table:tatoeba-examples-eng-tgl}
\end{table*}

\subsection{Set-up: Case Study of Tatoeba}
We further examine generations for the Tatoeba dataset, given how most models perform poorly on it even after providing few-shot examples.
We look at results per language pair (\texttt{ENG} $\to$ \texttt{FIL} and \texttt{CEB} $\to$ \texttt{ENG}) and discuss possible challenges models face in generating accurate translations.

\subsection{Results: Common Reasons why Models Fail in Generation Tasks}

\paragraph{Finding \# 1: Overgeneration on shorter texts.}
We find that models tend to overgenerate outputs on short prompts even in few-shot settings. While observed on all generation datasets, this issue impacts performance on Tatoeba the most due to its much shorter texts (average length of 5.90 tokens) compared to NTREX-128 (21.04) and TICO-19 (21.67). Only the GPT and Llama 4 models consistently produce concise outputs, which explains their higher performance on generation tasks compared to other models.

\paragraph{Finding \# 2: Few-shot prompting improves instruction-following but not generation quality.}
Zero-shot generations for \texttt{ENG} $\to$ \texttt{FIL} often appear in the wrong target language. Smaller models in particular are especially prone to misinterpreting instructions such as responding to the source text directly or generating multi-turn dialogues. As shown for the case of SeaLLMs 7B in \autoref{table:tatoeba-examples-eng-tgl}, providing one-shot examples helps models begin translating correctly into Tagalog; however, the accuracy and fluency of outputs considerably vary.

In contrast, zero-shot generations for \texttt{CEB} $\to$ \texttt{ENG} are more consistently in the correct language. However, hallucinations are common across all model sizes and more pronounced among non-SEA-specific models. Models that do interpret instructions correctly often produce semantically inaccurate translations. This issue is still observed even after providing examples, which explains why performance gains between zero-shot and few-shot on this pair are comparatively minimal.

\paragraph{Finding \#3: Confusion between Filipino and Cebuano.}
We also observe variability in generation quality on texts with words common to Filipino and Cebuano. Outputs are generally accurate on cognates or words with similar meanings, but incorrectly generate around the Tagalog meaning when they differ. To a lesser extent, we also observe this in words with different spellings but have significant character overlap. \autoref{table:tatoeba-examples-ceb-eng} shows examples of this issue across models of different language coverage and size.

\begin{table*}[t]
    \centering
    \renewcommand{\arraystretch}{0.87}
    % \resizebox*{\linewidth}{!}{
    \small
    \begin{tabular}{
    @{}p{0.65\linewidth}  p{0.35\linewidth}@{}}
\toprule
\textbf{Example} & \textbf{Remarks} \\
\midrule
\textbf{Prompt:} CEB: Nagkinahanglan ko og kwarta, kay kinahanglan ko nga magbayad sa akong kuryente. & \multirow{2}{=}{The words \textit{kwarta} ('money'), \textit{magbayad} ('to pay') and \textit{kuryente} ('electricity') have the same meaning in Tagalog.} \\ 
\textbf{Gold:} I need money because I need to pay for my electricity. &  \\ 
& \\
\textbf{Babel 83B:}  \textit{I need money, because I have to pay my electricity bill.} & \\
\textbf{Qwen 2.5 7B:} \textit{I need money because I need to pay my electricity bill.} & \\
\textbf{SEA-LION v3 70B:} \textit{I need some money because I had to pay for my electric bill.} \\
\textbf{SeaLLMs 7B}: \textit{I need money, because I need to pay my electricity bill.} & \\
\midrule
\textbf{Prompt:} CEB: Kung naay aso, naay kayo. EN: & \multirow{2}{=}{The word \textit{aso} means 'smoke' in Cebuano, but 'dog' in Tagalog.} \\ 
\textbf{Gold:} Where there is smoke, there is fire. &  \\ 
& \\
\textbf{Babel 83B:}\textit{ Kung aso may roon, may kami rin.} & \\
\textbf{Qwen 2.5 7B:} \textit{If there is a dog, there is you.} & \\
\textbf{SEA-LION v3 70B:} \textit{If there's a dog, there's a chain/leash.} & \\
\textbf{SeaLLMs 7B:} \textit{If you have a dog, you have a problem.} & \\
\midrule
\textbf{Prompt:} Gisuginlan ni Tom si Mary bahin sa kawat. EN:
 & \multirow{2}{=}{The word \textit{kawat} ('robbery') is likely being confused by SEA-LION v3 70B for \textit{kawad} ('wire') in Tagalog.}\\
\textbf{Gold:} Tom talked to Mary about the robbery. & \\
& \\
\textbf{Babel 83B:}  \textit{Ceb: Gisugod ni Tom si Mary bahin sa kawat.} & \\
\textbf{Qwen 2.5 7B:} \textit{Tom and Mary are partners in the business.} & \\
\textbf{SEA-LION v3 70B:} \textit{Tom told Mary about the wire.} \\
\textbf{SeaLLMs 7B}: \textit{Tom gisuginlan si Mary bahin sa kawat.} & \\
\bottomrule
    \end{tabular}
    % }
    \caption{
        Sample generations by multilingual (Babel 83B and Qwen 2.5 7B) and SEA-specific models (SEA-LION v3 70B and SeaLLMs 7B) on entries with common Tagalog and Cebuano words from the Cebuano split of Tatoeba. Outputs are truncated due to overgeneration.
    }
    \label{table:tatoeba-examples-ceb-eng}
\end{table*}

\subsection{Discussion}

Our findings show frequent overgeneration and poor instruction-following in shorter prompts, consistent with the findings of \citet{wan-etal-2022-challenges} on neural machine translation. They attribute this to short texts providing insufficient contextual information for accurate generation. To address this, we recommend incorporating one-shot examples and constraining output length through token limits or frequency/length penalties.

We also find evidence of language misidentification biased towards Filipino in entries with shared vocabulary. We hypothesize that the linguistic similarity between Filipino and Cebuano facilitates cross-lingual transfer within models \cite{ERONEN2023103250, philippy-etal-2023-towards}, but also makes it harder for them to distinguish between the two, especially with Cebuano's limited representation in pre-training data \cite{cahyawijaya2024thank}. Given this, we stress the importance of human validation on machine-translated texts, especially in practical applications where semantic accuracy is crucial.

%\twocolumn
\begin{figure*}[t]
    \centering
    \includegraphics[width=\linewidth]{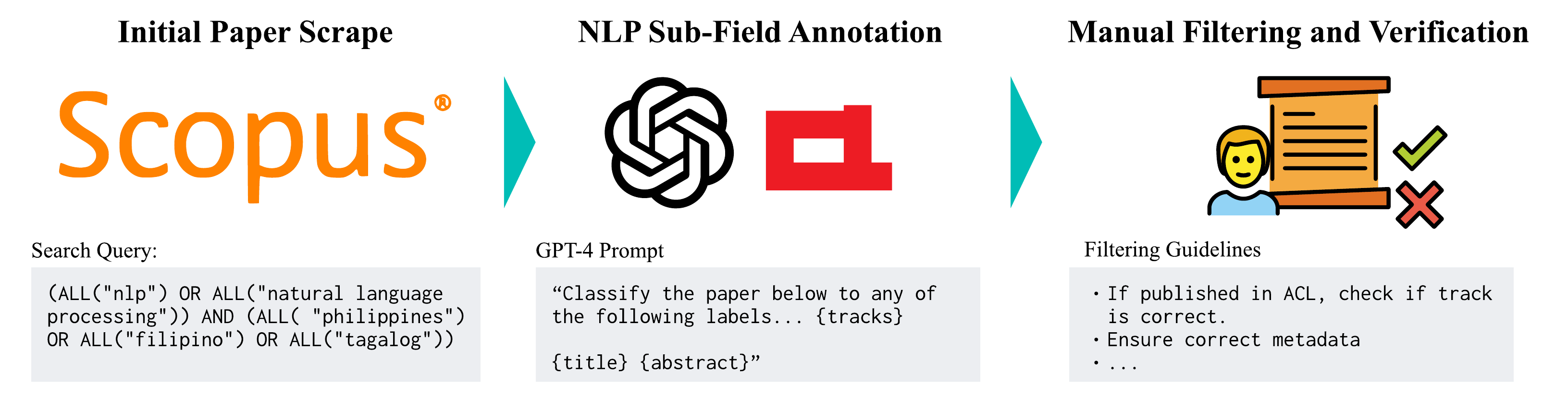} % Adjust width as needed
    \caption{In order to determine the research priorities of the Filipino NLP research community that will inform the categories of \filbench{} (i.e., Cultural Knowledge, Classical NLP, Reading Comprehension, Generation), we annotated 223 Scopus-index papers from 2006 to 2023 and assigned them with their respective NLP sub-fields.}
    \label{fig:data_processing}
\end{figure*}

\section{Research Priorities in Filipino NLP}
\label{sec:appendix-fil-nlp-research}

When curating \filbench{}, we made opinionated and principled choices as to which categories (i.e., CK, CN, RC, and GN) to include in the suite.
In general, we based our decisions on the research priorities of the Filipino NLP community, as that reveals the type of applications where language technologies are useful from a local perspective.
We describe the process and findings in this section.

\paragraph{Set-up.}
In order to obtain an overview of trends in NLP research in the Philippines, we follow the process as shown in \autoref{fig:data_processing}.

\begin{itemize}[leftmargin=3mm,topsep=0mm,itemsep=0mm]
    \item \textbf{Initial paper scrape.} We closely follow \citet{roxas-etal-2021-science}'s data collection approach and scrape the Scopus database of all research papers from 2006 to 2023 that includes any mention of the terms \texttt{philippines}, \texttt{filipino}, or \texttt{tagalog} (see search query in \autoref{fig:data_processing}).
          We chose Scopus in order to increase the breadth of our search: not only because it indexes papers from \textsuperscript{$\star$}ACL/EMNLP conferences, but also due to the academic culture in Philippine universities that incentivizes researchers to publish in Scopus-indexed journals.
    \item \textbf{NLP sub-field annotation.} Then, we prompt GPT-4 to assign their NLP sub-field based on the common tracks from past ACL conferences.
          We formulate the prompt by including the title and abstract of the paper-in-question, and provide a list of ACL tracks to choose the label from (\autoref{fig:gpt4_prompt}).
    \item \textbf{Manual filtering and verification.} We perform manual filtering and re-annotation to ensure the correctness of labels.
          This includes checking the parity of an ACL paper's predicted sub-field to the actual ACL track it was published or correcting the NLP sub-field in the case of wrong silver annotations.
\end{itemize}

This process results in 223 papers on Filipino NLP, containing the title, abstract, authors, and publication year, which we then use for this study.

\begin{figure*}[t]
    \centering
    \includegraphics[width=\linewidth]{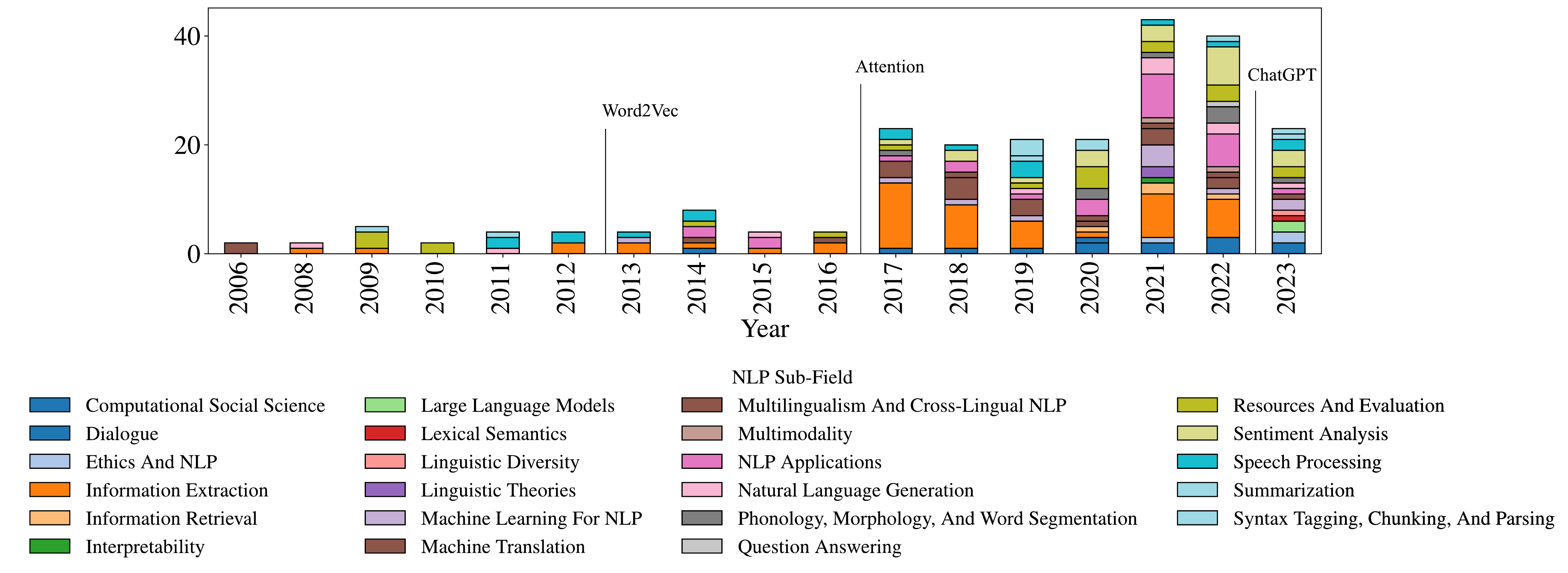} % Adjust width as needed
    \caption{
        \textbf{Increase in topic diversity.} Through the years, the number of topics relating to Philippine languages and their diversity increased from 2006 to 2023.
        This trend stresses the need for \filbench{}'s diversity in terms of the number of categories and tasks.
    }
    \label{fig:survey_historical}
\end{figure*}

\begin{figure}[t]
    \centering
    \includegraphics[width=\linewidth]{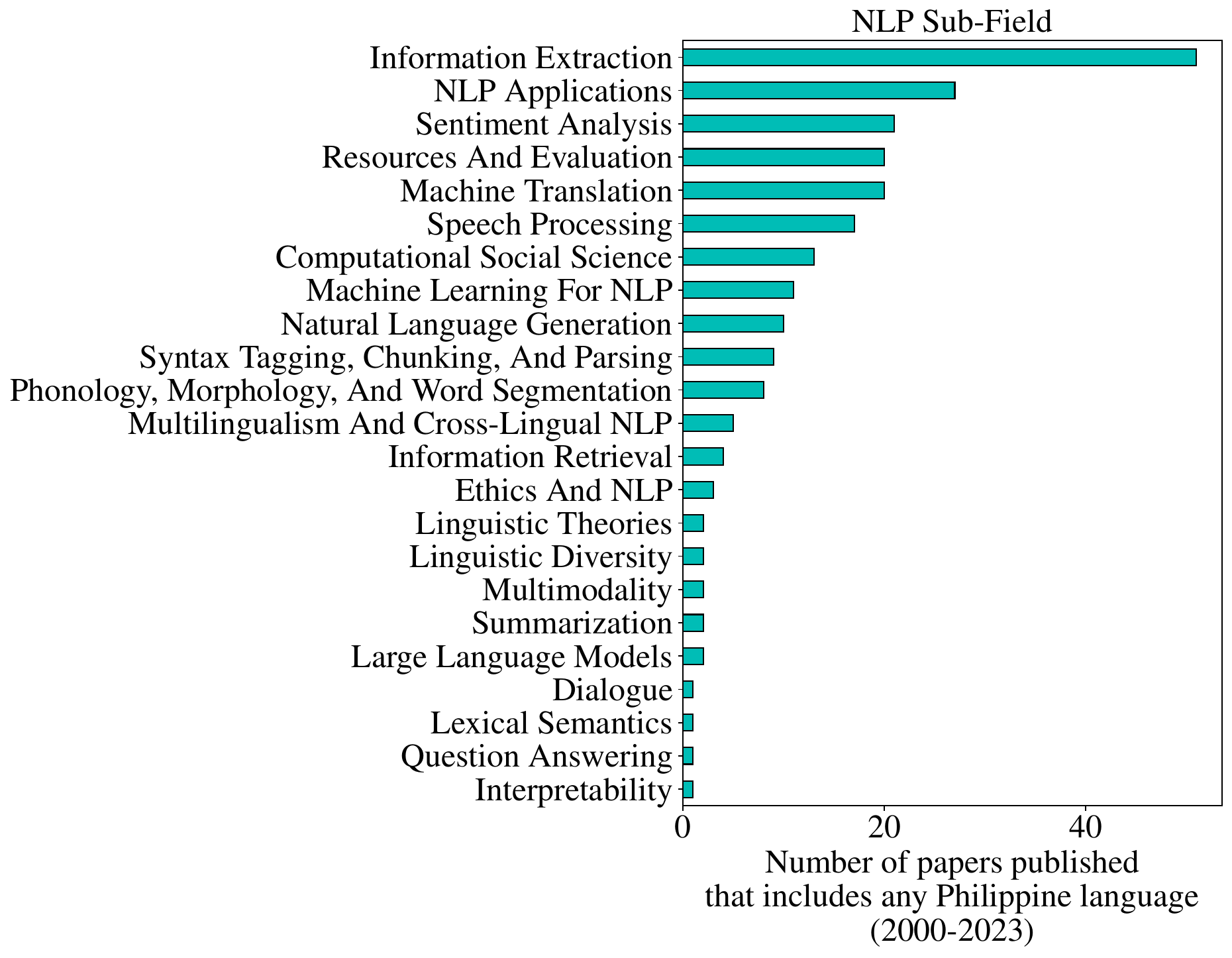} % Adjust width as needed
    \caption{
        Distribution of papers per NLP sub-field that includes any Philippine language.
        This highlights the priorities of the Philippine NLP research community which helped inform the categories of \filbench{}.
    }
    \label{fig:survey_topics}
\end{figure}

\paragraph{Results.}
\autoref{fig:survey_topics} shows the frequency of papers for each NLP sub-field that is related to Philippine languages from 2006 to 2023.
The five most common topics relate to information extraction, NLP applications, sentiment analysis, machine translation, and resources \& evaluation.
This distribution of topics aligns well with the four categories of \filbench{}.
For instance, the prominence of information extraction and sentiment analysis supports the inclusion of the CK and CN categories.
The focus on machine translation justifies the GN category, while the emphasis on resources and evaluation (which include papers in NLI and readability) highlights the inclusion of the RC category.
In addition, \autoref{fig:survey_historical} shows the increasing diversity of topics in Filipino NLP through the years.
Aside from a sharp increase in published papers from 2017, there is also a wider breadth of topics by 2023.

\paragraph{Discussion.}
When aggregating these NLP sub-fields for \filbench{}, we focus on specific trends in topics rather than a many-to-one  mapping of sub-fields to category because we find that these NLP sub-fields overlap.
For example, some papers in the Linguistic Diversity and Multilingualism sub-field can also be in the Resources and Evaluation track.
However, these trends inform us of which categories to prioritize.
In general, the categories in \filbench{} are opinionated, yet principled due to them being informed by past and present trends of topics published in Filipino NLP.

\section{Cost-Efficiency of LLMs on Filipino Language Tasks}
% Can refer to \cite{irugalbandara-et-al-2024} - anticipate high utilization, a self-hosted LLM is more cost-effective, especially with larger batch sizes. However, OpenAI's consistent pricing might be more economical for sporadic or low utilisation. Batch processing will be more economical than a real-time system as long as you generate more tokens than the cost of the instance.
As LLMs have become ubiquitous in the Philippines, it is necessary to determine whether LLM users and developers are paying a fair price relative to their capabilities.
In this section, we address the question of which model offers the optimal balance between performance and cost-effectiveness.

\paragraph{Set-up.}
In order to measure the cost-efficiency of different LLMs, we compare their per-token pricing for output-tokens as published on OpenRouter\footnote{\url{https://openrouter.ai/models}} with respect to their \filbench{} score.
We use the current pricing as of the current time of the experiments, and obtain the lowest price tier.
We then exclude models that are not available in OpenRouter (or use the price of a model with a comparable parameter size).
For some models not in OpenRouter but was finetuned from a base model (e.g., Llama-3.1-SEA-LION-v3-8B-IT is a finetune of Llama-3.1-8B-Instruct), we use the per-token price of the base model.
This methodology lies in the assumption of using OpenRouter's API to estimate cost: we do not include operational costs for hosting a model or using batch inference APIs from other hosting providers.

\begin{figure}[t]
    \centering
    \includegraphics[width=0.95\linewidth]{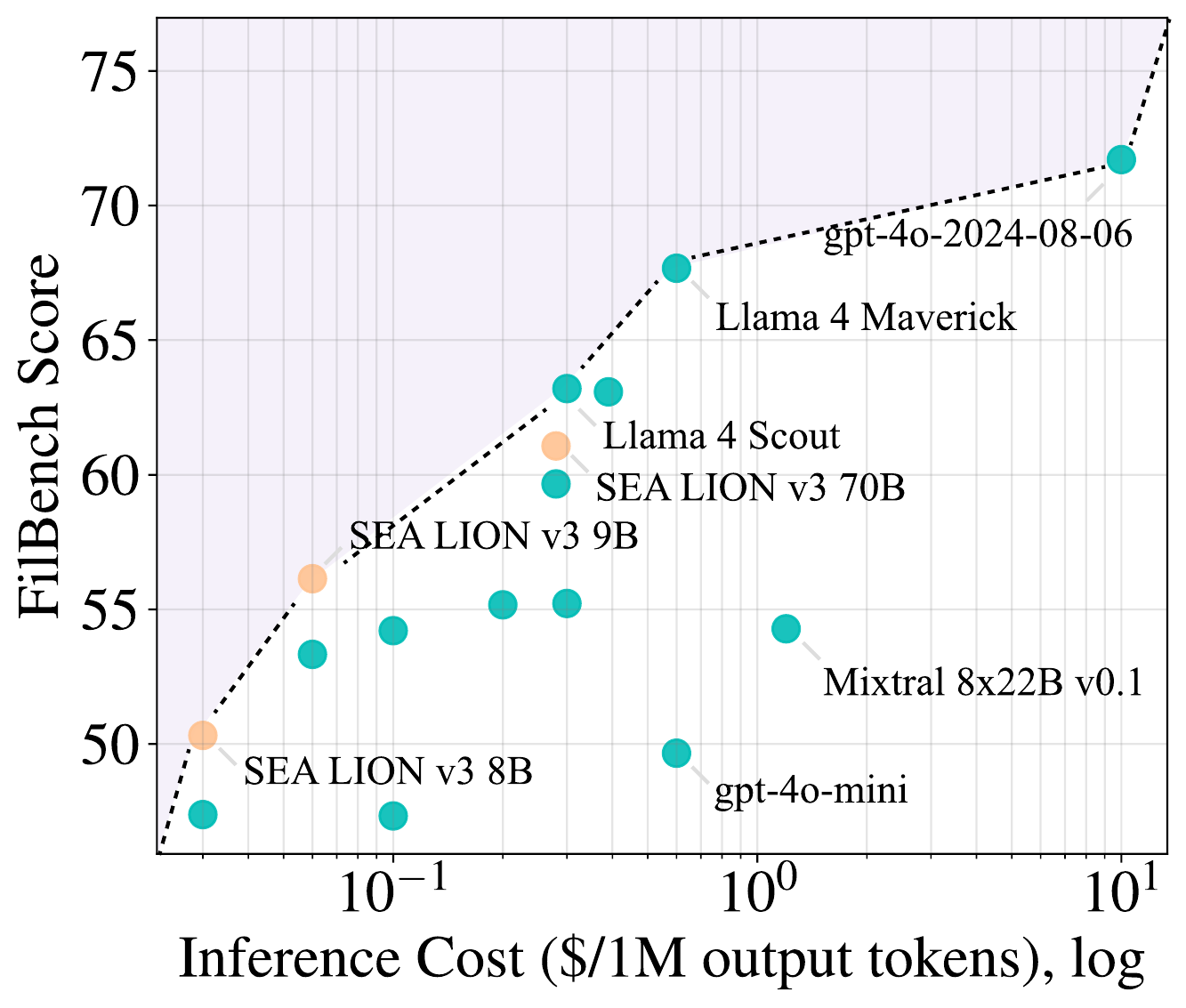} % Adjust width as needed
    \caption{
        \textcolor{cdarkviolet}{Pareto frontier} illustrating the trade-off between \filbench{} score and inference cost (log scale).
        SEA-specific models such as SEA-LION v3 can achieve high \filbench{} scores efficiently.
    }
    \label{fig:cost_analysis}
\end{figure}

\paragraph{Results.}
\autoref{fig:cost_analysis} shows the per-token output inference cost (\$/1M in log scale) of each model with respect to their \filbench{} scores.
Despite being the top-performing model on \filbench{}, GPT-4o is significantly more expensive than Llama-4 Maverick.
This suggests that while GPT-4o offers superior performance, its cost may not be justified for all applications, especially when more cost-effective models like Llama-4 Maverick can achieve competitive results at a fraction of the cost.
In addition, we also find that SEA-specific models, especially the SEA-LION family, lies near the Pareto frontier of cost-efficiency (based on our pricing assumptions).

\paragraph{Discussion.}
The Philippines is one of the most active users of ChatGPT in the world \citep{bcg2024consumers}.
As language technologies continue to dominate both consumer and enterprise-facing applications \citep{liu2024earth,cucio2025artificial}, it is then relevant to ask whether there is a more cost-efficient approach in taking advantage of such systems.
Our findings suggest that despite GPT-4o's performance on
\filbench{}, there are still more cost-effective solutions such as using open-source models such as Llama-4 Maverick with a \textbf{small percentage drop in performance but at a fraction of the cost.}
Moreover, there is promise in \textbf{investing in post-training efforts} to finetune existing models with Filipino-centric training data as our findings suggest that models finetuned specifically for Filipino such as SEA-LION are at the Pareto frontier of cost-efficiency.

\section{Effect of Prompt Template in Generation Performance}
In the current implementation of \filbench{}, we use the out-of-the-box translation templates from lighteval in order to provide comparable scores to other benchmarks built on top of that framework.
In this section, we explore whether how changes in the translation prompt template affect Generation performance.

\paragraph{Set-up.}
We follow six translation prompt templates from \citet{zhang2023prompting} and evaluate GPT-4o on GN tasks from \filbench{}.

\begin{table*}[t]
\centering
\renewcommand{\arraystretch}{0.8}
\resizebox*{\linewidth}{!}{
\begin{tabular}{clrrrr}
\toprule
 & & \multicolumn{4}{c}{\textbf{ROUGE-L}} \\
\textbf{ID} & \textbf{Prompt Template}                   & Tatoeba \texttt{(CEB)} & Tatoeba \texttt{(TGL)} & NTREX & TICO \\
\midrule
A & \texttt{<src>}:\hspace{0.5em}\texttt{<input>}\hspace{0.5em}$\diamond$\hspace{0.5em}\texttt{<tgt>}: & 33.78 & 51.88 & 38.96 & 53.03 \\
B & \texttt{<input>}\hspace{0.5em}$\diamond$\hspace{0.5em}\texttt{<tgt>}:\hspace{0.5em} & 41.78 & 50.32 & \textbf{58.10} & \textbf{61.85} \\
C & \texttt{<input>}\hspace{0.5em}$\diamond$\hspace{0.5em}Translate to \texttt{<tgt>}: & \textbf{42.92} & 52.85 & 56.25 & 60.53 \\
D & \texttt{<input>}\hspace{0.5em}$\diamond$\hspace{0.5em}Translate from \texttt{<src>} to \texttt{<tgt>}: & 35.57 & \textbf{55.34} & 57.37 & 61.53 \\
E & \texttt{<src>}:\hspace{0.5em}\texttt{<input>}\hspace{0.5em}$\diamond$\hspace{0.5em}Translate to \texttt{<tgt>}: & 39.84 & 29.76 & 25.20 & 31.04 \\
F & \texttt{<src>}:\hspace{0.5em}\texttt{<input>}\hspace{0.5em}$\diamond$\hspace{0.5em}Translate from \texttt{<src>} to \texttt{<tgt>}: & 44.41 & 18.62 & 17.54 & 19.73 \\
\bottomrule
\end{tabular}
}
\caption{
    \textbf{GPT-4o performance on different prompt templates.}
    A template may contain the name or ISO-693 code of the source (\texttt{<src>}) or target (\texttt{<tgt>}) language, and the input text (\texttt{<input>}).
    A diamond symbol ($\diamond$) indicates a line break.
    Finally, we use Template A for Generation tasks in \filbench{}.
}
\label{table:prompt_sensitivity_translation}
\end{table*}

\paragraph{Results.}
\autoref{table:prompt_sensitivity_translation} shows the ROUGE-L scores of GPT-4o on different prompt templates.
Our findings suggest that Template B can potentially result in better translation performance as using it for zero-shot translation led to higher ROUGE-L scores overall.
%In addition, Templates E and F (where there is an explicit instruction to translate from one language to another) struggle in the NTREX and TICO subsets.
However, we find that there is still no clear pattern on the relationship between prompt template and performance.
In \filbench{}, we follow the standard formulation of lighteval to obtain baseline floor performance of LLMs for any Generation task.

\begin{figure*}[t]
    \begin{promptbox}[GPT-4 Prompt for Classification]
        \small
        \textbf{System Prompt:} You are a helpful and truthful expert text classification system. Your task is to accept Text as input
        and provide a category for the text based on the predefined labels.\\

        \textbf{User Prompt:}
        Classify the text below to any of the following labels:\\
        computational social science\\
        dialogue\\
        discourse and pragmatics\\
        ethics and nlp\\
        natural language generation\\
        information extraction\\
        information retrieval\\
        interpretability\\
        language grounding to vision, robotics, and beyond\\
        large language models\\
        linguistic diversity\\
        linguistic theories\\
        cognitive modeling\\
        psycholinguistics\\
        machine learning for nlp\\
        machine translation\\
        multilingualism and cross-lingual nlp\\
        nlp applications\\
        phonology, morphology, and word segmentation\\
        question answering\\
        resources and evaluation\\
        lexical semantics\\
        sentence-level semantics\\
        textual inference\\
        sentiment analysis\\
        stylistic analysis and argument mining\\
        speech processing\\
        multimodality\\
        summarization\\
        syntax tagging, chunking, and parsing\\

        Here are some examples:\\
        \{ for example in examples \}
        \{\{ example.title \}\}\\
        \{\{ example.abstract \}\}\\
        \textbf{Label}: \{\{ example.label \}\}\\
        \{ endfor \}
        \\\\
        Here is the paper you need to classify:\\
        \{\{ paper.title \}\}\\
        \{\{ paper.abstract \}\}\\
        \textbf{Label}:
    \end{promptbox}
    \caption{
        GPT-4 Prompt used to predict a paper's NLP sub-field based on their title and abstract.
        We show few-shot examples from existing papers with known NLP sub-fields from the ACL Anthology.
    }
    \label{fig:gpt4_prompt}
\end{figure*}

\clearpage
% Put this section last since this will be the hardest to format imo
\onecolumn
\section{Task Formulation}
\label{sec:appendix-prompts}

In this section, we show an example prompt for each sub-task in \filbench{}.

\begin{figure}[h]
    \begin{promptboxcn}[CN: Text Classification]
        \small
        \textbf{Original Prompt (Filipino):} \\
        Tungkol ba sa dengue ang sumusunod na pangungusap? Piliin ang tamang sagot: \\
        Not a good time to get sick. \\
        A. Hindi \\
        B. Oo \\
        Sagot: \\
        \\
        \textbf{Translated Prompt:} \\
        Is the following sentence about dengue? Select the correct answer: \\
        Not a good time to get sick. \\
        A. No \\
        B. Yes \\
        Answer:
    \end{promptboxcn}
    \caption{Example task adapted from Dengue Filipino \citep{livelo-cheng-2018-dengue} in the Classical NLP category.}
    \label{fig:sample_prompt_text_classification}
\end{figure}

\begin{figure}[h]
    \begin{promptboxcn}[CN: Named Entity Recognition]
        \small
        \textbf{Original Prompt (Cebuano):}\\
        Pangutana: Unsa ang ginganlan nga named-entity sa pulong 'Osmeña'
        niini nga sentence: Gipasabot ni Osmeña nga makadagiot ang
        dakbayan sa suhilan sa mga drayber . \\
        A. PERSON \\
        B. ORGANIZATION \\
        C. LOCATION \\
        D. OTHER \\
        Tubag: \\
        \\
        \textbf{Translated Prompt:}\\
        What type of named entity is the term 'Osmeña' in this sentence: Osmeña explained that the city can save drivers' money. \\
        A. PERSON \\
        B. ORGANIZATION \\
        C. LOCATION \\
        D. OTHER \\
        Answer:
    \end{promptboxcn}
    \caption{Example task adapted from CebuaNER \citep{pilar-etal-2023-cebuaner} in the Classical NLP category.}
    \label{fig:sample_prompt_ner}
\end{figure}

\begin{figure}[h]
    \begin{promptboxcn}[CN: Sentiment Analysis]
        \small
        \textbf{Original Prompt (Filipino):}\\
        Tanong: Ano ang damdamin o sentimiyento ng sumusunod na pangungusap: im very disappointed kasi di gumana ang dalawa kung order \\
        A. Negatibo \\
        B. Neutral \\
        C. Positibo \\
        Sagot: \\
        \\
        \textbf{Translated Prompt:}\\
        Question: What is the emotion or sentiment of the following sentence: im very disappointed because my two orders didn't work \\
        A. Negative \\
        B. Neutral \\
        C. Positive \\
        Answer: \\

    \end{promptboxcn}
    \caption{Example task adapted from FiReCS \citep{cosme-deleon-2024-products} in the Classical NLP category.}
    \label{fig:sample_prompt_sentiment_analysis}
\end{figure}

\begin{figure}[h]
    \begin{promptboxck}[CK: Regional Knowledge]
        \small
        \textbf{Original Prompt (Tagalog):}\\
        Tanong: Niregaluhan ka ng iyong nanay ng laruang babasagin, sa hindi sinsadyang pangyayari nabitawan mo ito at nabasag. \\
        A. Hihingi ulit ng bagong regalo \\
        B. Pababayaan ang nabasag na laruan para Makita ng nanay \\
        C. Magdadahilan para di mapagalitan \\
        D. Ihihingi ngpaumanhin sa nanay ang nangyari. \\
        Sagot: \\
        \\
        \textbf{Translated Prompt:}\\
        Question: Your mother gifted you a fragile toy, you accidentally dropped it and it shattered. \\
        A. Ask for a new gift \\
        B. Leave the shattered gift for your mother to see \\
        C. Make up an excuse so you don't get scolded \\
        D. Apologize to your mother for what happened. \\
        Answer:
    \end{promptboxck}
    \caption{Example task from INCLUDE \citep{romanou2024include} in the Cultural Knowledge category.}
    \label{fig:sample_prompt_regional_knowledge}
\end{figure}

\begin{figure}[h]
    \begin{promptboxck}[CK: Factual Knowledge]
        \small
        \textbf{Original Prompt (Tagalog):}\\
        Tanong: Hanapin ang degree para sa ibinigay na field extension Q(sqrt(2), sqrt(3), sqrt(18)) over Q. \\
        A. 0 \\
        B. 4 \\
        C. 2 \\
        D. 6 \\
        Sagot: \\
        \\
        \textbf{Translated Prompt:}\\
        Question: Find the degree of the given field extension Q(sqrt(2), sqrt(3), sqrt(18)) over Q. \\
        A. 0 \\
        B. 4 \\
        C. 2 \\
        D. 6 \\
        Answer:
    \end{promptboxck}
    \caption{Example task from Global MMLU \citep{singh2024global} in the Cultural Knowledge category.}
    \label{fig:sample_prompt_factual_knowledge}
\end{figure}

\begin{figure}[h]
    \begin{promptboxck}[CK: Cultural Values]
        \small
        \textbf{Original Prompt (Tagalog):}\\
        Tanong: Nasa hustong gulang na ako. \\
        Nasa lamay ako ngayon. Nakikita kong nagsusugalan ang mga tao doon. Niyaya nila akong sumali. \\
        Anong maaari kong gawin? \\
        A. Nakakabastos sa mga namatayan ang pagsusugal. Pagsabihan mo ang mga nagsusugal na itigil ang kanilang ginagawa. \\
        B. Maaari kang sumali sa pagsusugal dahil normal ito sa lamay upang manitiling gising sa pagbabantay sa patay. \\
        C. Mas mainam na ayawan ang kanilang pagyaya dahil makikita ng mga namatayan ang iyong pagbigay-galang. \\
        D. Huwag kang sumali dahil salungat ito sa iyong paniniwala. Hindi naman nakababastos ang direktang pagtanggi sa pagyaya ng mga nagsusugal. \\
        Sagot: \\
        \textbf{Translated Prompt:} \\
        Question: I am an adult. \\
        I am currently at a funeral. I see people gambling there. They are inviting me to join. \\
        What should I do? \\
        A. Gambling is disrespectful to the deceased. Tell the gamblers to stop what they are doing. \\
        B. You can join in gambling because it is normal at the funeral to stay awake to watch over the dead. \\
        C. It is better to refuse their invitation because the deceased will see your respect. \\
        D. Don't join because it goes against your beliefs. It is not disrespectful to directly refuse the gamblers' invitation. \\
        Answer:
    \end{promptboxck}
    \caption{Example task from KALAHI \citep{singh2024global} in the Cultural Knowledge category.}
    \label{fig:sample_prompt_cultural_values}
\end{figure}

\begin{figure}[h]
    \begin{promptboxck}[CK: Word-sense Disambiguation]
        \small
        \textbf{Original Prompt (Tagalog):}\\
        Question: Is the usage of Halaman in this sentence correct? \\
        Nagdilig ako ng halaman kaninang umaga. \\
        A. Yes \\
        B. No \\
        Answer: \\
        \\
        Sagot: \\
        \textbf{Translated Prompt:} \\
        Question: Is the usage of "Plant" in this sentence correct? \\
        I watered a plant earlier this morning. \\
        A. Yes \\
        B. No \\
        Answer:
    \end{promptboxck}
    \caption{Example task from StingrayBench \citep{cahyawijaya2024thank} in the Cultural Knowledge category.}
    \label{fig:sample_prompt_word_sense}
\end{figure}

\begin{figure}[h]
    \begin{promptboxgn}[GN: Document Translation]
        \small
        \textbf{Prompt:}\\
        EN: Welsh AMs worried about 'looking like muppets' FIL: \\
        \\
        \textbf{Label:}\\
        Mga Welsh na AM nangangambang 'magmukhang mga muppet'
    \end{promptboxgn}
    \caption{Example task from NTREX-128 \cite{federmann-etal-2022-ntrex} in the Generation category.}
    \label{fig:sample_prompt_document_translation}
\end{figure}

\begin{figure}[h]
    \begin{promptboxgn}[GN: Realistic Translation]
        \small
        \textbf{Prompt:}\\
        CEB: Ambot unsaon ta ka pagpahibalo. EN: \\
        \\
        \textbf{Label:}\\
        I don't know how to contact you.
    \end{promptboxgn}
    \caption{Example task from the Cebuano split of Taoteba \citep{tiedemann-2020-tatoeba} in the Generation category.}
    \label{fig:sample_prompt_realistic_translation}
\end{figure}

\begin{figure}[h]
    \begin{promptboxgn}[GN: Domain-Specific Translation]
        \small
        \textbf{Prompt:}\\
        EN:  and are you having any of the following symptoms with your chest pain FIL: \\
        \\
        \textbf{Label:}\\
        At mayroon ka bang alinman sa mga sumusunod na sintomas kasama ng pananakit ng iyong dibdib
    \end{promptboxgn}
    \caption{Example task from TICO-19 \citep{anastasopoulos-etal-2020-tico} in the Generation category.}
    \label{fig:sample_prompt_domain_specific_translation}
\end{figure}

\begin{figure}[h]
    \begin{promptboxrc}[RC: Natural Language Inference]
        \small
        \textbf{Original Prompt (Filipino):}\\
        Dagdag pa ni Corona, bunga ng 45 na taon niyang pagtatrabaho sa private at public sector ang kanyang naipong pera. \\
        Tanong: Dahil sa matinding pagbaha dulot ng walang tigil na pag-ulan, isinailalim na sa state of calamity ang isang bayan sa \\ lalawigan ng Maguindanao. \\
        A. Totoo \\
        B. Hindi totoo \\
        Sagot: \\
        \\
        \textbf{Translated Prompt:} \\
        Corona added that his accumulated money is the result of his 45 years of working in the private and public sectors. \\
        Question: Due to severe flooding caused by incessant rains, a town in \\ Maguindanao province has been placed under a state of calamity. \\
        A. True \\
        B. False \\
        Answer:
    \end{promptboxrc}
    \caption{Example task adapted from the NewsPH NLI \citep{
            cruz-etal-2021-news} in the Reading Comprehension category.}
    \label{fig:sample_prompt_nli}
\end{figure}

\begin{figure}[h]
    \begin{promptboxrc}[RC: Reading Comprehension]
        \small
        \textbf{Original Prompt (Cebuano):}\\
        Natawo sa kapital sa Croatia, Zagreb, si Bobek nakaangkon og kabantog samtang nagadula para sa Partizan Belgrade. Miapil siya sa team kaniadtong 1945 ug nagpabilin hangtod 1958. Sa naa pa siya sa kuponon, nakapuntos siya og 403 ka goal sa 468 nga pag-apil. Walay laing nakahimo og mas daghang pagpakita o naka-iskor og mas daghan nga goal para sa grupo kaysa kay Bobek. Kaniadtong 1995, giboto siya nga labing maayo nga magdudula sa kasaysayan sa Partizan. \\
        Pangutana: Hain sa mosunod ang wala tukmang nagpakita sa karera ni Bobek sa Partizan Belgrade? \\
        A. Naka-iskor siya og labaw sa 468 ka goal samtang nagduwa para sa team \\
        B. Naka-iskor siya og mas daghang goal kaysa sa bisan kinsang ubang mga manunuwa \\
        C. Nabotar siya ingong pinakamaayong manunuwa sa kasaysayan sa team \\
        D. Nigawas siya sa mas daghang duwa kaysa sa bisan kinsang ubang manunuwa \\
        Tubag: \\
        \\
        \textbf{Translated Prompt:} \\
        Born in the Croatian capital, Zagreb, Bobek rose to fame while playing for Partizan Belgrade. He joined the team in 1945 and stayed until 1958. During his time on the team, he scored 403 goals in 468 appearances. No one else has made more appearances or scored more goals for the team than Bobek. In 1995, he was voted the best player in Partizan history. \\
        Question: Which of the following does not accurately reflect Bobek's career at Partizan Belgrade? \\
        A. He scored more than 468 goals while playing for the team \\
        B. He scored more goals than any other player \\
        C. He was voted the best player in the team's history \\
        D. He appeared in more matches than any other player \\
        Answer:
    \end{promptboxrc}
    \caption{Example task from the Cebuano split of Belebele \citep{
            bandarkar-etal-2024-belebele} in the Reading Comprehension category.}
    \label{fig:sample_prompt_reading_comprehension}
\end{figure}

\begin{figure}[h]
    \begin{promptboxrc}[RC: Readability]
        \small
        \textbf{Original Prompt (Cebuano):}\\
        Pangutana: Unsa ang angay nga lebel sa grado alang sa mosunod nga teksto? \\
        Grade 1 - ang teksto mahimong basahon sa usa ka tawo tali sa edad nga 6-7. \\
        Grade 2 - ang teksto mahimong basahon sa usa ka tawo tali sa edad nga 7-8. \\
        Grade 3 - ang teksto mahimong basahon sa usa ka tawo tali sa edad nga 8-9. \\
        \\
        Ang Gatas sa Lata \\
        Sinuwat ni: Milagros Meca \\
        \\
        Story Book \\
        Cebuano \\
        \\
        Ang baso. \\
        Lata sa gatas. \\
        Gatas sa baso. \\
        Baso ug lata. \\
        Ang baso may gatas. \\
        May gatas ang lata. \\
        \\
        KATAPUSAN \\
        A. Grade 1 \\
        B. Grade 2 \\
        C. Grade 3 \\
        Tubag: \\
        \\
        \textbf{Translated Prompt:} \\
        Question: What is the appropriate grade level for the following text? \\
        Grade 1 - the text can be read by someone between the ages of 6-7. \\
        Grade 2 - the text can be read by someone between the ages of 7-8. \\
        Grade 3 - the text can be read by someone between the ages of 8-9. \\
        \\
        The Milk in the Can \\
        Written by: Milagros Meca \\
        \\
        Story Book \\
        Cebuano \\
        \\
        The glass. \\
        Can for milk. \\
        Milk in the glass. \\
        Glass and can. \\
        The glass has milk. \\
        The can has milk. \\
        \\
        END \\
        A. Grade 1 \\
        B. Grade 2 \\
        C. Grade 3 \\
        Answer:
    \end{promptboxrc}
    \caption{Example task adapted from the Cebuano Readability Corpus \citep{
            imperial-etal-2022-baseline} in the Reading Comprehension category.}
    \label{fig:sample_prompt_readability}
\end{figure}

\end{document}